%% file: main.tex
\documentclass[10pt, a4paper, logo, copyright]{googledeepmind} %

\usepackage{times}

\usepackage{amsmath,amsfonts,bm}
\usepackage{longtable}
\usepackage{hyperref}
\usepackage{url}
\usepackage{graphicx}
\usepackage{subcaption}
\usepackage{booktabs} %
\usepackage{wrapfig} %
\usepackage{enumitem}
\usepackage{hyphenat}
\usepackage{fontawesome}
\usepackage{makecell}
\usepackage{caption}
\usepackage{hyperref}

\usepackage{mathtools}
\usepackage{amsthm}
\usepackage[most]{tcolorbox}
\usepackage{moresize}
\usepackage{listings}
\usepackage{courier}
\usepackage{array}

\usepackage[numbers]{natbib}
\usepackage{graphicx}
\usepackage{booktabs}
\usepackage{multirow}
\usepackage{float} 
\usepackage{arydshln}
\usepackage{parskip}
\usepackage{stfloats}
\usepackage{makecell}
\usepackage{pifont}
\usepackage{longtable}
\usepackage{cuted}    
\usepackage{caption}  
\usepackage{enumitem}
\usepackage{float}
\usepackage[capitalize,noabbrev]{cleveref}
\usepackage[numbers]{natbib}
\usepackage{booktabs}
\usepackage{tabularx}
\usepackage{threeparttable}

\renewcommand{\arraystretch}{0.6}
\newcolumntype{C}[1]{>{\centering\arraybackslash}m{#1}}
\newcommand{\cmark}{\ding{51}}
\newcommand{\xmark}{\ding{55}}

\title{A Survey on Vision-Language-Action Models: An Action Tokenization Perspective}

\author[1,2*]{Yifan Zhong}
\author[2*]{Fengshuo Bai}
\author[1,2]{Shaofei Cai}
\author[1,2]{Xuchuan Huang}
\author[1,2]{Zhang Chen}
\author[1,2]{Xiaowei Zhang}
\author[2,3]{Yuanfei Wang}
\author[1,2]{Shaoyang Guo}
\author[1,2]{Tianrui Guan}
\author[1,2]{Ka Nam Lui}
\author[1,2]{Zhiquan Qi}
\author[1,2]{\\Yitao Liang}
\author[1,2\dag]{Yuanpei Chen}
\author[1,2\dag]{Yaodong Yang}
\correspondingauthor{Yuanpei Chen\{yuanpei.chen312@gmail.com\},Yaodong Yang\{yaodong.yang@pku.edu.cn\}}

\affil[1]{Institute for AI, Peking University}
\affil[2]{PKU-PsiBot Joint Lab}
\affil[3]{School of Computer Science, Peking University}

\begin{abstract}

The remarkable advancements of vision and language foundation models in multimodal understanding, reasoning, and generation has sparked growing efforts to extend such intelligence to the physical world, fueling the flourishing of vision-language-action (VLA) models. Despite seemingly diverse approaches, we observe that current VLA models can be unified under a single framework: vision and language inputs are processed by a series of VLA modules, producing a chain of \textit{action tokens} that progressively encode more grounded and actionable information, ultimately generating executable actions. We further determine that the primary design choice distinguishing VLA models lies in how action tokens are formulated, which can be categorized into language description, code, affordance, trajectory, goal state, latent representation, raw action, and reasoning. However, there remains a lack of comprehensive understanding regarding action tokens, significantly impeding effective VLA development and obscuring future directions. Therefore, this survey aims to categorize and interpret existing VLA research through the lens of action tokenization, distill the strengths and limitations of each token type, and identify areas for improvement. Through this systematic review and analysis, we offer a synthesized outlook on the broader evolution of VLA models, highlight underexplored yet promising directions, and contribute guidance for future research, hoping to bring the field closer to general-purpose intelligence.

\end{abstract}

\begin{document}

\maketitle

\begin{figure}[h]
    \vspace{20pt}
    \centering
    \includegraphics[width=\linewidth]{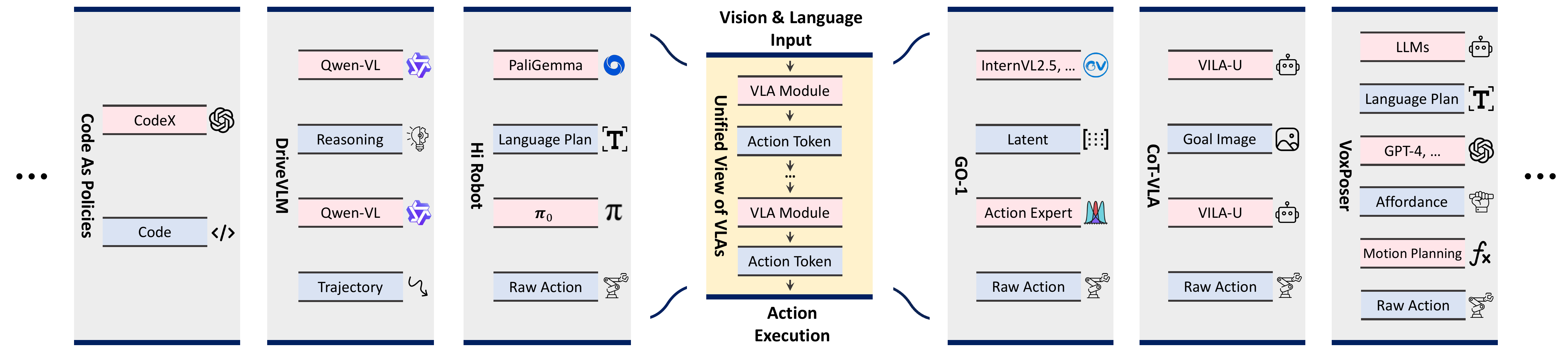}
    \caption{We present a unified framework of VLA from an \textit{action tokenization} perspective. Action token refers broadly to any descriptive guidance iteratively generated by VLAs that ultimately leads to action execution, extending beyond the notion of raw action.}
    \label{fig:unified_framework}
\end{figure}

\clearpage
\section*{Executive Summary}

\begin{itemize}
    \item \textbf{VLA Unified Framework and Action Token Taxonomy}. Current VLA models can be unified under a single framework: vision and language inputs are processed by a series of VLA modules to produce a chain of action tokens that progressively encode more grounded and actionable information, ultimately generating executable actions. Core to this framework, action tokens can be categorized into language description, code, affordance, trajectory, goal state, latent representation, raw action, and reasoning. Action tokens in VLAs are generalized counterparts to language tokens in LLMs.
    \vspace{5pt}
    \item \textbf{Action Token Trends}. The future of VLA models lies not in a single dominant action token, but in their strategic synthesis. Language motion, limited in expressiveness, is unlikely to become mainstream, while language plans remain essential for task decomposition. Code is a powerful alternative whose potential will be unlocked by building comprehensive function libraries that integrate perception and action primitives to solve complex, long-horizon tasks. A key synergy is forming between affordances that provide semantic what-to-do guidance and trajectories that define precise how-to-do paths. This pairing is powerfully supported by world models, which can predict visual goal states to ground the generation of both token types. Latent representations are promising but face training challenges. Raw actions represent the ideal for end-to-end learning but remain limited by data availability. Finally, reasoning serves as a meta-token to enhance all others, evolving from purely language-based to action-token-based reasoning with multimodal feedback and adaptive test-time computation.
    \vspace{5pt}
    \item \textbf{Emerging Action Token Types}. Action token types are shaped by foundation model capabilities. Stronger models and new modalities (e.g., audio, tactile) will give rise to new token types and subtypes.
    \vspace{5pt}
    \item \textbf{VLA Architecture Trends}. Effective VLA models are likely to adopt a hierarchical architecture, with the top layer using language description and code to perform long-horizon planning and logic control. In the near term, the lower layers are expected to closely integrate video prediction of goal state, flow modeling of trajectory, and 3D interaction prediction of affordance to form intermediate motion representations, which are finally mapped to raw actions. In the long term, the lower layers evolve toward a fully end-to-end approach, directly predicting raw actions from subtask-level inputs. Reasoning is always integrated throughout VLA models as needed.
    \vspace{5pt}
    \item \textbf{From Imitation to Reinforcement Learning}. By incorporating reinforcement learning, VLA models can overcome the limitations of imitation learning and achieve more human-like trial-and-error and self-guided exploration. However, real-world deployment requires more efficient RL algorithms to address high reset costs and low interaction efficiency. Additionally, VLMs can automate the generation of dense reward functions, accelerating model training and deployment.
    \vspace{5pt}
    \item \textbf{From VLA Models to VLA Agents}. A conscious effort should be made to evolve from VLA models to VLA agents, which are proactive systems that enhance perception-action capability with a broader cognitive architecture of memory, exploration, planning, and reflection. This shift also entails transitioning from the current linear processing architecture to more complex, bidirectional, and graph-structured topologies.
    \vspace{5pt}
    \item \textbf{The Triad of Progress: Model, Data, and Hardware}. Embodied AI aims to handle the unstructured, open-ended nature of the physical world—an ambition that demands synergy among models, data, and hardware. Despite this, progress is largely limited by constrained robotic platforms and scarce high-quality embodied data, forcing most research into simplified lab settings far from real-world complexity. As a result, the field remains in its infancy. Achieving robust, general-purpose intelligence requires the co-evolution of model, data, and hardware, advancing in tandem rather than in isolation.
    \vspace{5pt}
    \item \textbf{Safety and Alignment}. Current VLA research primarily focuses on model capability. Future work must place greater emphasis on ensuring safety and human alignment.
\end{itemize}

\clearpage
\begingroup
\hypersetup{linkcolor=black}
\renewcommand{\contentsname}{\centering \large \textbf{Table of Contents}}
\tableofcontents
\thispagestyle{empty}
\endgroup
\clearpage

\section{Introduction}

In recent years, Artificial Intelligence (AI) has made remarkable strides toward general-purpose intelligence. Central to this progress is the emergence of foundation models~\cite{bommasani2021opportunities, zhou2024comprehensive}---large neural networks trained on internet-scale data, which acquire broad and transferable capabilities by capturing the diverse knowledge and patterns embedded in their training corpora. 
As a prominent example, Large Language Models (LLMs), such as GPT-4~\cite{achiam2023gpt} and DeepSeek-R1~\cite{guo2025deepseek}, excel at natural language understanding, reasoning, and generation, forming the backbone of many text-based applications. In parallel, Vision Foundation Models (VFMs), such as CLIP~\cite{radford2021learning}, DINO~\cite{caron2021emerging, oquab2023dinov2}, and SAM~\cite{kirillov2023segment, ravi2024sam}, have shown strong generalization across a wide range of vision tasks. Building upon these, Vision-Language Models (VLMs), exemplified by GPT-4o~\cite{openai2024gpt4ocard}, Gemini 2.5 Pro~\cite{gemini25}, and Qwen2.5-VL~\cite{bai2025qwen2}, integrate visual and textual modalities to enable multimodal processing and generation. 
Collectively, these models encode vast world knowledge, exhibit strong performance on complex tasks, and generalize to novel scenarios---making them highly versatile and broadly applicable across domains.

However, despite their impressive capabilities, these models remain confined to the \emph{digital} world, limiting the impact on real-world tasks. To overcome this boundary, researchers have begun exploring ways to harness the perceptual and cognitive capabilities of foundation models to enhance task execution, thereby extending their intelligence into the \emph{physical} world. This line of work has led to the emergence of \textbf{Vision-Language-Action} (VLA) models, which we formally define as \textit{models that generate actions conditioned on visual and linguistic inputs, and are built upon at least one large-scale vision or language foundation model}. For example, SayCan~\cite{saycan2022arxiv}, PaLM-E~\cite{driess2023palme}, and Code as Policies~\cite{liang2023code} utilize the language and code generation abilities of LLMs and VLMs to produce high-level action plans expressed in natural language or executable code, which are then interpreted and executed by low-level controllers. Other works focus on extracting actionable knowledge from foundation models, such as generating affordances for task-relevant objects~\cite{huang2023voxposer} or predicting scene-level trajectories to guide downstream control~\cite{gu2023rt}. A separate line of research purposefully constructs latent representations of embodied action sequences via dedicated pretraining, and adapts VLMs to predict these representations, which are subsequently decoded and executed by a policy controller~\cite{bu2025agibot}. In addition, parallel efforts have sought to extend the scaling laws~\cite{hoffmann2022training, kaplan2020scaling} observed in vision and language domains to the embodied setting, collecting large-scale embodied datasets and training generalist agents end-to-end on top of vision-language foundation models~\cite{kim2024openvla, black2410pi0, liu2024rdt}. These diverse approaches have led to a rapid proliferation of VLA models in robotic manipulation~\cite{shi2025hi, team2025gemini}, navigation~\cite{zhou2024navgpt, serpiva2025racevla}, and autonomous driving~\cite{covla_wacv2025, hwang2024emma, tian2024drivevlm}, 
demonstrating promising capabilities in multitask learning~\cite{rth2024arxiv}, long-horizon task completion~\cite{black2410pi0}, and strong generalization~\cite{zhong2025dexgraspvla}. By leveraging foundation model intelligence, they offer new directions for addressing long-standing challenges in embodied AI, such as data scarcity and poor cross-embodiment transferability, and pave the way for \emph{agents capable of solving open-ended tasks expressed via open-vocabulary instructions in open-world physical environments}.

\begin{figure*}
    \vspace{-1em}
    \centering
    \includegraphics[height=0.95\textheight, keepaspectratio]{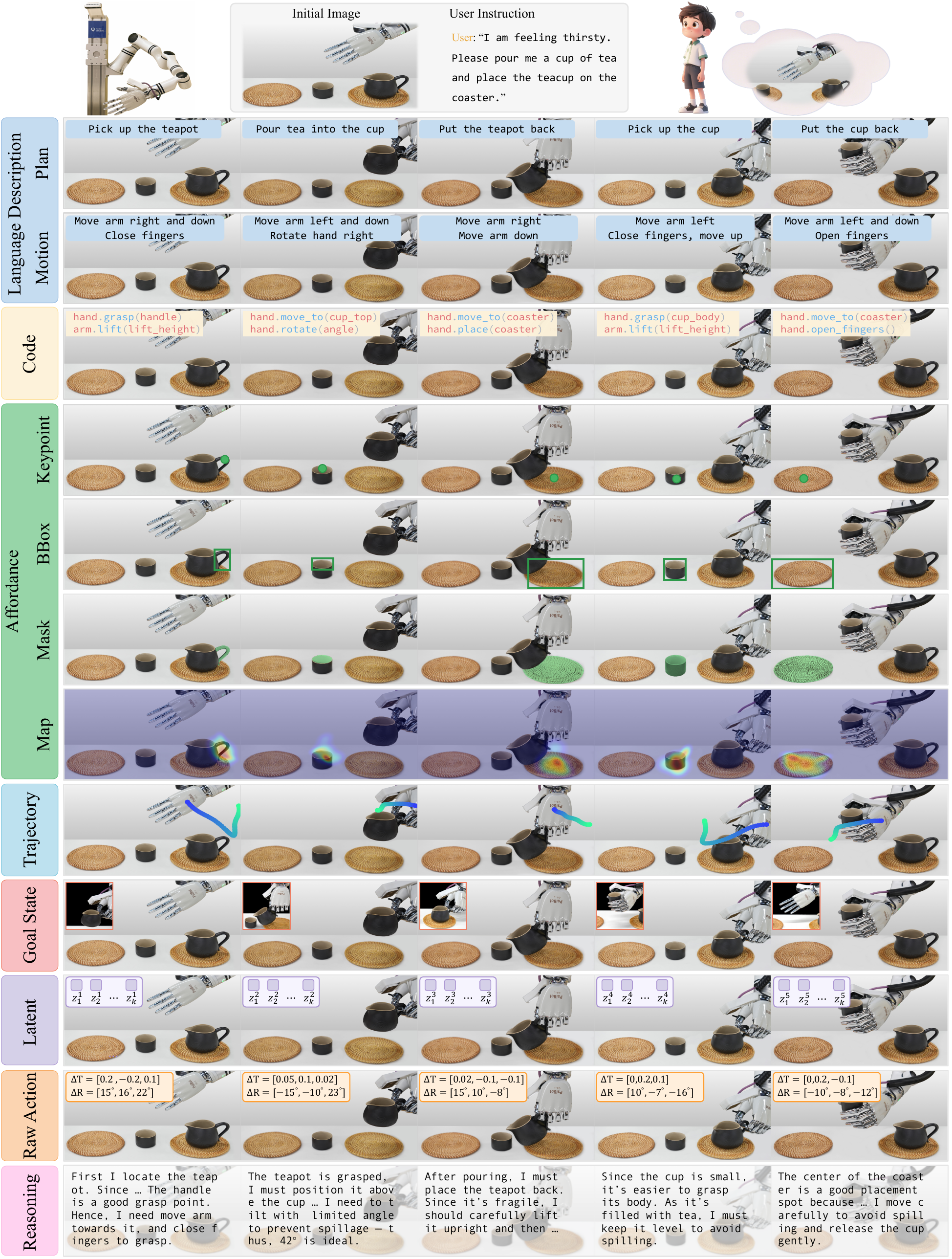}
    \caption{\textbf{Visualization of action tokens in a single embodied task.} Given the same vision and language inputs, different VLA models encode them into diverse action tokens, each conveying varying forms of actionable guidance and requiring distinct strategies for token generation and post-processing.}
    \label{fig:action_token}
\end{figure*}

The rapid progress, promising empirical results, and growing diversity of VLA models create an urgent need for a timely and systematic review to inform and guide future research. This need is further underscored by the underlying commonalities across seemingly disparate architectures. We observe that existing VLA models can generally be abstracted into a unified framework: vision and language inputs are iteratively processed through a sequence of \textbf{VLA modules}, producing a chain of \textbf{action tokens} that gradually encode increasingly informative and actionable guidance, ultimately producing executable actions. Formally, we define \textbf{VLA modules} as maximal differentiable subnetworks in VLA models that support end-to-end gradient flow, or non-differentiable functional units such as motion planning. If multiple neural components are connected and jointly optimized, they are regarded as parts of the same module. Following the naming convention of language and image tokens in VLMs, we refer to the outputs of VLA modules as \textbf{action tokens}. Additionally, we also consider semantically meaningful intermediate representations \emph{within} VLA modules---such as latent representations constructed via dedicated pretraining~\cite{bu2025agibot} and goal images~\cite{zhao2025cot}---as action tokens. \Cref{fig:unified_framework} illustrates the instantiations of VLA modules and action tokens in several representative VLAs, highlighting how they can be uniformly viewed, explained, and understood with our proposed framework. From this perspective, VLA models are primarily distinguished by how action tokens are formulated and organized. These tokens can be categorized into eight types: language description~\cite{shi2025hi, rth2024arxiv}, code~\cite{liang2023code, Singh2022ProgPromptGS}, affordance~\cite{huang2023voxposer, huang2024rekepspatiotemporalreasoningrelational}, trajectory~\cite{rt-trajectory, wen2023any}, goal state~\cite{zhen20243d, zhao2025cot}, latent representation~\cite{ye2024latent, bu2025agibot}, raw action~\cite{kim2024openvla, black2410pi0}, and reasoning~\cite{ECoT, clark2025action}.
\Cref{fig:action_token} visualizes their common forms using a single embodied task. Crucially, the design of action tokens shapes nearly every aspect of VLA models, including the choice of foundation model, data requirements, training and inference efficiency, interpretability, scalability, and applicability across tasks and environments. As such, action tokenization is central to the design of VLA models and necessitates a thorough understanding.

Despite its importance, the research community currently lacks a systematic and in-depth understanding of action tokenization. This survey aims to fill that gap by providing a structured overview of VLA research from the perspective of action tokenization. We begin by reviewing the evolution of vision and language foundation models, examining their design choices, scaling strategies, and capabilities. We then discuss the transition to embodied AI, in particular VLA models, and establish VLA as the next frontier (\Cref{sec:evolution-foundation-models}). With this background, we start introducing the VLA research landscape by presenting an overview of action tokens, including their taxonomy, definitions, comparisons, and organizational patterns within VLA models (\Cref{sec:overview-of-action-tokens}). Subsequent sections delve into each major category of action tokens, analyzing their motivations, representative approaches, properties, advantages, limitations, and future work (\Cref{sec:language-description,sec:code,sec:affordance,sec:trajectory,sec:goal-state,sec:latent,sec:raw-action,sec:reasoning}). 
We also summarize scalable data sources to inform and support future research (\Cref{sec:data}). Finally, based on the surveyed landscape and emerging trends, we outline future research directions for advancing the field of VLA (\Cref{sec:discussion-future-direction}). Through this perspective, we hope to offer valuable insights and actionable guidance for the development of next-generation embodied AI systems.

\section{The Evolution of Language and Vision Foundation Models}
\label{sec:evolution-foundation-models}

\begin{figure*}
    \vspace{-1.0\baselineskip}
    \centering
    \includegraphics[width=0.97\linewidth, keepaspectratio]{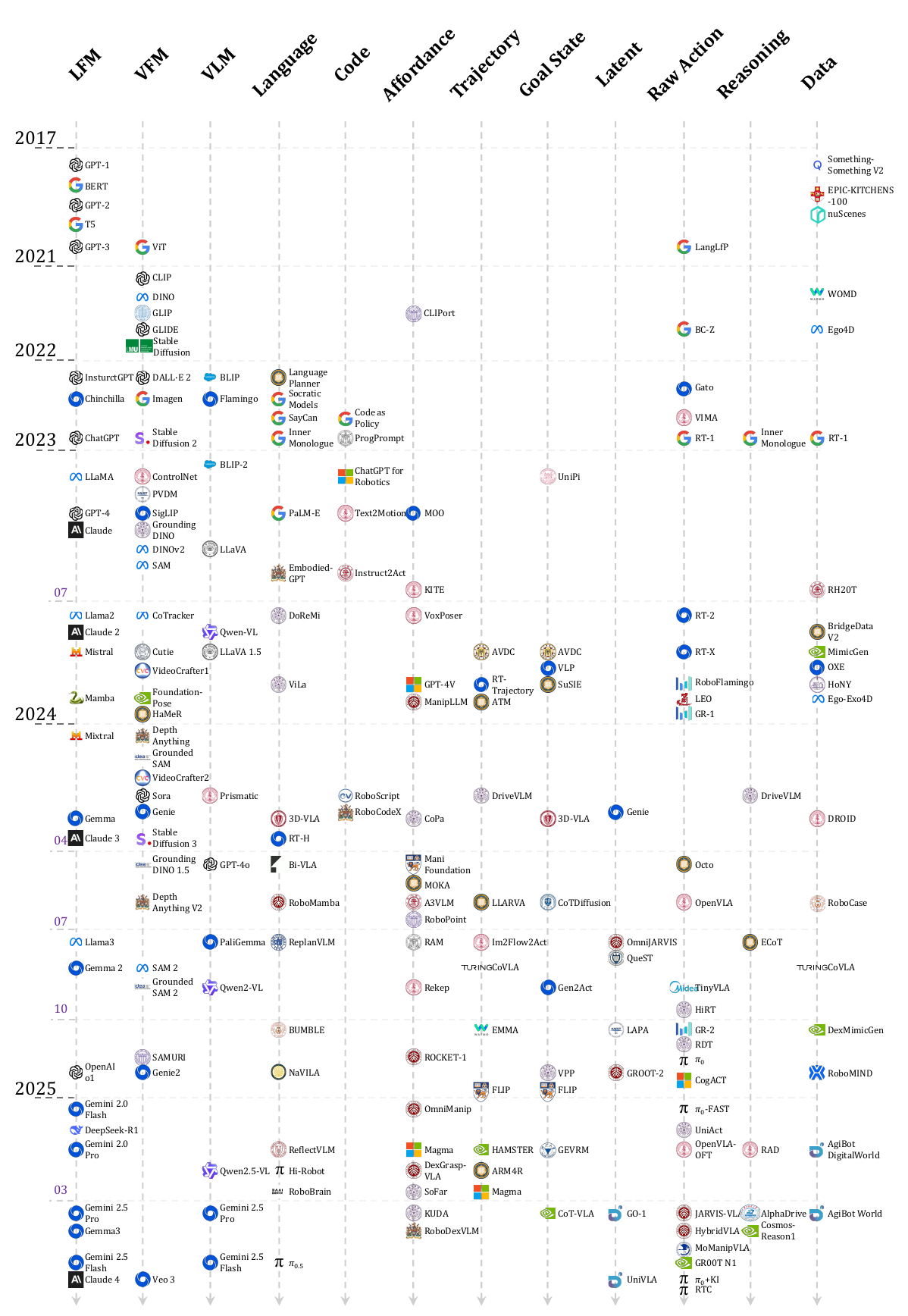}
    \caption{Evolution timeline of foundation models, VLA models, and data sources. The U-shape reflects how the growing proliferation of VLA is supported by progress in foundation models and data.}
    \label{fig:timeline}
\end{figure*}

This section first reviews the major advances in the evolution of language foundation models (LFMs, \Cref{sec:lfms}), vision foundation models (VFMs, \Cref{sec:vfms}), and vision-language models (VLMs, \Cref{sec:vlms}), elucidating their progress in terms of capabilities, technical innovations, and methodological approaches. Subsequently, we discuss the field's progression towards embodied AI, analyzing the significantly greater complexity of this domain and establishing embodied VLA as the next frontier.

\subsection{Language Foundation Models}
\label{sec:lfms}

The emergence of language foundation models can be largely traced back to the introduction of the Transformer architecture~\cite{vaswani2017attention}, which leverages multi-head self-attention and cross-attention mechanisms for scalable sequence modeling, and adopts an encoder-decoder structure for effective sequence-to-sequence generation. Building on this architecture, BERT~\cite{devlin2019bert} pretrains a bidirectional Transformer encoder in a self-supervised manner using masked language modeling and next sentence prediction objectives on large-scale unlabeled corpora, enabling the model to learn rich, context-aware representations that significantly improve downstream task performance. The Universal Sentence Encoders~\cite{cer2018universal, yang2019multilingual} similarly employ Transformer encoders to learn transferable sentence-level encodings. T5~\cite{raffel2020exploring} retains the encoder-decoder structure, reformulates all natural language processing tasks into a unified text-to-text format, and pretrains on the large-scale C4 dataset. Its pretrained encoder is widely used to produce high-quality language encoding for open-vocabulary inputs~\cite{liu2024rdt}. 

In contrast, GPT models~\cite{radford2018improving, radford2019language, brown2020language} formulate all NLP tasks as next-token prediction, motivating the use of decoder-only Transformer architectures, also referred to as causal or autoregressive Transformers. By scaling model size to 175 billion parameters and pretraining on internet-scale corpora, GPT-3~\cite{brown2020language} demonstrates impressive capabilities in language understanding and generation. More notably, it exhibits emergent behaviors such as in-context learning, where the model can perform tasks based solely on a few examples provided at inference time. This demonstrates that the scalability of model architecture, training objectives, and data sources enables learning to be effectively applied at scale, resulting in general-purpose models that outperform task-specific systems. This paradigm shift aligns with the core insight of the \textit{Bitter Lesson}~\cite{sutton2019bitter} and marks the beginning of the large language model (LLM) era.

To guide the efficient scaling of LLMs, scaling laws have been proposed to characterize predictable relationships between model size, data volume, compute requirements, and pretraining loss~\cite{kaplan2020scaling, hoffmann2022training}. These insights inform practical decisions about model design and resource allocation during large-scale training. InstructGPT~\cite{ouyang2022training} further advances the alignment of LLMs with human intent by applying supervised fine-tuning (SFT) on instruction-following datasets, followed by reinforcement learning from human feedback (RLHF). Since then, alignment techniques have been extensively studied to ensure that large AI models behave in accordance with safety considerations, human preferences, and values~\cite{ji2023ai}.

These technological advances have led to the development of highly capable commercial LLMs, such as GPT-4~\cite{achiam2023gpt} and Claude, which demonstrate strong performance in open-ended dialogue, code generation, and chain-of-thought reasoning~\cite{wei2022chain}. Orchestrated into an evolutionary coding agent called AlphaEvolve~\cite{novikov2025alphaevolve}, Gemini 2.0 Flash and Gemini 2.0 Pro have jointly enabled remarkable breakthroughs in open scientific problems, including matrix multiplication. However, due to their closed-source nature and restricted API-based access, these models are difficult to inspect, fine-tune, or integrate into broader research and development workflows.

To address these limitations, a number of open-source LLMs, such as Llama~\cite{touvron2023llama, touvron2023llama2, grattafiori2024llama}, Gemma~\cite{team2024gemma, team2024gemma2, team2025gemma}, and Mistral~\cite{jiang2023mistral7b}, have been released, with model sizes ranging from 2B to 70B parameters to accommodate varying requirements. Built on top of these models, parameter-efficient fine-tuning (PEFT) techniques~\cite{li2021prefix, liu2024gpt, liu2021p, lester2021power, houlsby2019parameter, dettmers2023qlora}, such as LoRA~\cite{hu2022lora}, enable task-specific adaptation using significantly fewer trainable parameters and lower computational cost, making fine-tuning feasible in resource-constrained settings.

To further scale model capacity without proportional increases in computation, Mixture-of-Experts (MoE) architectures have been introduced into LLMs, as exemplified by Switch Transformer~\cite{fedus2022switch} and Mixtral~\cite{jiang2024mixtral}. MoE models activate only a subset of expert sub-networks for each input, allowing for significantly larger effective model capacity while maintaining efficient inference.

Meanwhile, to address the quadratic time complexity of the Transformer architecture, alternative designs such as Mamba~\cite{gumamba} have been proposed. Mamba replaces self-attention with selective state space updates, enabling linear-time sequence modeling while preserving strong performance across long contexts.

Another line of work improves reasoning capabilities by scaling test-time computation. For example, OpenAI o1~\cite{jaech2024openai} and DeepSeek-R1~\cite{guo2025deepseek} dynamically allocate computational resources during inference to enhance performance on complex reasoning tasks. In particular, DeepSeek-R1 acquires this capability through large-scale reinforcement learning based on GRPO~\cite{shao2024deepseekmath}.

Finally, significant progress has been made in optimizing the infrastructure for training and deploying LLMs. A range of parallelism strategies, including data parallelism~\cite{LiAPSAJLSS14}, model parallelism~\cite{dean2012large}, pipeline parallelism~\cite{narayanan2019pipedream}, and tensor parallelism~\cite{shoeybi2019megatron}, are actively used to scale training across distributed compute environments. Additionally, inference acceleration techniques such as model quantization, weight pruning, and speculative decoding have been developed to reduce latency and computational overhead during deployment.

These advancements have made LLMs highly capable in knowledge, dialogue, code, and reasoning, while also enabling efficient training, deployment, and fine-tuning through mature infrastructure. They not only improve the usability of LLMs, but also support the development of vision and multimodal systems, forming key building blocks for embodied VLA models.

\subsection{Vision Foundation Models}
\label{sec:vfms}

Following the success of Transformer in the language domain, the computer vision community has begun to replace convolutional neural networks~\cite{lecun1989backpropagation,krizhevsky2012imagenet,he2016deep} with Vision Transformer (ViT)~\cite{dosovitskiy2020image} as the default backbone of vision models to attain better performance when trained with large-scale datasets. This architectural shift naturally treats images as sequences of visual tokens, a representational format that allows visual inputs to be handled similarly or jointly with textual inputs, facilitating cross-modal alignment and fusion in subsequent multimodal models. 
Additionally, the scalability of LLM training has also inspired researchers to explore scalable learning objectives in visual learning, in order to train general models on internet-scale visual data without human-annotated labels. As an early and successful attempt, CLIP~\cite{radford2021learning} utilizes natural language supervision for image representation learning by training on 400 million image-text pairs with a contrastive loss. This enables CLIP to learn robust and generalizable image representations and show impressive zero-shot transfer capabilities. SigLIP~\cite{zhai2023sigmoid} improves upon CLIP by replacing the original softmax operations with a sigmoid loss, boosting training efficiency and enhancing performance. Both CLIP and SigLIP have been widely employed as image encoders~\cite{liu2023visual,karamcheti2024prismatic}, especially in scenarios requiring multimodal understanding, due to their joint training with textual supervision. However, relying on textual supervision also constitutes their limitations. Since textual descriptions are often high-level and abstract, the encoded image features of CLIP and SigLIP could lack complex pixel-level information, which is undesirable for tasks requiring detailed visual understanding. To tackle this, DINO~\cite{caron2021emerging,oquab2023dinov2} directly learns from curated image datasets in a self-supervised manner, obtaining rich, general-purpose visual features helpful for fine-grained downstream tasks such as semantic segmentation and depth estimation. Importantly, its encoded features could match similar regions across different objects, such as the wings of a plane and a bird, showing in-depth semantic understanding and world knowledge.~\citet{darcet2024vision} proposes a simple yet effective improvement to these ViT-based models by adding learnable register tokens along with the original \texttt{[CLS]} token and patch tokens to remove artifacts otherwise present in feature maps and increase performance on dense prediction tasks.

Based on these pioneering image encoding efforts, subsequent research has developed foundation models tailored for specialized downstream vision tasks. Depth Anything~\cite{yang2024depth} effectively utilizes self-generated pseudo-labels from large-scale unlabeled data for robust monocular depth estimation (MDE), while Depth Anything V2~\cite{yang2024depthv2} leverages ground-truth labels from synthetic data to enhance fine-grained detail preservation. The Segment Anything Model (SAM)~\cite{kirillov2023segment} serves as a foundation model for promptable image segmentation, and its successor SAM 2~\cite{ravi2024sam} extends this capability to the video domain. These models can generate valid segmentation masks based on prompts in the form of points, bounding boxes, masks, and---in the case of SAM---text. Cutie~\cite{cheng2024putting} is an earlier model for video object segmentation (VOS) and has demonstrated robustness under diverse visual conditions~\cite{zhong2025dexgraspvla}. 
SAMURAI~\cite{yang2024samurai} improves the visual object tracking (VOT) performance of SAM 2 by incorporating motion modeling and motion-aware memory selection, enabling more effective handling of fast motion, occlusion, and crowded scenes. CoTracker~\cite{karaev2024cotracker} complements this line of work by introducing a transformer architecture for dense point tracking in long video sequences.

In the field of open-vocabulary detection and grounding, a series of models have progressively advanced region-level vision-language understanding. GLIP~\cite{li2022grounded} unifies detection and phrase grounding within a single pretraining framework by extending CLIP-style alignment to the region level. Grounding DINO~\cite{liu2024grounding} builds upon this with a DETR-style architecture and contrastive region-text alignment, achieving strong performance on open-vocabulary grounding tasks. Grounding DINO 1.5~\cite{ren2024grounding} scales model size and training data, improving generalization and setting new state-of-the-art results. Grounded SAM~\cite{ren2024grounded} further combines Grounding DINO with SAM to enable zero-shot language-driven segmentation. Grounded SAM 2 extends it towards grounding and track anything in videos.

For high-fidelity image and video generation, diffusion models~\cite{ho2020denoising, song2020denoising, song2020score} have become the dominant approach. Early models like GLIDE~\cite{nichol2022glide}, DALL$\cdot$E 2~\cite{ramesh2022hierarchical}, and Imagen~\cite{saharia2022photorealistic} demonstrate the power of text-guided image synthesis, while Stable Diffusion~\cite{rombach2022high, esser2024scaling} enables efficient, open-domain generation with wide adoption. ControlNet~\cite{zhang2023adding} introduces spatial conditioning to support fine-grained control over structure and layout. For video, models such as VideoCrafter~\cite{chen2023videocrafter1, chen2024videocrafter2} and PVDM~\cite{yu2023video} extend diffusion to the temporal domain for text-to-video synthesis. Sora~\cite{videoworldsimulators2024} advances this further by employing flow-matching~\cite{lipmanflow,lipman2024flow} and learns physical priors to generate long-duration, high-resolution videos with strong temporal coherence. More recently, Veo 3 showcases impressive full-modality generation, including synchronized audio and motion, pushing the boundaries of realistic video synthesis. These advanced image and video generative models are also referred to as world models, since they encode vast physical common sense and world knowledge. In parallel, other world models such as Genie~\cite{bruce2024genie} and Genie 2~\cite{parkerholder2024genie2} simulate future visual dynamics conditioned on action sequences, enabling accurate and coherent rollout of environment evolution across time.

Other efforts focus on developing foundation models for manipulation-relevant perception tasks. FoundationPose~\cite{wen2024foundationpose} is a unified vision foundation model for robust and generalizable 6D pose estimation and tracking of novel objects, regardless of the availability of CAD models. HaMeR~\cite{pavlakos2024reconstructing} leverages large-scale data and high-capacity transformer architectures to enable accurate and reliable hand mesh recovery from monocular input, facilitating hand pose extraction from human videos and supporting dexterous manipulation tasks. 

These advances in vision foundation models have provided general-purpose solutions for visual representation learning, vision-language alignment, common vision tasks, and generative modeling. Their capabilities in generalized visual understanding and generation have substantially accelerated the progress of multimodal learning and empowered a broad spectrum of real-world applications.

\subsection{Vision-Language Models}
\label{sec:vlms}

The advancement of vision and language foundation models has naturally driven research toward multimodal understanding, reasoning, and generation, resulting in the rise of vision-language models. As an early effort, BLIP~\cite{li2022blip} introduces a multimodal mixture of encoder-decoder (MED) architecture based on ViT and BERT for unified vision-language understanding and generation, along with a data bootstrapping strategy that synthesizes captions and filters noisy web data into high-quality image-text pairs. To better harness readily-available unimodal foundation models, BLIP-2~\cite{li2023blip} proposes a Q-Former connector and a two-stage training strategy to effectively align frozen pretrained image encoders with frozen LLMs, achieving strong vision-language performance with modest trainable parameters.

Different architectural paradigms were also explored. Flamingo~\cite{alayrac2022flamingo}, for instance, employs a Perceiver Resampler and gated cross-attention layers for cross-modal alignment. It also processes inputs in a manner inherently compatible with interleaved visual and textual sequences, thereby enabling strong few-shot learning capabilities. LLaVA~\cite{liu2023visual} represents a milestone in the development of VLM architecture, which simply links a CLIP vision encoder to the Vicuna LLM~\cite{chiang2023vicuna} via a linear projection and is trained on visual instruction-tuning data synthesized by GPT-4. LLaVA-1.5~\cite{liu2024improved} improves upon LLaVA by adopting a stronger vision encoder, replacing the linear projection with an MLP, and training on a larger dataset.

The Qwen-VL family represents another prominent line of work. The initial Qwen-VL~\cite{bai2023qwenvlversatilevisionlanguagemodel} combines the Qwen-7B LLM~\cite{qwen} with a ViT through a position-aware cross-attention adaptor. Its specially designed input-output interface for images and bounding boxes, together with a three-stage training strategy, enables interleaved image-text understanding and visual grounding capabilities. Its successor, Qwen2-VL~\cite{wang2024qwen2}, enhances spatial-temporal encoding with 2D RoPE and M-RoPE to support images and videos of varying resolutions and aspect ratios. It demonstrates strong multilingual capabilities and competitive performance on vision-language tasks such as captioning, VQA, and video understanding. Most recently, Qwen2.5-VL~\cite{bai2025qwen2} extends dynamic resolution to the temporal domain and aligns M-RoPE time IDs with absolute time, enabling more refined temporal understanding. It incorporates window attention in the vision encoder to improve inference efficiency. Together with extensive high-quality data curation, Qwen2.5-VL delivers enhanced visual recognition, precise object grounding, robust document parsing, and long-video comprehension.

\citet{karamcheti2024prismatic} explore key VLM design decisions around image preprocessing, architecture, and optimization, concluding that single-stage training, fused DINOv2 and SigLIP vision backbones, base LLMs, and co-training with language-only data are effective strategies. Building on these insights, they develop Prismatic VLMs, which consistently outperform LLaVA-1.5 across benchmarks and have been used later in OpenVLA~\cite{kim2024openvla}. PaliGemma~\cite{beyer2024paligemma}, a 3B VLM built on SigLIP~\cite{zhai2023sigmoid} So400m~\cite{alabdulmohsin2023getting} and Gemma 2B~\cite{team2024gemma}, is developed with a focus on transferability and subsequently adopted as the backbone for the $\pi_0$ series of VLA models~\cite{black2410pi0,Pi0-FAST,shi2025hi,intelligence2025pi05visionlanguageactionmodelopenworld,driess2025knowledge}. 

At the forefront of current capabilities are two proprietary models: GPT-4o~\cite{openai2024gpt4ocard} and Gemini 2.5 Pro~\cite{gemini25}. Both exhibit leading performance on general vision-language benchmarks and have seen widespread adoption in real-world applications. GPT-4o is distinguished by its native support for image generation, while Gemini 2.5 Pro is recognized for its powerful reasoning capabilities, underscoring the rapid progress and practical utility of modern VLMs.

\subsection{Embodied VLA Models as the Next Frontier}
\label{sec:next-frontier}

The rapid advancements in foundation models are increasingly fueling imagination and propelling the pursuit of Artificial General Intelligence (AGI). As current foundation models primarily operate within the \emph{digital} domain, representing \emph{digital AI}, researchers are naturally shifting their focus to \emph{embodied AI}, which aims to develop general-purpose agents capable of following human instructions in the \emph{physical} world. However, we emphasize that embodied AI presents a significantly bolder ambition than digital AI for several reasons.

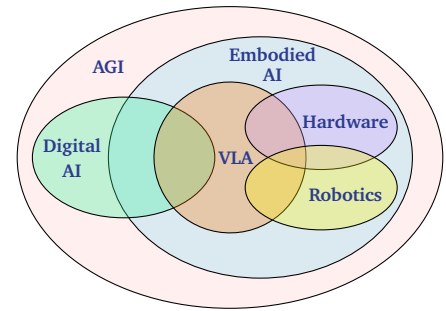
\begin{wrapfigure}{r}{0.36\textwidth}
\vspace{-10pt}
\centering
\input{figure/scope}
\caption{A Venn diagram showing the interrelationships among key AI fields. VLA models intersect with digital AI, hardware, and robotics, representing a core subfield of Embodied AI and a key area in the progression towards AGI.}
\label{fig:ai-fields}
\vspace{-10pt}
\end{wrapfigure}

Fundamentally, the problems that embodied AI must solve introduce novel forms of open-endedness and challenges absent in digital AI. Whereas difficult digital cases may involve out-of-distribution (OOD) or adversarial inputs, the physical world is inherently unstructured, and even routine settings can be highly challenging. Free-flowing human conversations, inadvertent interventions, fallen chairs, cluttered rooms, and occlusions are common examples, not to mention even more difficult situations. A comparable and perhaps more familiar problem, which we also consider part of embodied AI in this paper, is autonomous driving. While autonomous driving is already incredibly difficult, general-purpose embodied intelligence in the physical world must handle orders of magnitude more situations, leading to orders of magnitude greater challenges and difficulties. This imposes substantial demands on both model and data to support robust embodied AI.

Furthermore, a crucial realization is that embodied AI also involves requirements for robot hardware, which digital AI does not entail. To achieve general-purpose embodied intelligence, the hardware platform must possess the dexterity and robustness necessary for general tasks, a level that is currently far from being met. Representative gaps include dexterous hands and robotic arms that are far from achieving human-level dexterity, heavy reliance on grippers, diversity and isolation of embodiments, and the lack of sensitive, full-coverage tactile sensors. Since hardware perfection cannot be achieved in a short time frame, a reasonable expectation is that models, data, and hardware will develop synergistically, ultimately achieving general intelligence. The scope of this survey primarily focuses on the model and data aspects, but we also inform readers of the hardware challenges, which are often important considerations for model development. 

Given embodied AI's requirement for general visual and language capabilities, a natural strategy is to build on foundation models and endow them with action capabilities. This direction has given rise to embodied VLA models, now a central topic of investigation. Situated at the intersection of digital AI, robotics, and hardware, VLA constitutes a core subfield of embodied AI and a key area in the pursuit of AGI (\Cref{fig:ai-fields}). The hundreds of VLA papers proposed to date illustrate a rapidly expanding field (\Cref{fig:timeline}), showing early yet limited signs of intelligence and generalization. This survey systematically reviews and analyzes these papers from the perspective of action tokenization to outline the research landscape. Notwithstanding recent progress, most evaluations remain confined to simplified laboratory settings---predominantly gripper-based manipulation---and thus far from the requirements for general-purpose embodied agents in everyday environments. Consequently, the field is in its infancy, and substantial advances are still needed. The continued development of embodied VLA models is therefore poised to remain the next frontier of research for the foreseeable future.

\section{Overview of Action Tokens}
\label{sec:overview-of-action-tokens}

Research in VLA models focuses on processing vision and language input to generate action output, leveraging foundation models. We observe that in designing VLA architectures and formulating training strategies, the concepts of VLA modules and action tokens naturally emerge. To map raw perception to action, VLA models must effectively comprehend the scene and instruction, ground the instruction within the scene, plan the current subtask, anticipate subsequent movement, and generate executable actions. The complexity and generality of embodied tasks further necessitate the switching, repetition, and recursion of these capabilities. To facilitate task-relevant information flow and refinement, VLAs delegate these capabilities to distinct modules, manage their respective generations, and logically link these modules and their generations to derive final actions. Consequently, the design of generation formats and the training strategies for these modules are central to VLAs. This survey reviews existing research from this perspective.

We term the maximal differentiable subnetworks and non-differentiable functional units within a VLA ``VLA modules'', and their generations ``action tokens''. Furthermore, semantically meaningful intermediate generations within VLA modules are also considered ``action tokens''. The designation ``action token'' not only signifies that these generations encapsulate action-related information but also aligns with the naming convention of ``language token'' in LLMs. Indeed, \emph{action tokens in VLAs are generalized counterparts to language tokens in LLMs.}

To further clarify these concepts, \Cref{fig:unified_framework} highlights several representative examples. For a given language instruction in the current environment, Hi Robot~\cite{shi2025hi} employs a fine-tuned PaliGemma model to predict the next subtask in natural language. This is followed by a VLA model—trained in a manner similar to $\pi_0$~\cite{black2410pi0}—that generates low-level robot commands. In this case, both the fine-tuned PaliGemma and the customized $\pi_0$ constitute VLA modules, while the intermediate language plan and the resulting raw actions serve as action tokens. Another example is VoxPoser~\cite{huang2023voxposer}, which also begins by using LLMs to decompose a language instruction into subtasks. It then employs LLMs and VLMs to generate an affordance map for solving each subtask based on the current scene, and finally invokes a motion planning module to convert the affordance map into raw actions. Here, the LLMs, VLMs, and motion planning algorithm all function as VLA modules, while the language plan, affordance map, and raw actions represent the corresponding action tokens.

Other VLA models can similarly be analyzed by identifying their constituent VLA modules and action tokens according to this framework. Based on a broad survey of existing literature, we observe that most VLA models conform to a unified abstract framework, as illustrated in \Cref{fig:unified_framework}: vision and language inputs are iteratively processed by a sequence of VLA modules to produce a chain of action tokens that progressively encode more grounded and actionable guidance, ultimately resulting in executable actions. This abstraction offers a unified lens through which to interpret and compare diverse VLA architectures.

\input{table/action_token}

As VLA leverages foundation models for the development of VLA modules and action tokens, the inherent diversity in these underlying models results in a variety of action token formats. Existing VLA research has primarily investigated eight principal types of action tokens: language description, code, affordance, trajectory, goal state, latent representation, raw action, and reasoning. In \Cref{fig:action_token}, we visualize common formats of these action tokens, employing the illustrative task: ``prepare tea''. This visualization demonstrates that, for a given language instruction and observation, each type of action token encodes task-relevant guidance in a distinct manner. Formal definitions of these action tokens are provided below.

\vspace{0.5em}
\noindent\textbf{(1) Language description} (\Cref{sec:language-description}): A natural language expression that describes the intended action sequence, ranging from high-level and abstract language plan to low-level and concrete language motion.

\noindent\textbf{(2) Code} (\Cref{sec:code}): An executable code snippet or pseudocode that either constitutes a complete robot program or specifies low-level atomic operations.

\noindent\textbf{(3) Affordance} (\Cref{sec:affordance}): A spatially grounded representation that captures task-specific and interaction-relevant properties of objects, typically represented as keypoint, bounding box, segmentation mask, or affordance map.

\noindent\textbf{(4) Trajectory} (\Cref{sec:trajectory}): A temporally ordered sequence of spatial states that captures the dynamic evolution of an object, end-effector, or scene.

\noindent\textbf{(5) Goal state} (\Cref{sec:goal-state}): A predicted future observation---such as an image, point cloud, or video clip---that visually represents the expected outcome of the intended action sequence, serving as an intermediate target for planning and execution.

\noindent\textbf{(6) Latent representation} (\Cref{sec:latent}): A purposefully pretrained latent vector sequence that encodes action-relevant information over a temporal interval, typically extracted from large-scale datasets.

\noindent\textbf{(7) Raw action} (\Cref{sec:raw-action}): One or more low-level control commands that can be directly executed by a robot.

\noindent\textbf{(8) Reasoning} (\Cref{sec:reasoning}): Natural language expressions that explicitly describe the decision-making process leading to a specific action token.
\vspace{0.5em}

In the following sections, we systematically present VLA models categorized by each type of action token. For each category, we discuss the motivation for its adoption, review relevant literature, and analyze its advantages and limitations, while highlighting directions for future research. Each section also includes a table summarizing the surveyed works, examining similarities and differences across multiple dimensions pertinent to the respective action token. In particular, the ``previous module'' and ``next module'' columns refer to the design strategies of the VLA modules preceding and succeeding the action token, respectively, often reflecting key innovations and thoughtful design choices in how the token is generated and transformed to enable effective VLA models. Additionally, \Cref{tab:action_token} provides a summary of the most salient advantages, limitations, and notable empirical results for each type of action token, facilitating comparison, understanding, and insight across categories.

\section{Language Description as Action Tokens}
\label{sec:language-description}

\input{table/language}

The advancements of LLMs and VLMs naturally motivate the use of language description as action tokens in VLA models, enabling direct leverage of their strengths in language understanding, generation, reasoning, and planning. Moreover, representing actions through natural language aligns closely with the way humans conceptualize and communicate plans, especially for complex and long-horizon tasks.
Rather than executing primitive actions directly, humans tend to decompose a high-level instruction into intermediate, semantically meaningful sub-steps, and further into precise motion commands when necessary.
This hierarchical structure of tasks allows people to flexibly adapt their plans to different contexts and levels of control.
Inspired by this, these language-based tokens in VLA models are also designed with varying levels of abstraction, broadly categorized into two types.
At the upper end, \textbf{language plans}~\cite{huang2022inner, driess2023palme, saycan2022arxiv, ji2025robobrain, shi2025hi, huang2022language, huang2023voxposer} typically describe an entire subtask or a high-level goal in a single phrase.
Examples such as ``pick up the cup'' and ``place the cup on the table'' convey what the robot should accomplish, serving as semantic anchors that can be assigned to skills or policies.
In contrast, at a finer level, \textbf{language motions}~\cite{rth2024arxiv, cheng2024navila} specify low-level physical actions closer to motor control, using expressions such as ``move the arm forward'' and ``close gripper'', which detail the execution of specific movements.
This spectrum of abstraction provides a conceptual framework that enables VLA models to organize, interpret, and execute embodied tasks at different levels of granularity, with the potential to support more human-like hierarchical planning.
Motivated by these advantages, a growing body of work has explored the incorporation of language description as action tokens in VLA, leading to diverse strategies for task decomposition, action sequencing, and execution management. We list them in \Cref{tab:language_description}.

\subsection{Progress and Key Papers}

Early works, such as Language Planner~\cite{huang2022language}, Socratic Models~\cite{zeng2022socratic}, and SayCan~\cite{saycan2022arxiv}, demonstrate that LLMs can directly decompose high-level natural language instructions into semantically meaningful subgoals without task-specific training. This opens up the possibility of planning without domain-specific engineering. 
However, naive LLM-based planners face a fundamental limitation: the absence of perceptual grounding. Operating without direct access to visual, spatial, or sensory inputs, they struggle to align abstract plans with the actual state of the environment and to adapt effectively to unanticipated physical contexts.

To address this, these works introduce explicit grounding mechanisms. 
Socratic Models~\cite{zeng2022socratic} pairs LLMs with VLMs that detect relevant objects and provide visual context, bridging the gap between abstract plans and physical reality. 
SayCan~\cite{saycan2022arxiv} re-weights LLM-generated plans with an affordance function that estimates the feasibility of each action plan given the environment. 
Inner Monologue~\cite{huang2022inner} extends this further by introducing feedback loops: the system continuously prompts the LLM with signals like success detection, scene descriptions, or human feedback, enabling reflective, multi-turn reasoning and dynamic adjustment of plans. DoReMi~\cite{guo2024doremi} proposes a dual-role framework, where the LLM generated both high-level plans and explicit execution constraints. 
These constraints are monitored by VLM-based detectors at runtime, ensuring the system can react to dynamic contingencies.

Nevertheless, external grounding modules struggle to flexibly provide task-dependent information, cannot reason jointly with LLMs, and are often inadequate for handling fine-grained tasks in complex environments~\cite{hu2023look}. To address these challenges, subsequent works have shifted towards more natural grounding approaches by directly incorporating visual inputs into the planning process through VLMs.
PaLM-E~\cite{driess2023palme} is a large-scale embodied multimodal language model that unifies vision, language, and robot state information by encoding them into a single multimodal input. This design enables deep integration of perception, allowing the model to directly generate plans conditioned on sensory inputs.
To reduce training costs and enhance accessibility, EmbodiedGPT~\cite{mu2023embodiedgpt} adopts a lightweight VLM architecture composed of pretrained and frozen components, and trains it on a self-constructed EgoCOT dataset using a parameter-efficient strategy.
ViLa~\cite{hu2023look} utilizes GPT-4V as a planner, showing that advances in foundation models can directly translate to improvements in VLA models without additional task-specific training. 
3D-VLA~\cite{zhen20243d} and RoboMamba~\cite{RoboMamba} extend this paradigm by incorporating 3D scene understanding, spatial layouts, and visual affordance prediction into the planning loop.

Subsequent works have gone beyond grounding to tackle long-horizon, complex tasks by introducing memory and reflection mechanisms. 
BUMBLE~\cite{shah2024bumble} and ReflectVLM~\cite{feng2025reflective} incorporate these mechanisms, allowing systems to handle interdependent subtasks and plan across diverse and complex environments. 
These agent-like capabilities mark a shift from isolated planning to more integrated, adaptive behavior.

Previous papers constrain generated language plans within the scope of predefined skill sets~\cite{huang2022language, zeng2022socratic, saycan2022arxiv, huang2022inner} or scripted controllers~\cite{hu2023look, feng2025reflective}, limiting their flexibility in addressing complex instructions and open-ended scenarios. To overcome these limitations and enable handling of free-form prompts, research has increasingly focused on integrating more robust and generalizable low-level policies.
Notable among these are Hi Robot~\cite{shi2025hi} and $\pi_{0.5}$~\cite{intelligence2025pi05visionlanguageactionmodelopenworld}, which exemplify this transition.

Hi Robot~\cite{shi2025hi} proposes a hierarchical framework in which a high-level VLM interprets complex prompts and dynamic user feedback, producing free-form language commands executed by a low-level generalist control policy~\cite{black2410pi0}. The generality of both the high-level and low-level components enables the system to handle multi-stage tasks and situated corrections across diverse platforms.
Its successor $\pi_{0.5}$~\cite{intelligence2025pi05visionlanguageactionmodelopenworld}
unifies the planner and controller into a single VLA model, which first predicts high-level semantic subtasks and generates continuous low-level actions conditioned on these subtasks. By training on web-scale heterogeneous data, it can perform long-horizon, open-world tasks such as cleaning unseen kitchens with remarkable generalization.

While the aforementioned works largely focus on language plans at the subtask level, another line of research investigates language motions as fine-grained linguistic descriptions of low-level movements.
A representative work in this direction is RT-H~\cite{rth2024arxiv}, which introduces an intermediate layer of language motion between vision-language inputs and action outputs to facilitate multi-task data sharing across diverse high-level tasks. Building on this design, RT-H adopts a hierarchical architecture in which a VLM first predicts the current language motion (e.g. ``move arm forward'') conditioned on the instruction (e.g. ``pick coke can''), and subsequently generates the low-level action based on both the instruction and the predicted language motion. This approach improves performance and enables more effective intervention.
Beyond manipulation, NaVILA~\cite{cheng2024navila} applies this idea to navigation tasks by first generating mid-level spatial commands in natural language, such as “move forward 75 cm”, which are then executed by a visual locomotion policy.
These works collectively demonstrate that fine-grained language motions, by explicitly describing spatial and temporal micro-actions, can provide precise, interpretable guidance for low-level controllers.
A key advantage here is to enable better data sharing across different tasks at the language motion level, resulting in better language motion composition, generalization, and data efficiency. Also, the fine-grained language motions are more convenient for humans to correct in the context of current scenes.

\subsection{Advantages of Language Descriptions}

A primary advantage of using language descriptions as action tokens lies in their seamless integration with large \textbf{foundation models}. Both LLMs and VLMs possess strong out-of-the-box capabilities in understanding, reasoning, and planning, which enables zero-shot planning and significantly reduces the need for task-specific training. They can also directly benefit from ongoing advancements in in-context learning, memory, decoding strategies, and search techniques. Even when fine-tuning is required, the alignment between language descriptions and the model's native output space makes the process more efficient and less disruptive than with other forms of action tokens, which often suffer from greater modality mismatches.

Second, language description benefits from the abundance of \textbf{co-training data}. Empirical results in PaLM-E~\cite{driess2023palme} and $\pi_{0.5}$~\cite{intelligence2025pi05visionlanguageactionmodelopenworld} show that co-training on such data can transfer rich world knowledge into VLA models, thereby improving generalization.

Third, language description is particularly well-suited for long-horizon \textbf{planning}. In fact, language descriptions are almost \emph{necessary} if VLA models are to perform complex, temporally extended tasks.

Lastly, the \textbf{interpretability} of language descriptions facilitate human oversight and intervention, thereby enhancing safety, transparency, and controllability. Systems such as Hi Robot~\cite{shi2025hi} and YAY Robot~\cite{shi2024yell} exemplify how language-based plans enable seamless integration of human-in-the-loop corrections and dynamic feedback. Moreover, the correction data collected through online human interaction can be leveraged to iteratively improve model performance over time~\cite{rth2024arxiv}.

\subsection{Discussion and Future Directions}

One limitation of using language descriptions as action tokens arises from their imperfect \textbf{expressiveness}. While natural language is flexible and interpretable, it is inherently ambiguous and often insufficient for specifying fine-grained control behaviors --- particularly in contact-rich or deformable manipulation tasks~\cite{Lin_2023, du2023video}, where precise spatial and temporal details are critical. These issues may lead to miscommunication between system components and inadequate task grounding, both of which can hinder overall performance.

Another limitation concerns \textbf{latency}. Generating high-quality language descriptions often depends on large-scale models, which can incur inference delays and constrain applicability in dynamic or real-time scenarios. Potential remedies include employing inference acceleration techniques and developing asynchronous planning and execution frameworks.

Looking beyond these limitations, a promising research direction is to leverage language descriptions primarily for high-level planning - decomposing complex tasks into simpler subproblems that can then be more effectively addressed by VLA models utilizing alternative action token formats such as affordance (\Cref{sec:affordance}), trajectory (\Cref{sec:trajectory}), or goal state (\Cref{sec:goal-state}). These representations offer greater precision and efficiency for low-level execution, thereby enabling more reliable and scalable embodied intelligence.

\section{Code as Action Tokens}
\label{sec:code}

A key challenge in VLA models lies in planning and controlling complex, long-horizon manipulation tasks that require structured reasoning and adaptability to dynamic environments. Traditional action representations, such as discrete signals or direct language commands, often lack the expressiveness required for this complexity. In response, code-based action tokens emerge as a powerful alternative. These representations consist of executable code snippets or pseudocode that incorporate control structures like conditionals and loops. This format allows for direct execution through robot control APIs, enabling models to generate modular behaviors with explicit logic. It effectively supports both hierarchical planning and reactive control.

Code offers distinct advantages over other action formats. It provides clear logical structures and can leverage rich third-party libraries. Furthermore, it creates a transparent and verifiable bridge between high-level instructions and low-level robot primitives. Recent advances in LLMs have made it feasible to synthesize task-relevant codes from natural language and visual inputs. This paradigm has spurred a growing body of research exploring code as a structured and interpretable action representation for robotics~\cite{liang2023code, Singh2022ProgPromptGS, Vemprala2023ChatGPTFR, 10.5555/3692070.3693552, Chen2024RoboScriptCG, Lin_2023, Wang2025ChainofModalityLM, Huang2023Instruct2ActMM}. \Cref{tab:code_action} summarizes representative VLA models that utilize code-based action tokens.

\subsection{Evolution of Code-Based Action}

Two foundational works pioneer the use of code-based action representations in VLA research: Code as Policies~\cite{liang2023code} and ProgPrompt~\cite{Singh2022ProgPromptGS}. Code as Policies utilizes LLMs like GPT-3 or Codex~\cite{chen2021evaluating} to map language instructions to Python code snippets. This generated code processes perceptual inputs, parameterizes low-level robot APIs, and executes tasks on the robot platform. A key capability is its natural integration with third-party libraries like NumPy to perform complex spatial reasoning. At the same time, the system also generalizes effectively to new objects by bootstrapping from perception modules. This modularity allows its policy code to adapt to new behaviors through new instructions and APIs. Building on this, ProgPrompt extends the code generation process with a finite state machine (FSM) framework. Specifically, ProgPrompt employs programmatic structures in prompts to guide LLMs, which integrates import declarations to specify robot capabilities, natural language comments to scaffold high-level reasoning, and assertions to validate execution states. The FSM framework orchestrates overall task execution, which defines explicit subtask transitions and uses a reactive trigger mechanism, enabling the system to adapt to dynamic environmental changes.

Recent research extends code-based action tokens by integrating commonsense reasoning and improving the grounding of the generated code in the physical world. For instance, ChatGPT for Robotics~\cite{Vemprala2023ChatGPTFR} explores diverse prompting strategies, such as free-form dialogue, code prompting, XML tags, and closed-loop reasoning, to better parse human intent. To generate more effective and grounded code, it emphasizes the importance of descriptive API names and clear task specifications within the prompt. Crucially, the generated code undergoes a human-in-the-loop validation process, where feedback on its quality and safety is used for iterative improvement before final deployment on the robot. To address the perceptual limitations in Code as Policies~\cite{liang2023code}, Instruct2Act~\cite{Huang2023Instruct2ActMM} augments coding LLMs with specialized multi-modal foundation models for precise object segmentation and open-vocabulary classification. By offloading perception and semantic understanding, Instruct2Act effectively grounds high-level language instructions into precise, executable policy codes. Further advancing multimodal integration, RoboCodeX~\cite{10.5555/3692070.3693552} focuses on fusing information from diverse sources, such as various scene datasets, diverse object datasets, and procedural task descriptions. It introduces a novel tree-of-thought framework that synthesizes behaviors by combining visual, linguistic, and physical cues. The model's reasoning capabilities are enhanced through fine-tuning on a purpose-built multimodal dataset, leading to more accurate and generalizable robotic actions.

Code-based action tokens are also effective for high-level planning and task generalization. For instance, to tackle long-horizon tasks, Text2Motion~\cite{Lin_2023} leverages GPT-3 to generate valid goal states that define task success, providing a clear termination criterion for planning. To reach this goal, the framework employs a hybrid planner, which combines shooting-search planning for efficiency and greedy-search planning for reliable fallback. Addressing the practical deployment of such generated plans, RoboScript~\cite{Chen2024RoboScriptCG} introduces a unified code generation pipeline that standardizes inputs and integrates diverse perception and motion planning tools. This design significantly enhances code flexibility and adaptability across various robots. Pushing the boundaries of generalization further, Chain-of-Modality~\cite{Wang2025ChainofModalityLM} (not a VLA model) introduces a novel prompting strategy that guides VLMs to reason about multimodal human demonstrations (e.g., muscle or audio signals) to generate robot-executable code.



\subsection{Brittleness and Challenges}
Despite their advantages, code-based action tokens face several significant practical limitations. Their expressiveness is inherently constrained by the capabilities of a predefined perception and control API library~\cite{liang2023code}. When robots encounter highly dynamic, ambiguous, or previously unobserved environments, the pre-established APIs might be inadequate to accurately capture or express the novel behaviors required. Therefore, the system's adaptability and exploratory capacity in complex, open-world settings~\cite{Singh2022ProgPromptGS} are limited. For instance, if an API doesn't offer abstractions for environmental features like ``slippery surfaces'' or ``fragile objects,'' even perfectly written code will struggle to generate the nuanced actions needed for such scenarios.

This reliance on rigid symbolic representations also leads to execution brittleness. Robotic policies aren't just susceptible to internal generative errors from the LLMs (e.g., producing logically inconsistent or inefficient code); more critically, they fail when real-world environmental states violate an API's presumed preconditions. This is a core manifestation of the symbol grounding problem---where abstract symbols in code cannot reliably map to complex real-world perceptions. For example, a piece of code controlling a robotic arm for grasping might assume that the object's surface is always dry and flat. If the actual object is wet or irregularly shaped, the code, though syntactically correct, could lead to a failed grasp, object damage, or even hardware damage. This inherent brittleness directly translates into substantial safety risks, as seemingly innocuous code commands can trigger severe incidents in unforeseen circumstances. 

\input{table/code}

\subsection{Future Directions}
A promising direction for future work is the development of comprehensive API function libraries to fully unlock the potential of code-based action tokens. Such a framework should integrate a rich set of modular functions, including multi-modal perception APIs (e.g., object detection and tracking), reasoning modules (e.g., spatial relationship analysis), and robust action primitives. By providing a structured and reliable interface, this framework would enable VLMs to act as high-level orchestrators, generating executable code that composes these primitives to solve complex, long-horizon tasks in the real world.

A second future direction is integrating formal verification throughout the code lifecycle to enhance robustness. This includes verifying API libraries for consistency and safety and developing methods to dynamically verify LLM-generated code. Logical reasoning and constraint satisfaction can guide safe code generation, while static analysis and model checking catch errors or prove safety before deployment. Finally, runtime monitoring ensures API preconditions are met, triggering safe shutdowns or recovery if anomalies occur.

Another frontier is leveraging code’s interpretability to enable effective human-robot collaboration. Unlike black-box models, code’s transparency lets humans understand and intervene in a robot’s logic. This supports two key paradigms: interactive debugging, where failures can be traced and fixed in real time, and collaborative refinement, where humans iteratively guide program improvement. Such human-in-the-loop systems are crucial for developing robotic agents that are not only capable but also trustworthy and controllable.

\section{Affordance as Action Tokens}
\label{sec:affordance}

Within the VLA paradigm, affordance serves as structured and spatially grounded action tokens that bridges visual perception and physical interaction. Recent research~\cite{stone2023open, huanga3vlm, xu2024manifoundation, qi2025sofar} demonstrates that affordance representations utilize the spatial reasoning capabilities of vision-language foundation models to identify actionable regions and evaluate physical feasibility based on multimodal inputs. By abstracting away embodiment-specific control mechanisms, affordances enhance cross-platform generalization, allowing the same high-level instructions to be executed across various robotic systems. Moreover, they explicitly encode task-relevant interaction information, such as grasp points or manipulable surfaces, making them particularly effective for object-centric manipulation in real-world settings.

Affordance can be expressed in various forms, each offering distinct insights into how a robot may interact with objects in its environment. Recent research primarily explores keypoints~\cite{sundaresankite, huangrekep, kuangram}, bounding boxes~\cite{wake2024gpt, huanga3vlm, zhong2025dexgraspvla}, segmentation masks~\cite{liu2025robodexvlm, qi2025sofar}, and affordance maps~\cite{10656630, huang2023voxposer, xu2024manifoundation}. We summarize these efforts in \Cref{tab:affordance_action}. For a contact-rich task like \textit{kitchen cleanup}, the choice of representation is crucial. Keypoints provide precise targets, ideal for pinpointing a bowl's rim for grasping or pressing a small dishwasher button. Bounding boxes offer a simpler, coarse localization sufficient for general object selection. For operations requiring fine-grained interaction, such as wiping the irregular interior of a bowl, segmentation masks are superior as they capture the object's exact contour. Affordance maps provide a dense, scene-level understanding of interaction possibilities. They highlight all graspable or wipeable regions simultaneously, enabling more complex spatial reasoning across multiple objects. Ultimately, the selection of an affordance representation involves a fundamental trade-off between interaction precision, computational complexity, and the demands of the task. In the following parts, we analyze each representation’s characteristics in detail.

\input{table/affordance}

\subsection{Keypoints: Precise Interaction Anchors}
Keypoints provide a compact and precise representation of interaction targets, such as object handles or contact edges. They are typically defined as $\mathbf{k} = [\mathbf{x}, \mathbf{d}]$, where $\mathbf{x}, \mathbf{d} \in \mathbb{R}^3$, with $\mathbf{x}$ denoting the spatial contact position and $\mathbf{d}$ indicating the interaction direction. Benefiting from the precise spatial grounding capabilities of VLMs~\cite{oquab2023dinov2, kirillov2023segment, CLIP, deitke2024molmo, team2025gemini}, several early VLA models have adopted keypoints to directly link vision-language perception with control-level execution. KITE~\cite{sundaresankite} grounds language instructions in visual scenes by predicting task-relevant keypoints, which correspond to semantic object parts. These keypoints are then used in conditioned skills to carry out low-level actions. RoboPoint~\cite{yuanrobopoint} builds upon this idea by constructing a synthetic dataset to instruction-tune VLMs for spatial reasoning, allowing models to identify points satisfying relational constraints, which are subsequently executed through motion planning. CoPa~\cite{huang2024copa} further enhances spatial grounding by incorporating common sense priors from VLMs into a coarse-to-fine grounding pipeline, which first identifies plausible interaction regions and then refines them into actionable spatial constraints for subsequent motion planning. To ensure control robustness, KUDA~\cite{liu2025kuda} introduces a two-level closed-loop control mechanism to facilitate robust model-based planning. Specifically, it uses a VLM to generate task specifications that contain keypoints and their corresponding target positions. These specifications are then formulated as cost functions that guide the optimization of a two-level controller. Moreover, this system employs a retrieval-based prompt library, which strengthens few-shot grounding and system robustness.

Beyond direct grounding, keypoints have also been adopted within structured frameworks that incorporate task semantics, relational constraints, and cross-domain knowledge. RAM~\cite{kuangram} addresses the cost of in-domain data collection by constructing an affordance memory from diverse out-of-domain datasets. It uses VFMs for language-conditioned retrieval of relevant demonstrations, transferring 2D keypoints into 3D through probabilistic lifting, thus enabling zero-shot manipulation in novel environments. ReKep~\cite{huangrekep} formalizes manipulation as a constraint optimization problem over tracked keypoints, where task goals are encoded as Python functions that impose geometric and relational costs among robots and objects. A hierarchical solver plans SE(3) subgoals and optimizes actions via receding-horizon control, supporting bimanual and human-in-the-loop interaction with high spatial-temporal complexity. OmniManip~\cite{pan2025omnimanip} introduces an object-centric canonicalization process that maps objects to a functional space. Within this structured space, keypoints act as reasoning primitives over which VLMs predict spatial constraints and interaction goals. To mitigate hallucinations and execution drift, it incorporates a self-correcting loop that renders outcomes and resamples interaction points, while a dual-level controller handles high-level planning and fine-grained pose tracking.

An emerging direction extends static keypoints into temporal sequences, effectively transforming them into trajectory-based action tokens. This evolution enables systems to represent not only where to act but also how actions unfold over time. Magma~\cite{yang2025magma} and VidBot~\cite{chen2025vidbot} both predict sequences of keypoint positions conditioned on task instructions and visual observations, capturing fine-grained temporal dynamics for object-centric manipulation. By modeling temporally grounded keypoints, these systems support longer-horizon reasoning and enable temporally consistent action planning. This temporal extension enhances expressiveness and planning capability, offering a natural bridge between spatial affordance and trajectory-level representations.

\subsection{Bounding Boxes: Coarse Grounding}
Bounding boxes provide a coarse yet efficient representation for instance-level localization in the visual scene. A 2D bounding box is typically defined as $\mathcal{B} = \{({x}_\mathrm{tl}, {y}_\mathrm{tl}), ({x}_\mathrm{br}, {y}_\mathrm{br})\}$, marking the top-left and bottom-right image-plane corners. In 3D, bounding boxes are commonly represented by eight spatial corner points $\{(x_i, y_i, z_i) | i \in \{1, \ldots, 8\} \}$, encoding the object’s physical extent within the scene. While these representations lack fine-grained geometric detail, they offer robustness and computational simplicity. The advent of powerful open-vocabulary detectors (e.g., Grounding DINO~\cite{LiuZRLZYJLYSZZ24}, Detic~\cite{ZhouGJKM22}, and OWL-ViT~\cite{minderer2022simple}) and VLMs (e.g., Qwen2.5-VL~\cite{bai2025qwen2}) creates a strong connection between visual understanding and physical manipulation by effectively localizing objects based on free-form language queries into bounding boxes.

Several VLA models leverage bounding boxes to ground language instructions into object-centric visual inputs. DexGraspVLA~\cite{zhong2025dexgraspvla} grounds domain-varying referential expressions by localizing domain-invariant bounding boxes of the target objects, which are then converted into segmentation masks. These masks are tracked across time using Cutie~\cite{cheng2024putting}, enabling temporally consistent visual grounding throughout the grasping process. This pipeline illustrates a broader trend in recent work: using bounding boxes as modular interfaces that connect referential language to spatially localized object representations. Bounding boxes serve as a efficient perceptual abstraction that simplifies the mapping from language to actionable visual input, enabling task specification in open-vocabulary settings without requiring dense supervision.

Beyond object localization by language instructions, bounding boxes can also support interaction inference and downstream action generation. \citet{wake2024gpt} employs GPT-4V to process human demonstration videos, integrating hand and object bounding boxes to detect grasp and release events via spatial proximity. These spatiotemporal cues serve as the basis for extracting affordance-relevant information, including grasp strategies and waypoint trajectories, which are then translated into robot-executable code. Extending this direction, A3VLM~\cite{huanga3vlm} models object articulation using a structured triad comprising a 3D bounding box, a movement axis, and a semantic label. To enable the prediction of this triad, it introduces a dataset of object-level articulation annotations and fine-tunes the Llama-2-7B model with a projection layer. Crucially, this robot-agnostic representation can translate directly into low-level robot action by simple action primitives, enabling generalization across diverse platforms and significant manipulation performance.

\subsection{Segmentation Masks: Pixel-Level Regions}
Segmentation masks provide high-resolution spatial representations that capture fine-grained object contours and part-level geometry, enabling precise grounding of functional regions such as wipeable surfaces or graspable areas. Formally defined as binary matrices $\mathbf{M} \in \{0,1\}^{H \times W}$, masks offer pixel-level detail that surpasses coarser abstractions like bounding boxes. With the advent of foundation models such as SAM~\cite{kirillov2023segment} and Florence-2~\cite{xiao2024florence}, the quality and generalization of language-conditioned segmentation have significantly improved. Recent VLA models leverage these capabilities to extract affordance-aligned object regions from textual instructions. 
MOO~\cite{stone2023open} utilizes OWL-ViT to extract object representations, which are fused with textual instructions to inform policy learning in open-world manipulation tasks. SoFar~\cite{qi2025sofar} segments object masks using SAM, then uses them to construct object-centric point clouds and orientation-aware scene graphs. These representations guide PointSO in predicting functional directions (e.g., “handle facing up”) and support structured spatial reasoning. RoboDexVLM~\cite{liu2025robodexvlm} adopts a coarse-to-fine refinement pipeline to obtain high-quality masks, which are used to predict end-effector grasp poses via AnyGrasp~\cite{fang2023anygrasp}. Together, these methods demonstrate that segmentation masks provide structured, task-aligned representations that bridge perception and control in contact-rich manipulation tasks. A more recent direction explores the use of segmentation masks as temporally anchored interaction interfaces. ROCKET-1~\cite{cai2025rocket} introduces a hierarchical system that leverages segmentation sequences extracted and tracked across time via SAM 2~\cite{ravi2024sam} as persistent visual prompts. These temporally grounded masks support high-level reasoning and coherent action selection in dynamic environments, enabling robust object manipulation without fixed task templates.

\subsection{Affordance Maps: Dense Spatial Fields}
Affordance maps represent scenes as spatial fields that assign each region a graded suitability score for specific actions, reflecting prior interaction awareness. Typically it is formulated as $\mathbf{A} \in \mathbb{R}^{H \times W}$, where $H$ and $W$ denote spatial resolution. These maps encode object geometry, surface topology, and task-specific priors, enabling dense and instruction-conditioned interaction reasoning. CLIPort~\cite{shridhar2021cliportpathwaysroboticmanipulation} adopts a two-stream network to fuse semantic and spatial features for affordance prediction, guiding precise pick-place actions. IGANet~\cite{li2024learning} learns to generate pixel-wise affordance distributions conditioned on language inputs, allowing the same object to afford different actions under varying instructions. VoxPoser~\cite{huang2023voxposer} expands this concept by prompting LLMs to synthesize affordance and constraint specifications in code form, which are then grounded to perceptual space via VLMs to form 3D value maps. These maps enable zero-shot trajectory synthesis over diverse tasks and objects without retraining. 

Beyond spatial grounding, affordance maps also support reasoning about physical contact and manipulation dynamics. ManipLLM~\cite{10656630} incorporates affordance maps into a multimodal chain-of-thought framework, using them to encode region-level priors that guide manipulation-aware pose generation. The maps indicate where actions are most likely to induce meaningful object motion, improving precision and stability in complex scenes. ManiFoundation~\cite{xu2024manifoundation} further extends this line of work by treating manipulation as contact synthesis, leveraging force and motion heatmaps to represent contact-centric affordances. These maps encode where contact should occur, the force to apply, and the expected motion trajectory, enabling robust contact prediction for both rigid and deformable objects. As task complexity increases, such structured affordance priors offer a scalable solution for grounding low-level control in physically realistic interaction fields.

\subsection{Discussion and Future Directions}

Despite their advantages, affordance-based action tokens face several limitations that hinder effectiveness in real-world manipulation. First, most VLA models rely on 2D image representations, which inadequately capture the 3D geometry and spatial relationships required for precise control. Although models like A3VLM~\cite{huanga3vlm} and SoFar~\cite{qi2025sofar} incorporate partial 3D information, they still fall short in tasks involving complex object shapes and occlusions and scenarios common in dynamic (e.g., inserting components into moving assemblies) or delicate (e.g., fine-grained part assembly) manipulations. Second, affordance tokens typically encode static object properties such as a ``graspable handle'' or ``closeable door'' without modeling how these affordances evolve over time. These limitations impair their effectiveness in contact-rich tasks that demand continuous reasoning about changing affordance states. Finally, affordance representations are vulnerable to visual perturbations such as occlusion and motion blur. Specifically, keypoints degrade significantly under occlusion, and segmentation masks lose accuracy in visually challenging scenes, compromising manipulation performance.

To address these challenges, we identify three promising research directions.

\textbf{Learning True 3D Affordances.} A critical next step is to move beyond 2D or projected 3D and learn affordances directly within native 3D representations. By grounding policies in structures like Neural Radiance Fields~\cite{978-3-030-58452-8_24}, 3D Gaussian Splatting~\cite{10.1145/3592433}, or explicit meshes, models can develop a holistic understanding of object geometry, free space, and occlusion. This approach would unlock robust reasoning for complex tasks currently beyond reach, such as inserting a part into a hidden cavity or manipulating non-rigid objects in clutter.
    
\textbf{Modeling Temporal Affordance Dynamics.} Future models should learn to predict how actions alter an object's affordances over time. For example, a model should infer that executing a ``lift lid'' action transitions the affordance state from ``openable'' to ``pourable''. This temporal reasoning is fundamental for enabling long-horizon planning and succeeding in contact-rich, sequential tasks.

\textbf{Enhancing Policy Robustness and Uncertainty-Awareness.} Real-world deployment demands policies that are resilient to visual ambiguity and aware of their own limitations. This requires a dual focus. Models should be trained for greater robustness against visual perturbations using techniques like advanced data augmentation. And policies should quantify their own uncertainty by outputting probabilistic affordances.

\section{Trajectory as Action Tokens}
\label{sec:trajectory}

One of the central challenges in scaling VLA models lies in the limited availability of robot data, particularly those annotated with action labels. To address this constraint, recent studies~\cite{wen2023any, ko2023avdc, li2025hamster, xu2024flow} have proposed leveraging off-domain video data, which typically lacks explicit action annotations. These works use trajectories as a proxy for action representations since they can be readily extracted from videos and encapsulate rich, actionable information across the entire manipulation process. We summarize representative trajectory-based methods in \Cref{tab:trajectory_action}. In comparison with latent representations (\Cref{sec:latent}) proposed by other works~\cite{bruce2024genie, ye2024latent, bu2025agibot, cai2023groot, cai2024groot, mete2024quest, shafiullah2022behavior, lee2024behavior}, trajectory is a relatively explicit action representation that is both explainable and understandable by humans, facilitating training and debugging. 
Another major challenge in VLA research is task generalization. For instance, policies conditioned on language-based action tokens often struggle to generalize zero-shot across semantically different tasks with similar low-level motion patterns---such as generalizing from “wiping a table” to “sliding a block on a desk”. In contrast, trajectory-conditioned policies exhibit stronger generalization capabilities across such tasks, as demonstrated by RT-Trajectory~\cite{rt-trajectory}.

\subsection{Overview of Trajectories}

Trajectory-based action tokens can be categorized into three distinct forms: \textit{Point Trajectory}, \textit{Visual Trajectory}, and \textit{Optical Flow}. Each represents motion with a different level of abstraction and information density. 

\textbf{Point Trajectory} is the most direct approach, encoding an action as a sequence of discrete points, denoted as $\mathbf{P} \in \mathbb{R}^{T \times K \times 2}$. This method models the path of $K$ critical points over a time span $T$, which offers targeted and numerically precise guidance. In autonomous driving, models predict future vehicle waypoints in Bird's Eye View (BEV) space~\cite{covla_wacv2025, tian2024drivevlm, hwang2024emma, liu2025vlm}. For robotic manipulation tasks, they generate 2D coordinate paths for end-effectors or objects within the image plane~\cite{wen2023any, flip}. 

\textbf{Visual Trajectory} directly renders a path into the pixel space. Instead of just a list of coordinates, the output is a new image or video where the intended motion is visually depicted. This can be achieved by overlaying point sequences onto observation frames~\cite{li2025hamster, rt-trajectory} denoted as $\mathbf{I} \in \mathbb{R}^{H\times W\times 3}$ or by generating a video flow~\cite{xu2024flow} that materializes as visible curves over time, such as $\mathbf{I} \in \mathbb{R}^{T\times H\times W\times 3}$. This form is highly interpretable as it shows the action in its visual context.

\textbf{Optical Flow} offers the densest representation, formulated as a motion field $\mathbf{V}\in\mathbb{R}^{H\times W\times 2}$. This field describes the motion of every pixel between frames, capturing the holistic dynamics of the entire scene rather than a single path. By treating the collective movement of the scene as the action signal, this method can model complex, multi-object interactions implicitly~\cite{ko2023avdc, yao2025thinksmallactbig}.

\input{table/trajectory}
\subsection{Progress and Key Papers}


Data scarcity has long been a bottleneck in robotics. Trajectory-based action tokens offer a solution by enabling learning from abundant off-domain videos. AVDC~\cite{ko2023avdc} predicts future frames using a diffusion model trained on human or robot demonstration videos and generates optical flow using pretrained models, guiding downstream control with depth information. However, this is computationally expensive and prone to hallucinations. ATM~\cite{wen2023any} mitigates these issues by predicting trajectories of arbitrary points and requires only a small amount of in-domain action-labeled data for low-level policy training. In contrast, Im2Flow2Act~\cite{xu2024flow} requires no real-world robot data. It learns to generate video trajectories from human demonstration videos and trains a trajectory-conditioned policy using simulation data. To bridge the embodiment gap, Im2Flow2Act focuses on object flow instead of arbitrary point flow. FLIP~\cite{flip} incorporates a world model built from videos, including dynamics, action, and value modules. It performs model-based planning and predicts action conditioned on both flow and video plan. Compared to ATM, FLIP samples denser flow points and achieves better performance, demonstrating the effectiveness of dense flow in low-level control.


Trajectory-based action tokens demonstrate strong generalization across tasks, as well as visual and semantic variations. Even when tasks are semantically distinct, shared motion patterns in trajectory space enable cross-task generalization. For example, RT-Trajectory~\cite{rt-trajectory} encodes tasks via coarse 2D or 2.5D end-effector motion trajectories, on which an end-to-end policy (i.e., RT-1) is conditioned. RT-Trajectory outperforms RT-1~\cite{Brohan2022RT1RT}, RT-2~\cite{Brohan2023RT2VM}, and RT-1-Goal (RT-1 conditioned on goal images) on unseen tasks. In comparison with RT-Trajectory, HAMSTER~\cite{li2025hamster} adopts a hierarchical architecture, using a VLM to synthesize 2D trajectories and a low-level policy conditioned on 3D observations. This structure facilitates fine-tuning on large-scale off-domain datasets, such as RoboPoint~\cite{yuanrobopoint}, thus improving its visual and semantic generalization.


Another direction focuses on pretraining large models on trajectory-centric data. LLARVA~\cite{niu2024llarva} constructs a unified robotic LLM via instruction tuning, incorporating structured information such as control mode, task, and proprioception. It outputs 2D trajectories and robot actions in text, showing greater flexibility across control modes. Despite leveraging 8.5M vision-action pairs from Open X-Embodiment (OXE)~\cite{RT-X}, its scale remains smaller than conventional LLM/VLM datasets. To leverage broader datasets, ARM4R~\cite{niu2025pre} introduces a three-stage training paradigm: pretraining on EPIC-KITCHENS-100~\cite{Damen2020RESCALING}, fine-tuning on 1-2K robot demonstrations, and predicting proprioceptive states. 
Its 4D trajectory representation enables superior performance over LLARVA and ATM. Magma~\cite{yang2025magma} is a foundation model for both UI navigation and robotic manipulation, which is trained on heterogeneous datasets with Set-of-Mark and Trace-of-Mark, endowing it with spatial-temporal reasoning capabilities that surpass VLA models trained solely on robot data like OpenVLA~\cite{kim2024openvla}.

\subsection{Trajectory-Related Data}

Various types of data can be utilized to train trajectory-based VLA, such as internet-scale vision-language datasets, human videos, and existing robot data. Web-scale vision-language pairs can instill broad common sense into the policy. Some approaches~\cite{li2025hamster} utilize VLMs to directly output keypoint sequences, which require vision-language datasets such as object location tasks~\cite{yuanrobopoint} in the co-training phase to keep the VLM's generalization ability. Human and robot demonstrations further provide specific actionable knowledge. Trajectory labels can be directly extracted from existing videos without human annotations. One option is to use point tracking tools such as CoTracker~\cite{karaev2024cotracker}, TAPIR~\cite{doersch2023tapir}, or optical flow methods like RAFT~\cite{teed2020raftrecurrentallpairsfield}. Another line of work, such as RT-Trajectory~\cite{rt-trajectory}, extracts 2.5D trajectories from robot demonstrations using end-effector states. In either way, all the existing demonstration datasets, no matter human, simulated or real-robot, can be utilized with ease. In autonomous driving, trajectories and captions can also be automatically generated using pipelines like that in CoVLA~\cite{covla_wacv2025}, which combines Kalman Filter~\cite{kalman1960new}-based trajectory prediction with rule-based and VLM-driven captioning.


\subsection{Discussion and Future Directions}

Despite their advantages, trajectory-based action tokens face several key challenges. We identify three main areas: 3D spatial understanding, computational efficiency, and task suitability. Most work utilizes 2D trajectories, but 2D trajectories lack explicit 3D information. This can introduce ambiguity and restrict their applicability to non-planar tasks. Depth data serve as a critical supplement: AVDC~\cite{ko2023avdc}, RT-Trajectory~\cite{rt-trajectory}, and HAMSTER~\cite{li2025hamster} all incorporate depth information to mitigate this issue and provide a richer 3D understanding. A more fundamental challenge, however, is that point trajectories typically encode only position. They omit crucial orientation information, making them ill-suited for complex dexterous manipulation tasks. Future work could explore integrating full 3D spatial information into trajectory representations.

Another significant challenge is computational efficiency. Many methods employ generative models to predict trajectories or videos, which are computationally expensive to train and to inference~\cite{ko2023avdc, wen2023any, xu2024flow}. Other methods leverage VLMs to predict trajectories, but VLMs often output waypoints at a low frequency, insufficient for smooth control~\cite{rt-trajectory, niu2024llarva, li2025hamster}. One solution is to use traditional planning methods to refine these sparse outputs into high-frequency control signals~\cite{tian2024drivevlm}. To avoid re-planning at every timestep, other approaches predict a full trajectory once and use a temporal alignment module for real-time execution~\cite{xu2024flow}. Developing lightweight yet expressive trajectory generation models remains a critical research direction.

Finally, the suitability of a trajectory depends on the task and environment. Trajectories excel at tasks defined by precise motion paths, such as surface wiping or navigation. However, they are less effective in partially observed settings where a complete path cannot be planned upfront. Furthermore, they lack the semantic richness for tasks involving complex interaction logic and do not inherently capture concepts like applying force or understanding object affordances. A promising future direction involves creating hybrid action tokens that combine trajectory tokens with semantic concepts (e.g., ``grasp'', ``increase force''), enabling robots to handle a wider and more complex set of tasks.

\section{Goal State as Action Tokens}
\label{sec:goal-state}

When humans approach manipulation tasks, our brains don't just translate raw perception directly into action. Instead, we often engage in a mental simulation, envisioning the desired outcomes before executing any steps. For instance, if asked to ``clean up the table'', one first conceptualizes a neat and organized table, then works backward to determine the necessary actions. Drawing inspiration from this powerful human cognitive strategy, a growing body of research in VLA models proposes utilizing a predicted goal state---a visual representation of the task's intended outcome---as an intermediate action token. These works, including recent advancements like 3D-VLA~\cite{zhen20243d}, FLIP~\cite{flip}, and VPP~\cite{hu2025videopredictionpolicygeneralist}, aim to bridge the gap between high-level instructions and low-level actions by grounding the ``what to do'' in a visually rich and interpretable form. 

Typically, models employing goal states as action tokens adopt a hierarchical architecture. A high-level model, often a generative model like DiT~\citep{peebles2023scalable} or CVAE, is responsible for synthesizing the goal state based on the current observation and language instruction conditions. This generated goal state then conditions a lower-level model, such as a diffusion policy or MLP, which translates it into the final sequence of actions. This setup effectively establishes the goal state as a crucial mental simulation step, situated between comprehending the instruction and synthesizing the actions. 
Goal states can be broadly categorized into two primary types based on their temporal dimension: single-frame images and multi-frame videos. 
To give a concise overview, ~\Cref{tab:goal_state} lists the principal methods discussed in this section.

\input{table/goal_state}

\subsection{Single-Frame Image as Goal State}
Single-frame goal states typically take the form of 2D RGB images, 2.5D RGB-D images, or 3D point clouds to depict the entire desired scene, as demonstrated in recent works~\cite{zhao2025cotvlavisualchainofthoughtreasoning, Ni_2024_CVPR, zhen20243d, black2023zeroshotroboticmanipulationpretrained}, offering many key advantages. 
For instance, LangLfP~\cite{LangLfP} demonstrates how methods leveraging goal images can achieve easy data scalability via hindsight relabeling. This technique ingests unsegmented streams of robot play data, automatically samples short windows, and treats each window’s final frame as a goal image. This process autonomously generates a large-scale robot action dataset with goal image annotations, entirely bypassing the need for manual labeling. 
Building upon the utility of goal images for data scaling and low-level control, subsequent works integrate high-level goal image generation to create complete hierarchical VLA models. For example, SuSIE~\cite{black2023zeroshotroboticmanipulationpretrained} first leverages a simple image-generative model for visuo-semantic reasoning before deferring to a low-level policy to determine precise motor actuations.

Specifically, a high-level diffusion model generates goal images from language instructions, and a lower-level DDPM decodes those images into the required action sequence. CoTDiffusion~\cite{Ni_2024_CVPR} further extends SuSIE's hierarchical diffusion architecture by integrating a semantic alignment module, which enables the diffusion model to assess its own task completion progress. 
Another significant advantage of using goal images is their ability to leverage action-free videos for training the high-level goal image generator. CoT-VLA~\cite{zhao2025cotvlavisualchainofthoughtreasoning}, for example, exploits action-free human videos to train its goal image generator. Unlike the diffusion-based architectures mentioned above, both stages in CoT-VLA are autoregressive VLMs~\cite{vilau}: the high-level model uses causal attention to synthesize goal images, while the lower-level model uses non-causal attention to generate corresponding action sequences. Beyond standard RGB images, some works like 3D-VLA~\cite{zhen20243d} have extended single-frame goal states to encompass RGB-D images and point clouds. By enriching the visual encoding with depth and 3D geometric configuration, these approaches provide a more precisely grounded and perceptually rich depiction of task goals.

\subsection{Multi-Frame Video as Goal State}
Multi-frame goal states (typically short videos) offer a richer temporal context compared with single-frame goal states. By capturing how scenes evolve, this additional temporal dimension provides crucial ``how-to-do'' cues, significantly reducing execution ambiguity and offering finer-grained motion information. Research in this area leverages multi-frame goal states through various innovations: 
    \textbf{Generating from Large-Scale Data}---One approach focuses on generating future video content from vast datasets to inform action. UniPi~\cite{du2023learninguniversalpoliciestextguided}, for instance, pioneered using internet-scale data for text-conditioned video generation, with an inverse dynamics model (MLP) then computing actions from these predicted video sequences. 
    \textbf{Extracting Implicit Action Cues from Videos}---Other works concentrate on extracting explicit or implicit action-relevant information directly from the generated goal videos. AVDC~\cite{ko2023avdc}, for example, enables the model to leverage dense correspondences within the video without relying on any action labels. It achieves this by using a diffusion model to synthesize future video frames and then extracts dense pixel-wise optical flow from these frames, which can then guide the lower-level policy. This method effectively translates visual motion into actionable guidance. 
    \textbf{Enhancing Generalization and Robustness}---Multi-frame goal states are also explored for improving model generalization and robustness. Acknowledging that embodiment-specific strategies limit broader generalization, Gen2Act~\cite{bharadhwaj2024gen2acthumanvideogeneration} and FLIP~\cite{flip} enhance cross-embodiment generalization by generating human-executed goal videos rather than robot-specific ones, thereby reducing the reliance on robot-specific fine-tuning. Similarly, GEVRM~\cite{zhang2025gevrmgoalexpressivevideogeneration} introduces an auxiliary state-alignment loss specifically designed to improve robustness against external perturbations. 
    \textbf{Strategies for Complex Long-Horizon Tasks}---For complex, long-horizon tasks, researchers typically employ two main approaches. One common method, exemplified by works like Gen2Act, directly leverages LLMs to decompose long-range tasks into shorter subtasks, then subsequently runs the same model for each of these shorter segments.
    The second approach involves using multiple candidate goal videos for improved planning. VLP~\cite{du2023video} generates and scores multiple candidate goal videos with a separate VLM, using a beam-search-like algorithm to select optimal long-term strategies for subtasks. Similarly, FLIP~\cite{flip} adapts a language-image valuing model (LIV)~\cite{ma2023livlanguageimagerepresentationsrewards} to evaluate candidate human-executed goal videos (synthesized by a DiT network from keypoint trajectories), then uses a beam-search-like algorithm to choose the best long-term option. These methods demonstrate sophisticated planning with multi-frame goals. 


\subsection{Advantages of Goal State}
Goal states offer several key advantages that significantly boost their effectiveness as action tokens. Primarily, goal states provide great \textbf{data scalability}. This is enabled through hindsight goal relabeling, which allows for the autonomous generation of vast training datasets by extracting single-frame and multi-frame goal states from raw robot trajectories, fundamentally bypassing the action annotation bottleneck. 
Moreover, using goal states unlocks access to \textbf{broader training data sources and enhanced generalization capabilities}. Their generators can leverage large-scale action-free video data to learn real-world dynamics, improving overall generalization. Furthermore, training on human-executed goal states (e.g., Gen2Act~\cite{bharadhwaj2024gen2acthumanvideogeneration}) specifically boosts their cross-embodiment generalization, enhancing knowledge transfer across different robot platforms.
Beyond data, goal states also enhance \textbf{task specificity}. By encoding highly precise spatial and visual information, they act as clear action tokens that reduce ambiguity in complex tasks, providing lower-level policies with accurate visual instructions for fine-grained action execution. These models also boast \textbf{robust interpretability}; their ``white-box'' training and inference processes make human understanding, debugging, and intervention more feasible. Additionally, goal states lend themselves to \textbf{straightforward evaluation}. Off-the-shelf language-image valuing models, like those adapted in FLIP~\cite{flip}, can easily assess goal-state quality by checking their alignment with language instructions. 

\subsection{Limitations and Future Directions}
Despite their notable advantages, goal states inherently possess several limitations. 
\textbf{Generating high-quality and consistent goal states remains challenging}, often manifesting as overspecification or outright inaccuracies. Overspecification occurs when the generated goal state contains unnecessary or overly precise details. This can lead the lower-level policy to focus on trivial aspects, overconstrain its flexibility, or even make the task harder to complete if those exact details aren't critical, thereby undermining the policy's generalization to slight variations in the environment or task execution. To mitigate this problem, VPP~\cite{hu2025videopredictionpolicygeneralist} synthesizes goal videos by performing only a single denoising step with its high-level diffusion model, conveying only coarse action and omitting some fine-grained details, thus partially alleviating overspecification. Conversely, inaccuracies imply the generated goal state is fundamentally incorrect, inconsistent with the desired outcome, physically implausible, or exhibits temporal and spatial inconsistencies due to insufficient dynamics modeling~\cite{zhang2025gevrmgoalexpressivevideogeneration}. Such erroneous goals directly provide misleading guidance, inevitably causing the lower-level policy to attempt wrong actions and resulting in task failure. 
Additionally, \textbf{generating future images or videos inherently introduces high inference latency} due to significant computational overhead. For instance, AVDC~\cite{ko2023avdc} requires approximately 10 seconds to synthesize an 8-frame goal video. This substantial delay is further compounded by the lower-level policy's need to condition on these computationally intensive goal states for action sequence generation. Some approaches, like Gen2Act~\cite{bharadhwaj2024gen2acthumanvideogeneration}, achieve only 3 Hz inference speed, making real-time robotic control difficult. Even VPP, which mitigates some of this by performing only a single denoising step when generating goal states, can still only achieve a control frequency of 7-10 Hz.

Goal states as action tokens represent an auspicious direction in VLA model development, offering superior data scalability, rich visual guidance, and strong interpretability. The rapid advancement of image and video generation (exemplified by diffusion models and large-scale video generation models) provides an increasingly solid foundation for this paradigm, as higher-quality and more temporally consistent visual content will better leverage the goal-specified nature of this approach by providing embodied agents with precise and rich visual guidance. Google's recently unveiled Veo 3 video generation model demonstrates exceptional performance in both image quality and physical constraint adherence. Beyond improvements in generation quality, several key research directions warrant exploration: improving computational efficiency to enable real-time robotic control, enhancing robustness to environmental variations for deployment in real-world scenarios, and developing more efficient approaches for long-horizon task planning, as current methods either rely heavily on LLM-based task decomposition (which is limited by the quality of subtask segmentation) or employ computationally expensive beam-search-like strategies for candidate goal evaluation. Addressing these limitations will be crucial for establishing goal states as a highly effective and widely applicable action token in VLA models.

\section{Latent Representation as Action Tokens}
\label{sec:latent}

\begin{figure}[!t]
    \vspace{-1em}
    \centering
    \includegraphics[width=0.93\linewidth, keepaspectratio]{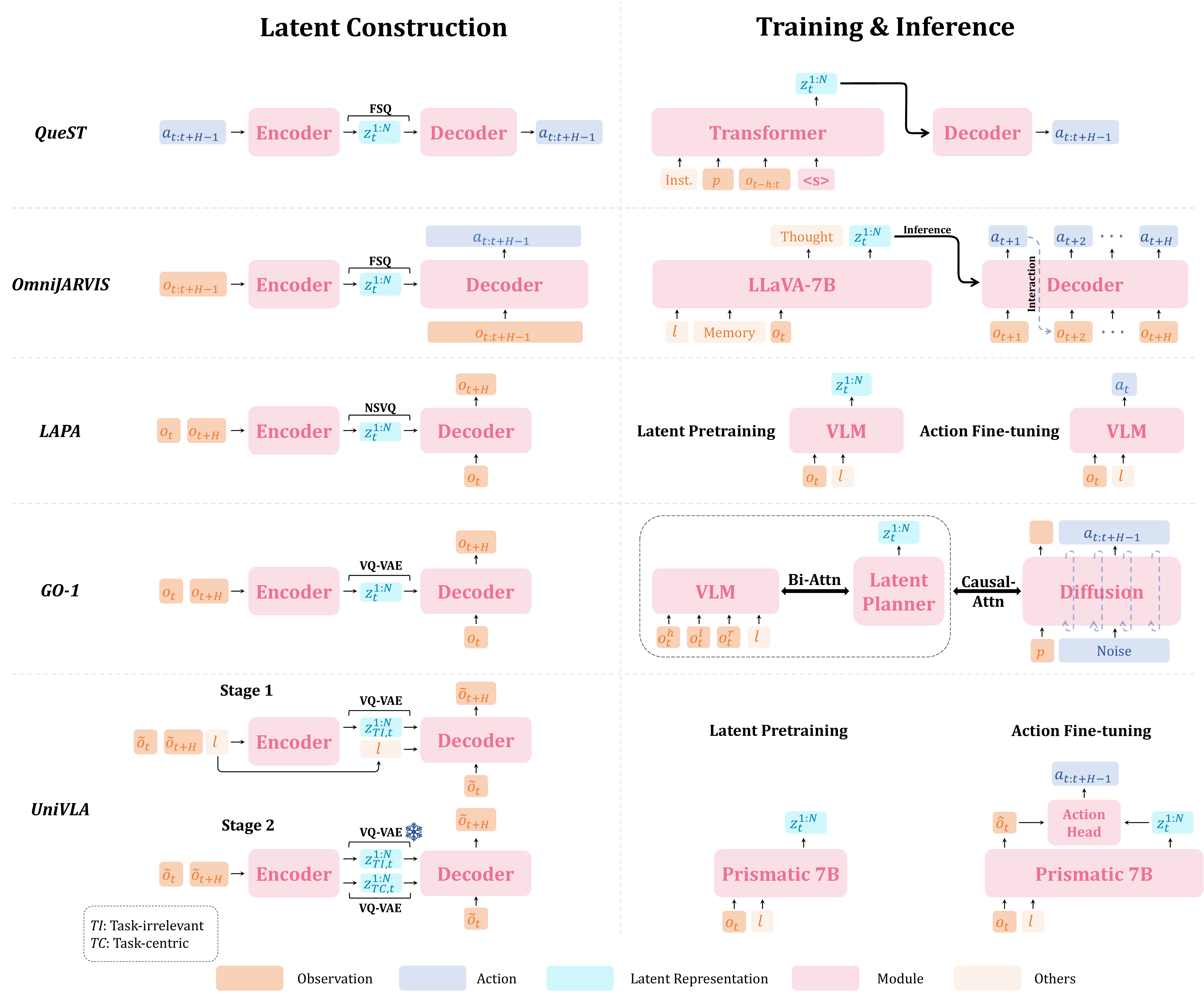}
    \caption{A unified visualization of representative methods (rows) that utilize latent representations as action tokens, highlighting their diverse strategies for latent space construction, training, and inference (columns). Inst.: Instruction, $p$: Proprioception, $l$: Language Instruction.}
    \label{fig:latent}
\end{figure}

\input{table/latent}


Embodied AI faces a fundamental challenge due to the limited availability of large-scale, embodiment-specific, and action-labeled datasets. To overcome this data bottleneck, researchers have turned to more scalable data sources, such as web-scale human activity videos (e.g., Ego4D~\cite{Ego4D2022CVPR}) and heterogeneous cross-embodiment robot datasets. Although these sources are abundant, they often lack explicit action annotations or suffer from significant embodiment gaps, making them difficult to leverage directly. A promising approach is to extract unified, embodiment-agnostic \emph{latent} action representations from such data, which encode high-level semantic behaviors---such as grasping or turning left---and effectively model real-world dynamics to support robot learning. This idea, along with its extensions and variants, has been explored in a series of VLA models that employ latent representations as action tokens.

Typically, these methods are realized through a three-stage pipeline, as illustrated in \Cref{fig:latent}. 
The initial \textbf{Latent Construction} stage constructs a latent action space from a large dataset in an unsupervised way, providing pseudo-labels for the subsequent stage.
Next, in the \textbf{Latent Pretraining} stage, a VLM is adapted to predict the appropriate latent actions given the current observation and instruction. The final \textbf{Action Fine-tuning} stage trains the VLA to translate the predicted high-level latent actions into low-level, executable commands for the target embodiment.
Based on what these latent actions represent, the approaches are broadly categorized as either vision-based, action-based, or goal-based.
\Cref{tab:latent} provides a comprehensive overview of the representative methods discussed in this section.

\subsection{Vision-Based Latent Representation}

Vision-based latent construction primarily utilizes a VQ-VAE~\cite{van2017neural} style architecture to model visual state transitions. The model learns by reconstructing a future goal observation from previous observations, conditioned on a sequence of latent codes $z ^ {1:N}$ from the VQ-VAE's codebook. The information bottleneck inherent to this framework compels these codes to distill the visual transformations between states, which contain information about the underlying actions. 
Genie~\cite{bruce2024genie} exemplifies this approach, training on internet game videos to produce a world model controlled entirely by latent actions. These learned actions demonstrate remarkable semantic consistency, enabling coherent control not only across different games but also when generalized to real-world robotic scenarios.
LAPA~\cite{ye2024latent} applies this method to robotic manipulation by tokenizing the learned discrete latent actions and employing a VLM for latent action prediction. This strategy demonstrates superior cross-embodiment learning capabilities, outperforming pretraining on ground-truth action labels when the agent's embodiment shifts between the pretraining and fine-tuning stages.
GO-1~\cite{bu2025agibot} further refines this approach using $\pi_0$~\cite{black2410pi0}-like architecture, which integrates a VLM, a latent planner, and a diffusion-based action head into a shared backbone through causal, layer-by-layer conditioning. This unified architecture predicts latent action and generates fine-grained, high-frequency motion for downstream tasks. Real-world experiments validate the latent planner's effectiveness by demonstrating performance gains over baselines without it.
However, a key challenge with vision-based methods is that the resulting latent space can inadvertently capture task-irrelevant visual variations, such as background clutter or camera shakiness. UniVLA~\cite{bu2025univla} mitigates this issue by first transforming raw pixels into patch-level semantic features via DINOv2~\cite{oquab2023dinov2}. It then employs a two-stage training scheme that uses language instructions to explicitly disentangle the latent space into task-centric and task-irrelevant action tokens. Ablation results show that the latent space constructed by UniVLA proves $6.4\%$ more effective than that produced using Genie's approach.

\subsection{Action-Based Latent Representation}

Different from vision-based approaches, another line of work adopts action-based latent representation, which learns a latent skill space by directly encoding and reconstructing action chunks of a fixed length $H$. For instance, QueST~\cite{mete2024quest} applies FSQ~\cite{mentzer2023finite} to these chunks from a multi-task manipulation dataset, learning a task-agnostic vocabulary of action primitives (e.g., reaching, grasping, or lifting). Experiments confirm the value of this approach: visualizations show that semantically similar behaviors cluster together, and the learned skills demonstrate effective few-shot transfer to new tasks. While effective, this approach's reliance on action-labeled data for the pretraining stage limits its scalability and cross-embodiment generalization.

\subsection{Goal-Based Latent Representation}

Distinct from methods that model short-term visual transitions or action primitives, goal-based representations encode an entire task's trajectory into latent vectors that represent the overall goal. This paradigm has proven particularly effective in virtual open-world environments, such as Minecraft~\cite{cai2024minsstudio}. 
Pioneering methods in this domain, such as GROOT~\cite{cai2023groot} and GROOT-2~\cite{cai2024groot}, employ a VAE~\cite{kingma2013auto} to encode the observation sequences of the entire task into a sequence of continuous latent vectors. Subsequently, a decoder, conditioned on these latent vectors, causally reconstructs the corresponding action sequence from observations. 
However, as discussed in GROOT-2, this latent space is prone to two failure modes---mechanical imitation of low-level trajectories and posterior collapse, leading to a deviation from the intended goal information. To better align the latent space with task-relevant goals and address these issues, GROOT-2 introduces weak supervision by encouraging the encoded latent goals to match the encoded language instructions through an MLE objective.
Despite these improvements, these methods lack reasoning and long-horizon planning capabilities. OmniJARVIS~\cite{wang2024omnijarvis} addresses this by adapting a VLM to jointly model discrete latent goals alongside vision and language tokens encompassing observation, instruction, memory, and thought. This approach ensures both strong reasoning and efficient decision-making capabilities, as demonstrated by its capacity to answer Minecraft-related questions and successfully execute complex, long-horizon tasks such as mining diamonds, which were previously unachievable.

\subsection{Advantages of Latent Representation}
Leveraging latent representations as action tokens yields several key advantages in scalability, training efficiency, and expressive power.
Primarily, vision-based latent representation enables models to scale across action-free, internet-scale human videos and cross-embodiment robot datasets, fostering an embodiment-agnostic understanding of physical dynamics that enhances generalization and allows for efficient downstream embodiment-specific fine-tuning. 
This scalability is complemented by significant gains in training efficiency. By encoding high-level kinematic semantics into a compact sequence, the latent space presents a much simpler pretraining target for VLMs than raw action. For instance, UniVLA achieves performance comparable to OpenVLA~\cite{kim2024openvla} using only $4.45\%$ of the training time.
Finally, latent representations offer strong expressive potential due to their ability to learn more compact and efficient structures, implicitly encode task-relevant semantics that are difficult to specify through explicit formats, and support the integration of non-visual and non-linguistic modalities—such as tactile feedback and audio—that are typically inaccessible to language- and vision-based action tokens such as language plans or keypoints.

\subsection{Limitations and Future Directions}
Although latent representations offer the aforementioned advantages, a key limitation lies in their inherent lack of explainability and controllability, which prevents humans from intervening or correcting policy failures, as is possible in methods like RT-H~\cite{rth2024arxiv}, thereby making interpretation and debugging more difficult. Therefore, latent representations may be unsuitable in scenarios where strict safety or reliability guarantees are required. 

Given the inherent uninterpretability of latent representations, the properties and quality of their construction become critically important.
Future research should therefore concentrate on three key directions. The first is achieving appropriate \textbf{granularity}: the latent space must be fine-grained enough to represent the subtle variations required for dexterous tasks, yet abstract enough to avoid unnecessary complexity and rote memorization. Current vision-based approaches often suffer from inadequate granularity and low reconstruction fidelity, limiting their effectiveness in highly dexterous tasks such as fine-grained manipulation.
The second is \textbf{comprehensiveness}: the latent space must encompass the full spectrum of behaviors required for a given task domain, as an incomplete vocabulary of behaviors will inevitably lead to policy failures when the agent encounters situations outside its learned repertoire.
The third crucial focus is ensuring strong \textbf{alignment} with human intention. As highlighted in the discussions of UniVLA and GROOT-2, latent spaces derived from both vision and action data can inadvertently encode information irrelevant to the given instruction. Developing robust methods to disentangle task-centric signals from this noise is therefore essential. We believe that progress focusing on these three axes---improving representational granularity, comprehensiveness and strengthening alignment with human intention---will be critical for advancing the capabilities and reliability of approaches that leverage latent representation as action tokens.

\section{Raw Action as Action Tokens}
\label{sec:raw-action}

In the previous sections, we discussed various forms of action tokens that encode actionable guidance. These tokens typically serve as intermediate outputs of VLA modules, which are ultimately mapped to raw actions. Each form of action token exhibits distinct characteristics, making it suitable for particular domains. However, choosing an appropriate token representation can be non-trivial. In such cases, a straightforward and intuitive alternative is to formulate VLA models as a direct mapping from vision and language inputs to raw actions.

This strategy is further motivated by the success of foundation models, which are trained on large-scale, diverse, task-agnostic datasets and are able to achieve strong performance on downstream tasks in a zero-shot or few-shot manner, demonstrating generalization and scalability. Similarly, the typical approach is to collect large-scale real-world robot datasets with natural language annotations and train VLA models end-to-end to directly predict raw actions. The overarching objective is that, as the dataset grows in size and diversity and base models become more capable, the resulting VLA model can learn a general-purpose robotic policy. Given the strong parallels between this training paradigm and that of foundation models, many techniques and best practices developed in the foundation model community can be inherited and adapted to this setting.

This section reviews the progress along this direction, with representative works summarized in \Cref{tab:raw_action_p1}.

\input{table/raw_action}

\subsection{Vision-Language Feature Fusion}

In the early stages, the most common approach involves fusing vision and language modules to obtain multi-modal features for downstream tasks. These fused representations are then mapped to raw actions through simple layers. LangLfP~\cite{LangLfP} represents one of the earliest VLA models. It uses MLP and CNN to encode the inputs and employs a CVAE decoder to generate action sequences. To scale up data volume, LangLfP combines 10M goal-image-conditioned state-action pairs with 10K human-labeled language-conditioned samples. BC-Z~\cite{BC-Z} is one of the first works to collect a large dataset (26K robot data and 19K human videos) to study how data scaling helps generalizable policy training. It utilizes ResNet~\cite{he2016deep} and multilingual sentence encoder~\cite{yang2019multilingual} but improves the fusion process by using multi-stage FiLM conditioning~\cite{perez2018film}, which dynamically modulates visual features based on language inputs. This approach allows more fine-grained instructions grounding and decodes actions with a simpler MLP.

\subsection{Transformer-Based Generalists}

Building on the success of scaling laws in LLMs, subsequent works have taken further steps to construct larger datasets, include more diverse task domains, and adopt autoregressive transformer backbones, with the goal of training generalists. VIMA~\cite{jiang2022vima} uses Mask R-CNN~\cite{he2017mask} and ViT~\cite{dosovitskiy2020image} to extract object tokens from visual observations, which are then concatenated with language tokens and processed by a pretrained T5 model~\cite{raffel2020exploring} to produce multi-modal prompt tokens. These tokens are used as inputs to cross-attention layers for decoding robot actions.
Gato~\cite{Gato} successfully trains a large decoder-only transformer model (1.2B parameters) on the combination of 596 control tasks (totaling 1.5T tokens) and 8 vision-language datasets. The Gato model is capable of performing a wide range of tasks across different domains, such as Atari games, robotic manipulation, VQA, and chatting tasks. By unifying vision, language, and action tokens, Gato demonstrates that a single autoregressive model can serve as a multi-modal, multi-task, and multi-embodiment generalist policy. LEO~\cite{huang2023embodied} extends this concept by incorporating additional 3D datasets to enhance the model's 3D reasoning ability, elevating the generalist model into the 3D space. This improvement strengthens LEO in embodied reasoning and planning tasks. JARVIS-VLA~\cite{JARVIS-VLA} is a Minecraft VLA model fine-tuned from pretrained VLM models (Qwen2-VL or LLaVA-NeXT). While previous VLA models typically apply imitation learning directly to fine-tune VLM on large-scale datasets for action prediction, JARVIS-VLA adopts a three-stage fine-tuning strategy: (1) text-only world knowledge fine-tuning, (2) multi-modal vision-language alignment and spatial grounding, and (3) instruction-following imitation learning.

\subsection{Autoregressive Robot VLA}
With increasing attention on robotics, RT-1~\cite{Brohan2022RT1RT} introduces the largest robotic manipulation dataset at the time, featuring 130K demonstrations across over 700 tasks, and trains a transformer-based model for real robots. It utilizes FiLM-conditioned EfficientNet, allowing language to modulate visual features. The transformer decoder then autoregressively generates raw actions. 
RT-1 demonstrates strong performance on seen tasks, generalizes well to unseen tasks, and shows robustness to distractors and varying backgrounds. Its performance further improves with the incorporation of simulation data.
Moreover, incorporating data from different robotic platforms (Everyday Robots and Kuka) enables generalization across diverse embodiments. 
RT-2~\cite{Brohan2023RT2VM} further advances this with a more streamlined, end-to-end design that maximizes knowledge transfer from foundation models. It fine-tunes web-scale pretrained VLMs (PaLI-X~\cite{PaLI-X} and PaLM-E~\cite{PaLM-E}) into end-to-end VLAs (RT-2-PaLI-X and RT-2-PaLM-E), which directly output raw actions. 
The raw robot actions are discretized into action bins, enabling autoregressive inference in the same manner as VLM.
Importantly, this approach mitigates the need to modify the original architecture of the foundation models. 
By leveraging foundation VLM as backbones and co-training on both vision-language and robot action data, RT-2 exhibits enhanced reasoning and generalization capabilities. It demonstrates emergent abilities beyond its training data during test-time inference. Moreover, RT-2 with chain-of-thought reasoning can interpret and respond to complex commands, highlighting the significant advantages of using VLMs as the backbone for VLA models.
To enhance dataset scale and diversity for improved policy generalization, OXE~\cite{RT-X} introduces a unified dataset comprising over 1 million trajectories collected from 22 different robots. 
Empirical results of retraining RT-1/2 on this dataset show that cross-embodiment training leads to substantial performance gains, and model capacity plays a critical role in data-rich settings.

Although the RT-2 model has a significant impact, its training code and models have not been publicly released. To advance open-source VLA models, several initiatives have been made. RoboFlamingo~\cite{RoboFlamingo} builds upon the open-source VLM OpenFlamingo~\cite{awadalla2023openflamingo} and focuses on single-step vision-language comprehension. It models sequential history through an explicit policy head and is lightly fine-tuned via imitation learning on language-conditioned manipulation datasets. While RoboFlamingo focuses on cost-effective, open-source solutions, it has only been evaluated in the simulation benchmark CALVIN~\cite{mees2022calvin}. 

OpenVLA~\cite{kim2024openvla} is a widely recognized open-source, transformer-based VLA model that has been adopted as the backbone for numerous subsequent works. It carefully curates a high-quality OXE subset comprising 970K trajectories from heterogeneous robots, builds on the advanced Prismatic-7B model, and provides a well-structured, ready-to-use codebase. OpenVLA demonstrates strong zero-shot generalization after pretraining and shows that joint training on cross-embodiment datasets enables efficient robot-specific fine-tuning, requiring only 10 to 150 trajectories per robot to optimize performance. This paradigm of joint pretraining followed by lightweight adaptation has gained significant traction. Additionally, OpenVLA explores parameter-efficient fine-tuning and memory-efficient inference through quantization. By open-sourcing both its model and dataset, OpenVLA has catalyzed community-scale development and inspired a wave of follow-up research.
MoManipVLA~\cite{MoManipVLA} proposes a bi-level optimization framework to adapt the stationary-based OpenVLA~\cite{kim2024openvla} for mobile manipulation tasks. It employs OpenVLA to generate the 3D position, posture, and state of the gripper $(\Delta x, \Delta \theta, \Delta G)$, using a downstream search algorithm to solve for coordinated motion between the manipulator and the base. This transition requires minimal fine-tuning (200 simulation demonstrations and 50 real-world episodes), highlighting the generalization capability of VLAs across embodiments. To improve training and inference speed, boost task success rates, and identify optimal fine-tuning strategies, OpenVLA-OFT~\cite{openvla-oft} conducts a comprehensive empirical analysis of fine-tuning recipes and proposes integrating parallel decoding, action chunking, and continuous action representations into the fine-tuning process.
TinyVLA~\cite{wen2025tinyvlafastdataefficientvisionlanguageaction} aims to develop a more efficient model for both pretraining and inference. It first pretrains a VLM to initialize the policy backbone, then freezes the pretrained components and applies the parameter-efficient fine-tuning method LoRA~\cite{hu2022lora}. A diffusion-based action decoder head is appended to the pretrained VLM via a linear projection layer.
To reduce inference latency caused by large model size and better handle dynamic tasks, HiRT~\cite{zhang2024hirt} further introduces a hierarchical robot transformer framework that enhances execution frequency and performance. In this design, the VLM (InstructBLIP~\cite{dai2023instructblipgeneralpurposevisionlanguagemodels}) encodes language-image inputs into latent representations at a lower frequency. A lightweight policy head then asynchronously generates low-level actions by conditioning this latent representation along with real-time observations.

\subsection{Video Pretraining and Robot Data Fine-Tuning}
Another line of research has explored large-scale video generative pretraining to capture world dynamics and facilitate robot learning. 
For example, GR-1~\cite{wu2023unleashinglargescalevideogenerative} adopts a GPT-style transformer model that learns to predict future frames during video pretraining and is subsequently fine-tuned on robot datasets to incorporate action generation. Experimental results on the CALVIN simulation benchmark and a real robot demonstrate the effectiveness of video-based pretraining. Its successor GR-2~\cite{cheang2024gr2generativevideolanguageactionmodel} scales up this approach by pretraining on a much larger dataset (38 million text-video pairs compared to GR-1’s 0.8 million clips) and replaces the MLP action head with a CVAE. The model learns to capture essential world dynamics and semantic information from videos, which are critical for downstream policy learning. GR-2’s video generation capability effectively serves as a planner for action generation, with the generated videos closely aligning with real-world rollouts.

\subsection{Diffusion-Based Action Chunking}
Although transformer-based autoregressive models have led to significant advancements, several limitations remain. First, discrete autoregressive tokenization can struggle to represent continuous or multi-modal actions, which are especially crucial for dexterous tasks. Additionally, the standard autoregressive generation process produces one action at a time, limiting action inference frequency. To address these issues, a new class of VLA models has emerged as an alternative to pure GPT-style architectures: using diffusion-based action heads with action chunking~\cite{ActionChunking}. Diffusion policies have demonstrated superior ability to model multi-modal action distributions~\cite{Chi2023DiffusionPV}, while action chunking allows the model to output sequential actions simultaneously. This approach improves temporal consistency, reduces compounding error, and significantly boosts control frequency.

Octo~\cite{Octo} is an early work that introduces a transformer-based policy with a diffusion head, trained on a subset of 25 datasets from the OXE. The model processes images with CNN and ViT, while language is handled by a frozen T5 model. The block-wise attention structure of the transformer allows for the addition or removal of inputs and outputs during fine-tuning, enabling adaptation to cross-embodiment action and observation spaces. This design enhances flexibility in input sources and fine-tuning. 
A more recent and influential advancement is $\pi_0$~\cite{black2410pi0}, which combines flow matching~\cite{lipmanflow} with action chunking to improve policy. The VLM backbone of $\pi_0$ is initialized from PaliGemma~\cite{beyer2024paligemma}. The model is pretrained on a mixture of OXE Magic Soup and the $\pi$ dataset, covering a wide range of scenes, corrective behaviors, and recovery strategies. In the post-training phase, $\pi_0$ is fine-tuned on a smaller, task-specific dataset to adapt to particular downstream tasks. The results show that comprehensive pretraining enables strong zero-shot generalization, while requiring only minimal fine-tuning data to achieve high performance in complex, multi-stage tasks such as laundry folding, box building and egg packing. Additionally, $\pi_0$ supports a control frequency of up to 50 Hz—an order of magnitude improvement over RT-2’s 5 Hz.

RDT~\cite{liu2024rdt} further extends diffusion-based VLA models to bimanual manipulation, demonstrating impressive few-shot learning capabilities. It employs frozen SigLIP~\cite{zhai2023sigmoid} and T5-XXL~\cite{raffel2020exploring} for image and language encoding, and scales the DiT head to 1B parameters. RDT can acquire new skills from as few as 1–5 demonstrations, marking a significant step toward highly data-efficient learning in complex robotic tasks. 
CogACT~\cite{li2024cogact} adds a diffusion-based action head to OpenVLA and introduces an ensemble strategy to mitigate inter-chunk mode shifts by aggregating chunked sequences.
HybridVLA~\cite{HybridVLA} integrates autoregressive and diffusion policies into a unified VLA model.

\subsection{Heterogeneous Datasets and Unified Action Space}
GR00T N1~\cite{nvidia2025gr00tn1openfoundation} introduces the data pyramid to enhance data diversity and quantity for training robot foundation models. The pyramid comprises large-scale web and human video data, mid-scale synthetic simulation data, and small-scale real-world data. It leverages the entire pyramid by extracting latent actions from human videos and synthetic demonstrations generated via DexMimicGen~\cite{jiang2024dexmimicgen}, combined with real-world data for training. 
GR00T N1 adopts a hierarchical architecture, where the high-level model is a slow (10 Hz) autoregressive VLM (Eagle-2, 1.34B parameters) responsible for high-level contextual reasoning and planning from visual and language inputs. The low-level model is a fast (120 Hz) diffusion transformer (0.86B parameters) dedicated to real-time motor control, generating smooth and responsive actions. The two models are tightly integrated and jointly trained end-to-end. 
To better leverage cross-embodiment datasets, UniAct~\cite{UniAct} learns a universal action space compatible across diverse embodiments, represented by vector-quantized codes, where each code encodes common atomic behaviors shared across different robots.

\subsection{Recent Advancements}
Despite advances in diffusion-based action heads with action chunking, the inference latency problem remains, as the model requires time to generate the next action chunk. 
If the robot continues executing the previous chunk while the next chunk is still being inferred, the new action chunk will be based on outdated observations, lacking real-time environmental feedback. Additionally, multiple plausible action modes in the diffusion process may exist at chunk boundaries, and mode shifts can lead to discontinuities between chunks, resulting in jerky or out-of-distribution movements.
Real-Time Chunking~\cite{black2025real} shows that simple averaging-based smoothing strategies can actually degrade performance, producing trajectories worse than those of individual chunks. Instead, it frames chunk fusion as inference-time inpainting via flow matching and introduces soft masking to improve cross-chunk continuity. During inference, the model generates the next action chunk while executing the current one, freezing actions that are guaranteed to be executed and inpainting the remaining steps. Soft masking ensures that the rest part of the chunk is still considered during generation, further improving continuity across chunks.

$\pi_0$-FAST~\cite{Pi0-FAST}, an extension of $\pi_0$, shows that the naive binning tokenization method would produce poor results due to strong correlation between consecutive tokens at high frequencies. To address this, it applies a discrete cosine transform (DCT) to encode action chunks. The DCT-based representation provides substantial token compression (up to $13.2\times$) across tasks while producing smoother action trajectories---both critical for high-precision manipulation.

Another limitation in prior work is that VLM pretraining alone does not yield representations fully aligned with robotic tasks, and naive fine-tuning with action supervision can degrade previously learned knowledge. To address this, $\pi_{0.5}$ with knowledge insulation~\cite{driess2025knowledge} proposes pretraining the VLM backbone on discretized actions and general vision-language data to develop robust, transferable representations. The action expert is trained separately using flow matching on continuous actions. To preserve the backbone’s pretrained knowledge, gradients from the action expert are blocked from flowing back, effectively insulating its representations. During inference, the lightweight action expert generates continuous actions, while the frozen backbone contributes broad visual-linguistic understanding derived from diverse pretraining data.

\subsection{Conclusions and Discussions}

To sum up, raw actions serve as the most direct and executable form of action representation, making them a natural choice for VLA models. This approach typically involves minimal prior human knowledge and less structural constraint, favoring end-to-end learning. As real-world data are collected in raw action format, it also requires minimal action token annotation. In line with the ``bitter lesson'' observed in LLM development, which emphasizes the power of scaling over manual engineering, raw action-based end-to-end VLA models are likely to evolve as base models grow stronger and larger datasets become available.

Indeed, the evolution of VLA models using raw action tokens reflects broader trends in the foundation model era---scaling up both data and model size, improving base model architectures, and transitioning from pure pretraining to post-training strategies. Recent works such as $\pi_0$, RDT, and GR00T N1 demonstrate that comprehensive pretraining enables both strong zero-shot generalization and efficient task-specific fine-tuning. This progression mirrors the development trajectory of LLMs. 

However, raw action data lacks the internet-scale accessibility of language data. It is costly to collect, often requiring teleoperation and real robot interaction, which limits scalability. Furthermore, raw actions cannot directly generalize across embodiments, and fine-tuning or post-training on downstream tasks can lead to catastrophic forgetting of pretrained visual-language knowledge. Also, directly generating raw control commands without intermediate representations is less practical for long-term control tasks, as the required context length, computational cost, and inference latency can become prohibitively high.
Addressing these challenges while preserving foundation model knowledge remains a critical direction for future research.

\section{Reasoning as Action Tokens}

\label{sec:reasoning}

Embodied tasks, such as robotic manipulation and autonomous driving, often demand sophisticated cognitive abilities from AI agents. Their inherent complexity stems from the need for long-horizon reasoning, a deep understanding of space, semantics, and common sense, and the ability to operate effectively within dynamic environments~\citep{zhang2024vlabench}. Even advanced foundation models face considerable challenges in these areas. While a single VLA model is anticipated to address a wide array of embodied tasks, simply scaling up model parameters often proves insufficient to tackle the inherent complexities of real-world scenarios, particularly those demanding robust logical and embodied reasoning. Equipping VLAs with enhanced reasoning capabilities thus emerges as a promising solution. ~\Cref{tab:reasoning_action} summarizes  representative works that explicitly use reasoning as action tokens.

In the context of VLA, reasoning refers to a deliberative thinking process that is explicitly externalized in the form of natural language and serves to enhance the generation of the target action token. Unlike other action tokens that directly represent physical movements or emphasize object interactions, these reasoning tokens serve an intermediary role, facilitating the generation of subsequent executable action tokens. This concept allows the model to ``think step-by-step'' and externalize its internal decision-making process. For instance, RAD~\cite{clark2025action} uses reasoning to produce raw actions informed by language plans, and DriveVLM~\cite{tian2024drivevlm} processes reasoning before generating vehicle motion trajectories. 

\subsection{Evolution of Reasoning in VLA Models}
The core idea of externalizing internal reasoning processes finds its roots in Chain of Thought (CoT) prompting~\citep{wei2022chain}. Originally developed for LLMs to articulate intermediate steps (e.g., via the prompt ``think step by step'') before a final output, CoT has since transcended text-only domains. Its extension into visual and multi-modal contexts laid the groundwork for how reasoning could function within VLA models. For example, CoT has been applied to generate visual intermediates, such as bounding boxes of target objects, before computing final actions in visual tasks~\citep{hao2024visualcot}.

Early pioneering works in embodied reasoning often leveraged LLMs, augmented with additional modules to interpret visual scenes. A notable example is Inner Monologue~\cite{huang2022inner}, which uses LLMs to accept human instructions, scene descriptions (generated by MDETR~\cite{kamath2021mdetr}), and action feedback (from leveraged perception models). This setup allows for recursive multi-step language planning until a task is successfully completed. 

However, the field has rapidly evolved. Today, the mainstream approach for VLA models integrating reasoning is to utilize VLM. VLMs possess inherent and proficient multi-modal prior knowledge, simplifying the model architecture by reducing the need for numerous additional modules. Their intrinsic ability to process both linguistic and visual modalities significantly enhances the reasoning process for complex embodied tasks. To tailor these VLMs to specific reasoning patterns crucial for embodied tasks, methods like fine-tuning or retraining models are commonly employed, as demonstrated by Embodied CoT (ECoT)~\cite{ECoT} and RAD~\cite{clark2025action}. 

\input{table/reasoning}

\subsection{Key Implementations and Applications}

ECoT~\cite{ECoT} serves as a typical example of adopting reasoning for embodied tasks. Built on Prismatic VLMs~\cite{karamcheti2024prismatic}, an OpenVLA model~\cite{kim2024openvla} is specifically trained with reasoning data. A significant challenge in this domain is obtaining high-quality, large-scale reasoning datasets. While human annotation yields superior quality, it is impractical at scale. ECoT introduces an automated data synthesis pipeline that structures reasoning into a fixed sequence, from task decomposition to gripper position and object box prediction.

Following ECoT, RAD~\cite{clark2025action} adopts a similar framework but substantially extends data collection. It not only synthesizes reasoning data automatically from robot trajectories but also from easily accessible action-free human videos. The synthesis from human videos mirrors that from robot data, replacing motion primitive extraction with HaMeR~\cite{pavlakos2024reconstructing}, a method for hand keypoint and pose tracking. This innovation facilitates co-training on both robot data and human videos, broadening the scope of available data.

Furthermore, some VLMs are specifically trained for embodied reasoning, such as Cosmos-Reason1~\cite{nvidia2025cosmos}. This model is trained via reinforcement learning (specifically, GRPO) and supervised fine-tuning (SFT) on physical common sense, embodied reasoning, and intuitive physics, tailoring it for embodied applications.

Beyond robotic manipulation, autonomous driving presents another critical application area for reasoning due to its highly complex, dynamic, interactive environment, and the paramount need for enhanced safety. DriveVLM~\cite{tian2024drivevlm} applies CoT across its three key modules: scene description, scene analysis, and hierarchical planning. The scene description module identifies critical objects in the driving environment. The scene analysis module then evaluates their characteristics and potential influences on the ego vehicle. Finally, the hierarchical planning module formulates step-by-step plans, progressing from language motions to decision descriptions and ultimately to waypoints. This demanding task requires sophisticated deduction and common-sense understanding of diverse objects and scenarios, making reasoning with VLMs particularly well-suited. Models like AlphaDrive~\cite{bo2025alpha}, trained with SFT warm-up followed by GRPO-based RL exploration, are examples of VLMs exclusively developed for reasoning in autonomous driving contexts.

\subsection{Advantages of Reasoning as Action Tokens}
Integrating reasoning as action tokens offers several compelling advantages for VLA models: 
\noindent \textbf{Bridging the Instruction-Action Gap and Enhanced Generalization}---Reasoning significantly mitigates the gap between high-level instructions and low-level executable actions by introducing intermediate thinking steps. This allows VLMs to leverage their prior knowledge to handle tasks involving various scenes and objects to enhance the generalization ability and performance in complex long-horizon tasks. For instance, ECoT~\cite{ECoT} demonstrates substantial performance improvements in complex manipulation tasks like ``put the edible object in the bowl''. This task requires sophisticated reasoning, including identifying the bowl, checking all existing objects, and selecting edible ones based on common sense. ECoT also shows enhanced generalization capabilities to unseen objects and scenes, demonstrating the power of reasoning. 
\noindent \textbf{Improved Interpretability and Human-in-the-Loop Capabilities}---By externalizing the agent's thought process, reasoning enhances the interpretability of the model. Humans can clearly review the agent's decisions, trace failures to specific points in the reasoning chain, and even intervene in real-time when errors are detected. This transparency also facilitates human-in-the-loop interactions, allowing flexible processing of uncertain human input for subsequent actions, as exemplified by Inner Monologue's ability to let humans select objects in real-time. 
\noindent \textbf{Enabling Cross-Embodiment Capability}---While different embodiments may have distinct architectures and action token formats, the high-level plan for completing a task often remains consistent. Reasoning can extract these abstract plans, shifting the primary challenge to projecting them into final executions. The rich prior knowledge of VLMs, combined with training or fine-tuning on cross-embodiment datasets like OXE~\cite{RT-X}, can facilitate reasoning across various embodiments. ECoT validates its cross-embodiment capability, showing that a fine-tuned model can perform ECoT reasoning effectively in new embodiments.

\subsection{Limitations and Future Directions}
Despite its numerous advantages, applying reasoning in embodied tasks still faces several limitations: \textbf{Increased Inference Time and Reduced Execution Speed.} Reasoning often necessitates the model to generate a lengthy thinking process or multiple reasoning steps, leading to high inference time and low execution speed. This is a critical constraint for real-time, high-frequency tasks common in embodied AI. While solutions like ECoT's asynchronous execution can speed up inference by around $40\%$, further acceleration techniques are crucial. 
\textbf{Fixed Reasoning Steps and Data Challenges.} In current implementations, reasoning steps are frequently fixed manually. While this provides stability for certain tasks, it can limit the model's generalization capability and hinder the exploration of potentially superior reasoning pathways. Furthermore, building high-quality, large-scale reasoning datasets remains costly and challenging.

Concluding from the advantages and limitations, reasoning is particularly well-suited for complicated, long-horizon, deductive tasks that require decomposition into multiple subtasks, especially those with relatively low-frequency execution demands due to current inference speed limitations.

Looking ahead, future work in this area promises exciting advancements:
\textbf{Improved Inference Speed and Foundation Model Capabilities.} Enhancements in the inference speed and inherent reasoning capabilities of foundation models are anticipated. \textbf{Better Data Collection Methods.} Developing more efficient and scalable methods for collecting high-quality reasoning data is essential. 
\textbf{Advanced Test-Time Compute.} Leveraging test-time compute, which involves additional computation during inference, holds potential for boosting the performance, generalization, and robustness of reasoning models. Techniques explored in AlphaDrive and Cosmos-Reason1 are just early examples. 
\textbf{Novel Reasoning Paradigm Design.} Insights into designing the paradigm of the reasoning module in VLA are eagerly awaited. This may include multimodal forms of reasoning and eventually generalize to a wider range, perhaps all, of embodied tasks and robot embodiments.

\section{Scalable Data Sources}
\label{sec:data}

The development of VLA models critically relies on learning action tokens that are grounded in multimodal observations, compositional to support skill sequencing, and executable for embodied policy control. Effective learning of such representations requires data that jointly provides visual-linguistic grounding, fine-grained action supervision, and embodiment-aligned sensorimotor control. However, individual data sources typically provide supervision signals with complementary strengths. To address this, modern VLA frameworks adopt a hierarchical multi-source data paradigm, integrating web data and human video for visual-linguistic grounding, synthetic and simulation data for skill composition, and real-world embodied data for embodiment-specific control grounding. As these three types of data sources decrease in quantity and grow in embodiment specificity, they constitute the bottom, middle, and top layers of the “Data Pyramid”~\cite{nvidia2025gr00tn1openfoundation}. This multi-layered supervision enables scalable and transferable action token learning across diverse tasks, embodiments, and control modalities.
~\Cref{tab:vla-datasets} showcases representative works for scalable data sources.

\input{table/data_sources}

\subsection{Bottom Layer: Web Data and Human Videos}
The bottom layer consists of large-scale web data and human video datasets to support visual-linguistic grounding~\cite{ravi2024sam, miech2019howto100m}, world modeling~\cite{MendoncaBP23, assran2025v}, and temporal prediction~\cite{Villar-Corrales_2022_BMVC}. Since web data primarily consists of vision-language pairs and is mainly used to enhance foundation model capabilities, which falls outside the scope of our survey, we focus primarily on human video in our discussion. Representative human video datasets include Ego4D~\cite{Ego4D2022CVPR}, EPIC-KITCHENS-100~\cite{Damen2021RESCALING}, and Something-Something V2~\cite{goyal2017something}. Although these datasets do not contain action labels directly usable for policy learning, they capture diverse human-object interactions, complex manipulation skills, and rich physical common sense, which are valuable sources of world knowledge. Their scale and diversity enable pretraining of temporal visual encoders and facilitate the learning of action token representations. Recent VLA models have utilized these datasets to extract trajectory~\cite{yang2025magma, niu2025pre}, infer latent state transitions~\cite{ye2024latent}, and generate latent action~\cite{nvidia2025gr00tn1openfoundation}. The resulting pretrained modules provide temporally grounded, semantically structured, and partially embodied priors that enhance downstream policy learning across tasks and embodiments.

In addition to explicit perceptual content, these videos implicitly encode mappings between visual observations and physical actions. This implicit structure allows models to acquire coarse affordance priors and latent dynamics conditioned on observed states and estimated actions~\cite{wu2023unleashinglargescalevideogenerative, cheang2024gr2generativevideolanguageactionmodel, mendonca2023structured}. Egocentric viewpoints reduce the embodiment gap by approximating robot perspectives, particularly for manipulation and navigation tasks. Recent works further exploit weak supervision to extract actionable representations from large-scale videos. Frame-level captioning and temporal alignment provide indirect supervision signals for generating trajectory-based and goal state action tokens. For instance, Magma~\cite{yang2025magma} introduces Set-of-Mark and Trace-of-Mark abstractions to anchor action grounding within video streams. Ego-Exo4D~\cite{grauman2024ego} augments egocentric data with third-person views for 3D motion grounding, facilitating embodiment transfer. These approaches enable VLA models to build temporal grounding and language-conditioned policy priors in open-world settings.

\subsection{Middle Layer: Synthetic and Simulation Data}

To provide a critical bridge between the human video and the high cost of real-world data collection, VLA research extensively utilizes simulation and synthetic data. This paradigm provides scalable access to structured, task-centric data crucial for learning compositional skills and robust control policies. Two complementary methodologies are central to this approach.

\textbf{Synthetic Dataset Generation.} The first methodology is offline synthetic data generation. It employs procedural generation pipelines like MimicGen~\cite{mandlekar2023mimicgen}, DexMimicGen~\cite{jiang2024dexmimicen}, and RoboCasa~\cite{nasiriany2024robocasa} to programmatically augment or synthesize large-scale datasets from a limited set of expert demonstrations. For instance, MimicGen~\cite{mandlekar2023mimicgen} establishes a paradigm of applying procedural variations, such as spatial transformations and scene reconfigurations, to existing trajectories to enhance data diversity. Building on this, RoboCasa~\cite{nasiriany2024robocasa} scales this process to generate over 100K trajectories for diverse manipulation tasks, while DexMimicGen~\cite{jiang2024dexmimicen} extends it to complex bimanual manipulation by combining kinematic retargeting with contact dynamics randomization. These methods substantially enrich the quantity and diversity of datasets at low costs, as demonstrated by GR00T N1, which leverages such data to train policies for complex bimanual assembly tasks~\cite{nvidia2025gr00tn1openfoundation}.

\textbf{Interactive Simulation Platforms.} Complementary to synthetic datasets, the second methodology involves interactive simulation platforms like robosuite~\cite{zhu2020robosuite}, Habitat~\cite{savva2019habitat}, Isaac Gym~\cite{S2021_28dd2c79}, Isaac Lab~\cite{mittal2023orbit}, and others~\cite{ehsani2023spoc, yenamandra2023homerobot, mees2022calvin, RLBench, Puig_2018_CVPR, li2024evaluating, LIBERO, cheng2024navila, FMB, walke2024bridgedatav2datasetrobot, yu2020meta, iTHOR, pmlr-v100-gupta20a, srivastava2022behavior, Puig_2018_CVPR, Shah2017AirSimHV, zeng2021transporter, jiang2022vima, ebert2021bridgedataboostinggeneralization, li2024simpler, beattie2016deepmindlab, bellemare2013arcade, lee2021beyond, ramrakhya2022habitat, lin2023mcu, nasiriany2024robocasa, lynch2023interactive}. Data generation within these simulators follows several key paradigms. First, it involves teleoperation, where a human operator uses VR controllers, keyboards, or other interfaces to control the simulated robot and perform tasks. The second method involves algorithmic solvers, such as classical motion planners, which generate successful trajectories for tasks that have clear solution paths. Third, learned policies, often trained via RL, can autonomously collect vast amounts of data. Beyond generating robot trajectories, these platforms also diversify the training environment itself. Procedural content generation systematically randomizes environmental factors, including objects, textures, and lighting conditions. Additionally, platforms like AgiBot Digital World~\cite{contributors2025agibotdigitalworld} integrate realistic 3D assets with high-fidelity physics simulation, facilitating exploration of rare, failure-prone, and complex interaction scenarios. These environments enable agents to learn through direct interaction with a physics-based world, facilitating large-scale reinforcement learning and imitation learning. Simulation is particularly valuable for high-risk or safety-critical scenarios, such as tool misuse or complex contact dynamics, which are essential for robust policies with recovery capabilities. However, addressing the persistent \textit{sim-to-real gap} remains essential, where discrepancies in visual fidelity and physics modeling necessitate further fine-tuning in real-world settings.

\subsection{Top Layer: Real-World Robot Data}

Real-world robot data contains the most critical resource for training VLA models, providing direct supervision for learning physically grounded and executable policies. Unlike simulation or human video, real robot datasets capture the complex dynamics, sensory noise, and unpredictable variations inherent in physical environments. This high-fidelity information is indispensable for bridging the \textit{sim-to-real gap} and instilling crucial embodiment-specific characteristics, such as kinematic constraints and contact dynamics. Consequently, real-world data is paramount for training policies to generate low-level action, which demands precise physical realism for successful execution.

A primary objective in VLA research is to develop generalist agents capable of operating across diverse robotic platforms. This motivates the curation of large-scale, multi-embodiment datasets that aggregate experiences from various robot morphologies and environments. For instance, OXE~\cite{RT-X} assembles over 1 million manipulation episodes from multiple datasets spanning 22 robots, facilitating the learning of policies with significant cross-embodiment transfer capabilities. Recent efforts have further enriched these collections. RoboMIND~\cite{wu2024robomind}, for example, uniquely incorporates negative data, providing 5K failure trajectories annotated with causal reasons to enable more robust policy learning through contrastive or corrective mechanisms. Similarly, RH20T~\cite{fang2023rh20t} further provides multi-modal information, including force-torque and audio data, to support policies that reason about physical contact and environmental sounds.

In contrast to the broad coverage of multi-embodiment datasets, single-embodiment and task-specific datasets provide complementary data for mastering complex, specialized skills. These datasets are crucial for learning fine-grained manipulation and long-horizon tasks. 
For instance, RT-1~\cite{Brohan2022RT1RT} represents one of the earliest and most well-known efforts to collect large-scale single-embodiment datasets.
Subsequently, DROID~\cite{khazatsky2025droidlargescaleinthewildrobot} introduces a unified robotic platform, deploys it across multiple institutions worldwide, where researchers jointly collect a large-scale dataset that spans a wide range of tasks, objects, scenes, viewpoints, and interaction locations. Such unified yet diverse data facilitates the training of generalizable VLA models~\cite{Pi0-FAST}.
In addition, AgiBot World~\cite{contributors2024agibotworldrepo} provides 1M episodes with the Genie-1 robot across 5 different domains. 
For long-horizon planning, BridgeData V2~\cite{walke2024bridgedatav2datasetrobot} contains 60K demonstrations of complex kitchen tasks, providing supervision for learning causal dependencies in multi-step manipulation. 
Datasets like HoNY~\cite{shafiullah2023bringing} focus on capturing data in unstructured ``in-the-wild'' home environments, presenting challenges such as object clutter and variable lighting. This principle of data collection extends to autonomous driving, where datasets like nuScenes~\cite{fong2021panoptic} and the Waymo Open Dataset-Motion~\cite{Ettinger_2021_ICCV, Kan_2024_icra} integrate rich sensor suites (e.g., LiDAR, RADAR) to train safety-critical driving policies, typically using trajectory tokens as the primary action representation. 

Despite their indispensability, the acquisition of real-world robot data remains a significant bottleneck due to high costs, operational complexity, and the slow pace of teleoperation or portable motion capture devices. This scalability challenge fundamentally shapes the data strategy for most state-of-the-art VLA models. A prevalent and effective paradigm involves large-scale pretraining on abundant simulation or web-scraped data to learn generalizable visual, linguistic, and semantic representations. Subsequently, models are fine-tuned on smaller, high-quality, real-world datasets to adapt these general representations to specific physical embodiments and task requirements. This hierarchical approach strategically balances the need for broad world knowledge with the precise physical grounding required for reliable real-world execution, effectively mitigating the data scarcity problem while maximizing performance.

\section{General Discussions and Future Directions}
\label{sec:discussion-future-direction}

The preceding sections reveal that each category of action tokens has been explored through a series of influential papers. These studies have uncovered the expressive capacity of different action tokens, effectively leveraged the strengths of foundation models, and developed scalable data strategies, culminating in VLA models that demonstrate promising empirical performance. Clearly, each type of action token comes with its own strengths and limitations and remains in an early stage of exploration, holding significant potential for future development. At present, no single type exhibits absolute dominance or clear inferiority, nor has the research community converged on a dominant action token paradigm, making definitive recommendations challenging. As such, we provide our assessment of future trends in action tokens and the development of VLA models in \Cref{subsec:trends}. In \Cref{subsec:vla-agents,subsec:rl,subsec:full-dexterity,subsec:safety,subsec:data-scalability}, we further present a set of general observations and reflections, identifying underexplored areas in VLA research to inform and guide future directions.

\subsection{Trends of Action Tokens and VLA Models}
\label{subsec:trends}

Based on the summarized advantages and limitations of each token in \Cref{tab:action_token}, we observe that different action tokens exhibit complementary strengths and are best suited to different levels within a VLA model. This suggests that \emph{the future of VLA lies not in a single dominant token type, but in their strategic combination}, motivating a \emph{hierarchical} architecture. Language plans and code offer unique advantages in long-horizon planning and logical control—capabilities that are difficult to substitute with other action token types—making them ideal for the top layer. For subtasks derived from these high-level plans, a combination of 3D affordance, trajectory modeling, and goal video prediction can provide precise and interpretable motion representations, making them well-suited for the intermediate layer. In contrast, language motion and API-based code are comparatively less expressive and can generally be replaced by the former three. Finally, a policy module can be trained to map these vision-based representations into raw actions.

While latent representations hold strong potential, we do not include them in our proposed architecture due to current training challenges—particularly in achieving appropriate granularity, semantic comprehensiveness, and task-centric alignment. These limitations are not easily addressed and may compromise reliability in practical applications. As such, we currently favor more explicit forms of action tokens, which are generally easier to train and inspect, and offer greater interpretability and control. Nonetheless, we remain optimistic about future advances in latent representations and their eventual integration as the field evolves.

An end-to-end low-level policy that directly maps subtasks to raw actions offers fundamental scalability, though it remains constrained by limited data availability. In the short term, the aforementioned hierarchical design facilitates data collection to achieve a data flywheel effect; in the long run, it could enable fully end-to-end controller learning that bypasses intermediate tokens and directly predicts raw actions from subtasks.

Reasoning features a crucial action token in VLA models. While reasoning has been incorporated into current VLA models, it is generally rudimentary and applied only to relatively simple tasks. As discussed in \Cref{sec:next-frontier}, action tokens in VLA models play a role analogous to that of language tokens in LLMs. It is therefore natural to envision reasoning processes in VLA models being constructed not from language tokens, but from action tokens. This mirrors how humans solve complex tasks---not only through linguistic planning and reflection, but also by engaging in physical-world grounding and imagination. Furthermore, action-token-based reasoning should be designed to leverage test-time computation adaptively, adjusting its length according to task complexity, as is commonly done in language-based reasoning. Such reasoning should be integrated throughout the VLA hierarchy as needed to enhance the generation of all other action tokens, offering a promising path toward more general and human-like intelligence.

The above analysis presents our perspective on the future development of VLA models from an action tokenization perspective. Fundamentally, the existence of current action tokens stems from the capabilities of foundation models to generate and interpret them. As foundation models continue to evolve and new modalities (e.g., audio, tactile) become increasingly accessible, we anticipate the emergence of new types and subtypes of action tokens that will further expand the expressiveness and effectiveness of VLA models. Continued investigation and thoughtful integration of all action tokens will be essential to fully harness their complementary strengths and advance toward more capable, general-purpose embodied intelligence.

\subsection{From VLA Models to VLA Agents}
\label{subsec:vla-agents}

A natural next step is to consciously evolve beyond VLA models toward VLA \emph{agents}, complementing core capabilities through an agent-centric paradigm. While current VLA models primarily focus on learning an effective mapping from vision-language inputs to action outputs, building more general and robust embodied intelligence likely necessitates agent-level systems with comprehensive and integrated functionalities. Most existing VLA models lack mechanisms for incorporating history. Even when present, such context is typically limited to a few frames or simple language-based planning. This is insufficient for long-horizon tasks in the real world, particularly those involving progress tracking, subtask dependencies, or online exploration. Addressing these challenges calls for robust and structured mechanisms for memory, planning, and reflection---components that have been extensively studied in the broader agent research community~\cite{wang2024survey} and can be effectively integrated into VLA. Preliminary efforts such as RoboOS~\cite{tan2025roboos} represent an early step in this direction, though the current design remains relatively simple. Additionally, planning and online exploration in VLA agents could also be substantially enhanced by integrating advances in world models. 

While existing research can generally be described within our proposed framework of a chain of interleaved VLA modules and action tokens, future agent systems should not be constrained to linear architectures. Instead, modules and generated action tokens should be adaptively invoked and managed by the agent to fully process information and generate effective outputs.

Finally, the evolution toward VLA agents---and the broader vision of deploying embodied agents in real-world environments---also calls for increased attention to multi-agent systems and human-agent co-existence, both of which are essential for the future integration of robots into everyday human life.

\subsection{From Imitation Learning to Reinforcement Learning}
\label{subsec:rl}


Our third observation centers on the training paradigm of VLA models. Currently, the vast majority of VLA models are trained using imitation learning, which presents several limitations. These include an inherent upper bound imposed by the capabilities of human demonstrators, a lack of goal-conditioned execution mechanisms, and difficulties in achieving consistent, near-perfect performance. Even worse, human demonstrations are often suboptimal and may lack dexterity due to factors such as fatigue, inattentiveness, individual idiosyncrasies, and the technical constraints of data collection devices—such as sensor imprecision and latency. These limitations of imitation learning naturally prompt reflection: much of human learning does not arise merely from observation or instruction, but instead depends fundamentally on hands-on trial-and-error and self-guided exploration. This suggests a promising direction for future research: applying reinforcement learning (RL) to optimize VLA models. By enabling models to learn directly from goal feedback and autonomously explore the environment, such approaches could yield more robust, dexterous, and high-success-rate behaviors. Reinforcement learning thus offers a pathway toward more human-like learning processes and capabilities in VLA models.

While the promise of RL for VLA models is clear~\cite{hu2024flare, lu2025vla, chen2025conrft, liu2025can, li2025simplevlarl, guo2025improving}, its direct application to real-world scenarios presents significant challenges. Deploying VLA models in the physical world often incurs a high reset cost, demanding substantial time and resources to reset the environment after each trial. Furthermore, the low interaction efficiency of real-world environments means that models require a vast number of interactions to learn effectively, which is often impractical. Safety concerns also loom large, as exploratory actions during RL training could lead to damage to the robot or its surroundings. These challenges highlight the critical need for developing more efficient RL algorithms that can enable VLA models to be grounded on real machines with minimal interaction. This could involve techniques such as in-context reinforcement learning~\citep{moeini2025survey}, which leverages large, pre-trained models to learn new tasks with limited data by adapting to new contexts.

Another crucial area for future research lies in automating the design of dense reward functions by leveraging existing VLMs~\citep{rlgpt, minedojo}. Crafting effective reward functions for complex robotic tasks is notoriously difficult and often requires significant manual effort and domain expertise. VLMs, with their impressive understanding of visual and textual information, hold the potential to interpret high-level task descriptions and automatically generate fine-grained, dense reward signals that guide the RL agent towards successful completion. This approach could significantly reduce the burden of reward engineering, accelerating the development and deployment of RL-driven VLA models in diverse real-world applications. 

\subsection{From Restrictive Hardware to Full Dexterity and Modalities}
\label{subsec:full-dexterity}

Another critical limitation of current VLA models lies in their underlying hardware configurations. While in daily life, most complex and fine-grained manipulation tasks are performed using human hands, the vast majority of existing VLA research relies solely on simple grippers, which severely restricts the action space and dexterity of manipulation. To advance toward more capable VLA models, future research must incorporate dexterous hands as a central component.

Moreover, existing work primarily focuses on three common modalities: vision, language, and action. However, such sensor configurations are insufficient for developing truly general-purpose agents. Broader sensory modalities---including tactile sense, audition, olfaction, and even gustation---are essential for enabling agents to handle a wider range of real-world tasks with the robustness and adaptability required for general intelligence.

\subsection{From Capability-Centric to Safety-Aware}
\label{subsec:safety}

VLA models must also place greater emphasis on safety considerations~\cite{zhang2025safevla}. Embodied intelligence not only inherits many of the alignment~\cite{ji2023ai} and safety challenges present in digital AI systems, but also introduces additional risks---such as physical damage to hardware and even potential harm to humans---due to its interaction with the real world. These high-stakes consequences demand that safety be treated as a first-class concern in algorithm design. However, this remains an underexplored area in current research and calls for more systematic investigation and proactive development of safety-aware methodologies.

\subsection{From Data Scarcity to Data Scalability}
\label{subsec:data-scalability}

The history of deep learning has repeatedly demonstrated that data serves as the fossil fuel powering the development of powerful models. However, current efforts in robot data collection have made it clear that in the near term data availability will remain insufficient in several critical dimensions.

First, the overall quantity of robot data is severely limited. In contrast to language and vision data, which benefit from massive and continuously expanding internet-scale corpora, robot data must be collected manually through labor-intensive processes. Despite significant efforts from the community, the quantity of available robot data still falls short of vision-language data by orders of magnitude and is unlikely to reach comparable levels in the short term. In fact, the total number of tokens in OXE datasets is estimated to be only about 1/200,000 of that in large-scale language model corpora, further highlighting the scarcity of robot data.

Second, robot data lacks sufficient modality coverage. Most existing datasets are limited to vision, language, and action, while other important sensory modalities such as tactile sense, audition, olfaction, and gustation remain largely unrepresented. These gaps are difficult to fill shortly due to hardware constraints.

Third, robot embodiments are diverse and often incompatible with each other. Although a large amount of data has been collected on different platforms, these datasets are fragmented across various embodiments and are not easily shared or reused, which further reduces the amount of usable data.

Fourth, the quality of robot data is often inadequate, especially in scenarios involving dexterous manipulation. Existing data collection devices for dexterous hands are not yet sufficiently advanced in terms of precision, responsiveness, and reliability. As a result, it remains difficult to acquire high-quality data for complex tasks. This challenge is even more pronounced for high-degree-of-freedom hands with force feedback.

Due to these limitations, VLA models, which may ultimately require far greater data volume than digital AI systems, face a significant bottleneck in terms of data availability. Future research should address these challenges along two key directions. On one hand, simulation and internet-scale resources should be better leveraged to provide scalable supervision. On the other hand, it’s crucial to develop more versatile, reliable, multimodal, and in-the-wild data collection systems~\cite{chi2024universal, xudexumi, wang2024dexcap} that can operate effectively in real-world environments. These efforts are essential to support the continued progress and scalability of VLA models.

\section{Conclusion}

This survey positions VLA models as a central pathway toward embodied AI and presents a comprehensive review of existing research from an action tokenization perspective. For each category of action tokens, we systematically examine representative VLAs, analyze their strengths and limitations, and highlight potential directions for future investigation. We further summarize major efforts in scalable data sources, aiming to inform and support ongoing research. Finally, grounded in the current state of VLA development, we outline future trends and underexplored areas to help guide the next stages of progress. As vision and language foundation models continue to thrive, research in VLA is gaining momentum and holds immense promise. We hope this survey helps clarify the field’s evolution, map out its trajectory, and contribute meaningfully to its growth---ultimately bringing us closer to the pursuit of Artificial General Intelligence.

\clearpage

\bibliography{references}

\end{document}

%% file: figure/scope.tex
\begin{tikzpicture}[scale=0.8]

    \definecolor{AGI}{RGB}{255, 204, 204}
    \definecolor{EmbodiedAI}{RGB}{137, 221, 246}
    \definecolor{DigitalAI}{RGB}{49, 246, 154}
    \definecolor{VLA}{RGB}{255, 119, 0}
    \definecolor{Hardware}{RGB}{204, 153, 255}
    \definecolor{Robotics}{RGB}{255, 247, 0}

    \filldraw[fill=AGI, fill opacity=0.3, draw=black] (0.0, 0.0) ellipse (3.5 and 2.5);
    \filldraw[fill=EmbodiedAI, fill opacity=0.3, draw=black] (0.5, 0.0) ellipse (2.5 and 2.0);
    \filldraw[fill=DigitalAI, fill opacity=0.3, draw=black] (-1.75, 0.0) ellipse (1.5 and 1.0);
    \filldraw[fill=VLA, fill opacity=0.3, draw=black, rotate=30] (0.0, 0.0) ellipse (1.25 and 1.25);
    \filldraw[fill=Hardware, fill opacity=0.3, draw=black] (1.5, 0.5) ellipse (1.25 and 0.7);
    \filldraw[fill=Robotics, fill opacity=0.3, draw=black] (1.5, -0.5) ellipse (1.25 and 0.7);

    \node at (-2.0, 1.5) {\scriptsize \textbf{AGI}};
    \node at (0.7, 1.55) {\scriptsize \textbf{\shortstack{Embodied\\AI}}};
    \node at (-2.6, 0.0) {\scriptsize \textbf{\shortstack{Digital\\AI}}};
    \node at (0.1, 0.0) {\scriptsize \textbf{VLA}};
    \node at (1.9, 0.6) {\scriptsize \textbf{Hardware}};
    \node at (1.9, -0.6) {\scriptsize \textbf{Robotics}};

\end{tikzpicture}

%% file: table/action_token.tex
\begin{table}[!t]
  \centering
  \caption{Overview of key advantages, limitations, and empirical results of each type of action token.}
  \label{tab:action_token}
  \renewcommand{\arraystretch}{1.3}
  \resizebox{\textwidth}{!}{%
    \begin{tabular}{
      >{\centering\arraybackslash}m{3.0cm}
      >{\centering\arraybackslash}m{3.0cm}
      >{\centering\arraybackslash}m{6.0cm}
      >{\centering\arraybackslash}m{6.0cm}
      >{\centering\arraybackslash}m{6.0cm}
    }
    \toprule
    \multicolumn{2}{c}
    {\large \textbf{Action Tokens}} &
    \large  \textbf{Advantages} &
    \large  \textbf{Limitations} &
    \large  \textbf{Notable Empirical Achievements} \\
    \midrule
    \multirow{2}{*}[-0.2cm]{\makecell{Language\\ Description\\ (\Cref{sec:language-description})}} 
    & Language Plan & \makecell{Well-supported by LLMs and VLMs;\\abundant co-training data;\\necessary for long-horizon planning} & \multirow{2}{*}[-0.32cm]{\makecell{Imperfect expressiveness\\(ambiguous; hard to \\describe dexterous\\ manipulation); high latency}} & \makecell{Make bed ($\pi_{0.5}$~\cite{intelligence2025pi05visionlanguageactionmodelopenworld}));\\make a sandwich (Hi Robot~\cite{shi2025hi})} \\ 
    \cmidrule(lr){2-3} \cmidrule(lr){5-5}
    & Language Motion & \makecell{Multi-task data sharing} & & \makecell{Pull napkin from dispenser (RT-H~\cite{rth2024arxiv})} \\ 
    \cmidrule(lr){1-5}
    \multicolumn{2}{c}{\makecell{Code \\ (\Cref{sec:code})}} 
    & \makecell{Well-supported \\by LLMs; clear logic \\for planning and control;\\rich third-party libraries} 
    & \makecell{Overly rely on predefined APIs;\\ brittle runtime execution} 
    & \makecell{Rearrange restore (Instruct2Act~\cite{Huang2023Instruct2ActMM})} \\ 
    \cmidrule(lr){1-5}
    \multirow{4}{*}[-0.8cm]{\makecell{Affordance \\ (\Cref{sec:affordance})}} 
    & Keypoint 
    & \makecell{Precise interaction targets} 
    & \multirow{4}{*}[-1cm]{\makecell{Future work should better \\capture 3D spatial structures; \\lacks temporal modeling of\\ evolving affordance prediction;\\ sensitive to visual noise, including\\ occlusion and motion artifacts}} 
    & \makecell{Pour tea (ReKep~\cite{huangrekep})} \\ 
    \cmidrule(lr){2-3} \cmidrule(lr){5-5}
    & Bounding Box 
    & \makecell{Well-supported by VLMs;\\ efficient for instance-level\\localization} & 
    & \makecell{Dexterous grasping in cluttered scenes\\ (DexGraspVLA~\cite{zhong2025dexgraspvla})} \\ 
    \cmidrule(lr){2-3} \cmidrule(lr){5-5}
    & Segmentation Mask 
    & \makecell{Capture fine-grained \\ contours, geometry for \\ functional region grounding} & 
    & \makecell{Decision-making in open world \\ (ROCKET-1~\cite{cai2024rocket})} \\ 
    \cmidrule(lr){2-3} \cmidrule(lr){5-5}
    & Affordance Map & \makecell{Dense, interaction-centric, full-scene} & 
    & \makecell{Deformable object manipulation\\ (ManiFoundation~\cite{xu2024manifoundation})} \\ 
    \cmidrule(lr){1-5}
    \multicolumn{2}{c}{\makecell{Trajectory \\ (\Cref{sec:trajectory})}} 
    & \makecell{Trainable from off-domain \\ human videos; cross-task \\ generalization} 
    & \makecell{Limited 3D expressiveness;\\ limited support from VLMs;\\ insufficient semantic grounding} 
    & \makecell{Clean the table with a duster\\(RT-Trajectory~\cite{rt-trajectory})} \\ 
    \cmidrule(lr){1-5}
    \multicolumn{2}{c}{\makecell{Goal State \\ (\Cref{sec:goal-state})}} 
    & \makecell{Well supported by \\foundation models;\\high data scalability via\\hindsight relabeling and\\action-free video utilization;\\task specificity} & \makecell{Challenging to \\generate high-quality, \\consistent goal states; \\high latency} 
    & \makecell{Transfer liquid using a pipette\\ (VPP~\cite{hu2025videopredictionpolicygeneralist})} \\ 
    \cmidrule(lr){1-5}
    \multicolumn{2}{c}{\makecell{Latent Representation \\ (\Cref{sec:latent})}} 
    & \makecell{High data scalability by\\utilizing action-free human videos\\and cross-embodiment data;\\strong expressive potential\\(compact structure,\\implicit semantics,\\multi-modal integration)} & \makecell{Uninterpretable;\\future work should\\improve the granularity,\\comprehensiveness, \\and task-centric alignment \\ of the latent space} 
    & \makecell{Fold shorts (GO-1~\cite{bu2025agibot});\\mine diamond in Minecraft\\ (OmniJARVIS~\cite{wang2024omnijarvis})} \\ 
    \cmidrule(lr){1-5}
    \multicolumn{2}{c}{\makecell{Raw Action \\ (\Cref{sec:raw-action})}} 
    & \makecell{Minimal human knowledge;\\minimal action token annotation;\\similar training strategy and\\scaling potential to VLMs;\\efficient fine-tuning} & \makecell{Data scarcity;\\difficulty in data collection;\\high latency;\\poor cross-embodiment\\generalization} 
    & \makecell{Laundry folding ($\pi_0$~\cite{black2410pi0});\\light a match and light a candle\\ (Real-Time Chunking~\cite{black2025real})} \\ 
    \cmidrule(lr){1-5}
    \multicolumn{2}{c}{\makecell{Reasoning \\ (\Cref{sec:reasoning})}} 
    & \makecell{Enhance generation\\of target action tokens;\\complex problem solving} & \makecell{High latency;\\future work\\should develop\\flexible reasoning paradigms} 
    & \makecell{Autonomous driving\\(DriveVLM~\cite{tian2024drivevlm})} \\ 
    \bottomrule
    \end{tabular}
  }
\end{table}

%% file: table/language.tex
\begin{table}[htbp]
\centering
\caption{Overview of VLA research using language description as action tokens.}
\label{tab:language_description}
\renewcommand{\arraystretch}{1.08}
\resizebox{\textwidth}{!}{%
\begin{tabular}{
>{\centering\arraybackslash}m{0.7cm} 
>{\centering\arraybackslash}m{2.5cm} 
>{\centering\arraybackslash}m{2.2cm} 
>{\centering\arraybackslash}m{3.1cm} 
>{\centering\arraybackslash}m{3.1cm} 
>{\centering\arraybackslash}m{2.7cm} 
>{\centering\arraybackslash}m{2.4cm} 
>{\centering\arraybackslash}m{2.6cm} 
>{\centering\arraybackslash}m{2.9cm} 
>{\centering\arraybackslash}m{3.2cm}
}
\toprule
\multirow{2}{*}{\large \textbf{Format}} & \multirow{2}{*}{\large \textbf{Paper}} & \multicolumn{3}{c}{\large \textbf{Previous Module}} & \multirow{2}{*}{\large \makecell{\textbf{Action Token}\\\textbf{Restrictiveness}}} & \multicolumn{2}{c}{\large \textbf{Next Module}} & \multirow{2}{*}{\large \textbf{Task}} & \multirow{2}{*}{\large \textbf{Embodiment}} \\
\cmidrule(lr){3-5} \cmidrule(lr){7-8}
& & Model & Training Strategy & Generation Strategy & & Model & Training Strategy & & \\
\midrule
\multirow{14}{*}[-13.5cm]{\rotatebox[origin=c]{90}{\large Language Plans}} 
 & Language Planner~\cite{huang2022language} & Codex-12B, GPT-3-175B, Sentence-RoBERTa-355M & Frozen & LLM generates plans, RoBERTa finds the best match within skill set & Predefined & N/A & N/A & VirtualHome & N/A \\ \cmidrule(lr){2-10}
 & Socratic Models~\cite{zeng2022socratic} & ViLD, LLM & Frozen & VLM detects objects, LLM generates individual steps & Predefined & CLIPort-inspired policy & Trained & Tabletop rearrangement (simulation) & \makecell{UR5 with a gripper \\(simulation)} \\ \cmidrule(lr){2-10}
 & SayCan~\cite{saycan2022arxiv} & PaLM-540B & Frozen & The atomic skill with the highest combined rating from LLM and affordance function is selected & Predefined & BC-Z & Trained on 80K demonstrations &\makecell{Mobile manipulation \\(office kitchen)} & Everyday Robots \\ \cmidrule(lr){2-10}
 & Inner Monologue~\cite{huang2022inner} & PaLM-540B, InstructGPT & Frozen & LLM generates and updates plans with textual feedback, utilizing few-shot prompting & Predefined & CLIPort, BC-Z & CLIPort trained on 20K Pick-Place demonstrations; BC-Z trained on 80K demonstrations & Tabletop rearrangement (simulation, real-world); mobile manipulation (office kitchen) & \makecell{UR5e with a gripper;\\ Everyday Robots} \\ \cmidrule(lr){2-10}
 & PaLM-E~\cite{driess2023palme} & PaLM-E-562B & \makecell{Trained on \\VQA, web text, \\ 
manipulation datasets} & VLM generates plans & Free-form & Interactive Language policy, RT-1 & Trained as in the original papers & TAMP~\cite{PaLM-E}; Language-Table~\cite{lynch2023interactive}; mobile manipulation & xArm6 with a cylindrical gripper; Everyday Robots \\ \cmidrule(lr){2-10}
 & \makecell{EmbodiedGPT \\ ~\cite{mu2023embodiedgpt}} & EmbodiedGPT & Trained on EgoCOT via prefix tuning & VLM generates a sequence of sub-goals & Free-form & MLP policy network & Trained on 10 / 25 / 50 demonstrations per task & Meta-World~\cite{yu2020meta}; Franka Kitchen~\cite{pmlr-v100-gupta20a} & \makecell{Franka Panda\\ (simulation)} \\ \cmidrule(lr){2-10}
 & DoReMi~\cite{guo2024doremi} & Vicuna-13B, BLIP-2 & LLM frozen, BLIP-2 fine-tuned with LoRA & LLM generates plans and constraints through few-shot in-context learning & Predefined & CLIPort, Transporter Nets, DeepMimic for locomotion & Trained on via imitation learning or RL & Tabletop manipulation; humanoid manipulation & \makecell{UR5e with a gripper;\\ Humanoid} \\ \cmidrule(lr){2-10}
 & ViLa~\cite{hu2023look} & GPT-4V & N/A & VLM generates plans via CoT reasoning in a zero-shot mode & Predefined & Scripted, RL, BC policies & Trained & Tabletop manipulation (simulation, real-world) & Franka Panda \\ \cmidrule(lr){2-10}
 & 3D-VLA\cite{zhen20243d} & BLIP-2 FlanT5XL & Fine-tuned on a curated 3D embodied instruction tuning dataset containing 2M scene-language-action pairs & VLM generates plans with interactive tokens & Free-form & Stable Diffusion v1.4, Point-E & Fine-tuned & Long-horizon tasks in RLBench~\cite{RLBench} and CALVIN~\cite{mees2022calvin} & Franka Panda \\ \cmidrule(lr){2-10}
 & RoboMamba~\cite{RoboMamba} & CLIP, Mamba & Alignment pretraining, instruction co-training & VLM generates plans & Free-form & Simple policy head & Trained on 10K end-effector Pose Predictions & Vision-language tasks, pose prediction & Franka Panda\\ \cmidrule(lr){2-10}
 & ReplanVLM~\cite{mei2024replanvlm} & GPT-4V & N/A & VLM generates task plans, another two VLMs detect internal and external errors & Free-form & NR & NR & Tabletop manipulation & JAKA Zu 7 arm with a gripper \\ \cmidrule(lr){2-10}
 & BUMBLE~\cite{shah2024bumble} & GPT-4o & N/A & VLM predicts subtasks and selects parameterized skills & Predefined & NR & NR & Long-horizon building-wide mobile manipulation & NR \\ \cmidrule(lr){2-10}
 & ReflectVLM~\cite{feng2025reflective} & LLaVA-1.5-13B & Trained on demonstrations & VLM proposes plans, diffusion model imagines future images, and VLM reflects on the plans & Predefined & Rule-based script controller & N/A & Manipulation tasks (1K interlocking object) & Franka Panda\\ \cmidrule(lr){2-10}
 & Hi Robot~\cite{shi2025hi} & PaliGemma-3B & Trained on teleoperated demonstrations segmented into short skills and synthetic prompts & VLM generates plans & Free-form & $\pi_0$ & Trained on teleoperated demonstrations & Table bussing, make a sandwich, grocery shopping & UR5e with a gripper; ARX with a gripper; ARX with a gripper and mobile base \\ \cmidrule(lr){2-10}
 & $\pi_{0.5}$~\cite{intelligence2025pi05visionlanguageactionmodelopenworld} & PaliGemma-3B & Trained on robot data, high-level subtask prediction data, and multi-modal web data & VLM generates plans & Free-form &  $\pi_{0.5}$ & Trained on robot data, high-level subtask prediction data, and multi-modal web data & Household (real-world) & Two mobile manipulator platforms \\ 
 \midrule
\multirow{2}{*}[0.2cm]{\rotatebox[origin=c]{90}{\large Language Motion}} 
 & RT-H~\cite{rth2024arxiv} & PaLI-X 55B & Trained on \textit{Kitchen} and \textit{Diverse} datasets labeled with language motions & VLM predicts fine-grained language motion phrases & Free-form & PaLI-X 55B & Trained on \textit{Kitchen} and \textit{Diverse} datasets labeled with language motions & Tabletop manipulation & NR \\ \cmidrule(lr){2-10}
 & NaVILA~\cite{cheng2024navila} & ViLa & Trained on 2K YouTube egocentric touring videos & VLM generates mid-level actions with spatial information & Free-form & Visual locomotion policy & Trained via PPO & \makecell{VLN-CE-Isaac~\cite{cheng2024navila};\\navigation \\(25 tasks, real-world)} & \makecell{Unitree Go2;\\ Unitree H1;\\ Booster T1} \\
\bottomrule
\end{tabular}%
}
\vspace{5pt}
\begin{minipage}{\textwidth}
\footnotesize
``N/A'' indicates not applicable; ``NR'' indicates not reported.
\end{minipage}
\end{table}

%% file: table/code.tex
{
\begin{table}[!t]
\centering
\caption{Overview of VLA research using code as action tokens.}
\label{tab:code_action}
\renewcommand{\arraystretch}{1.3}
\resizebox{\textwidth}{!}{%
\begin{tabular}{
>{\centering\arraybackslash}m{3.8cm} 
>{\centering\arraybackslash}m{3.1cm} 
>{\centering\arraybackslash}m{3.1cm} 
>{\centering\arraybackslash}m{3.8cm} 
>{\centering\arraybackslash}m{3.1cm} 
>{\centering\arraybackslash}m{3.8cm} 
>{\centering\arraybackslash}m{3.8cm}}
\toprule
\multirow{2}{*}{\large \textbf{Paper}} &
\multicolumn{2}{c}{\large \textbf{Previous Module}}  &
\multicolumn{2}{c}{\large \textbf{Next Module}} & 
\multirow{2}{*}{\large \textbf{Task}} & 
\multirow{2}{*}{\large \textbf{Embodiment}} \\
\cmidrule(lr){2-3} \cmidrule(lr){4-5}
& Model & Training Details & Model & Training Details & & \\ \midrule
Code as Policies~\cite{liang2023code} & Codex (\texttt{code-davinci-002}) & API Calling & \makecell{ViLD, MDETR, \\impedance controller,\\ trajectory-based controller} & ViLD, MDETR frozen & \makecell{Draw shape,\\ pick-place;\\ mobile manipulation} & \makecell{UR5e with a \\Robotiq 2F85 gripper\\ (RealSense D435);\\ Everyday Robots}\\ \midrule
ProgPrompt~\cite{Singh2022ProgPromptGS} & GPT-3 & API Calling & \makecell{ViLD, Contact-GraspNet, \\ SceneCollisionNet, \\ motion planning (MPPI)} & Contact-GraspNet, SceneCollisionNet, ViLD frozen & \makecell{VirtualHome \\(simulation);\\ pick-place, \\sort objects \\(real-world)} & Franka Panda\\ \midrule
ChatGPT for Robotics~\cite{Vemprala2023ChatGPTFR} & ChatGPT & API Calling &  Robot function library, YOLOv8 & YOLOv8 frozen & \makecell{Navigation, \\object manipulation, \\ AirSim~\cite{Shah2017AirSimHV} industrial inspection, \\ AirSim obstacle avoidance\\ (simulation);\\ drone flight (real-world)} & NR \\ \midrule
Text2Motion~\cite{Lin_2023} &  GPT-3 (\texttt{text-davinci-003}) & API Calling & \makecell{Skills library, \\geometric feasibility \\planner} & N/A & \makecell{Pick-place \\(simulation)} & \makecell{Franka Panda \\(Kinect V2)} \\ \midrule
Instruct2Act~\cite{Huang2023Instruct2ActMM} & GPT-3 (\texttt{text-davinci-003}) & API Calling & \makecell{SAM, CLIP} & Frozen & Visual manipulation; scene understanding; rotate, rearrange; rearrange restore, pick-restore & NR \\ \midrule
RoboScript~\cite{Chen2024RoboScriptCG} & \makecell{GPT-3.5-turbo / \\GPT-4 / Gemini Pro} & API Calling &  \makecell{GLIP, AnyGrasp, \\GAMMA, GIGA, \\motion planning (RRT)}& GLIP, AnyGrasp, GAMMA, GIGA frozen &  \makecell{Pick-place,\\insert into drawer} & \makecell{Franka Panda, UR5 \\with a Robotiq 2F-85 \\gripper (RGB-D camera)} \\ \midrule
RoboCodeX~\cite{10.5555/3692070.3693552} & RoboCodeX & Pretrained and fine-tuned on self-collected dataset & \makecell{AnyGrasp, GAMMA,\\ motion planning, ROS} & AnyGrasp, GAMMA frozen & \makecell{Pick-place,\\insert into drawer} & \makecell{UR5 with a gripper,\\ Franka Panda \\(3 RGB-D camera)} \\ 
\bottomrule
\end{tabular}%
}
\begin{minipage}{\textwidth}
\footnotesize
``N/A'' indicates not applicable; ``NR'' indicates not reported.
\end{minipage}
\vspace{-15pt}
\end{table}

%% file: table/affordance.tex
{
\begin{table}[!t]
\centering
\caption{Overview of VLA research using affordance as action tokens.}
\label{tab:affordance_action}
\renewcommand{\arraystretch}{1.3}
\resizebox{\textwidth}{!}{%
\begin{tabular}{
>{\centering\arraybackslash}m{1.0cm} 
>{\centering\arraybackslash}m{2.5cm} 
>{\centering\arraybackslash}m{3.0cm} 
>{\centering\arraybackslash}m{3.5cm}
>{\centering\arraybackslash}m{3.0cm} 
>{\centering\arraybackslash}m{3.5cm} 
>{\centering\arraybackslash}m{3.5cm} 
>{\centering\arraybackslash}m{4.0cm}
}
\toprule
\multirow{2}{*}{\large \textbf{Format}} & 
\multirow{2}{*}{\large  \textbf{Paper}} & 
\multicolumn{2}{c}{\large \textbf{Previous Module}} & 
\multicolumn{2}{c}{\large \textbf{Next Module}} 
& \multirow{2}{*}{\large \textbf{Task}} & 
\multirow{2}{*}{\large \textbf{Embodiment}} \\
\cmidrule(lr){3-4} \cmidrule(lr){5-6}
& & Model & Training Details & Model & Training Details & & \\
\midrule
\multirow{7}{*}[-2.5cm]{\centering\rotatebox[origin=c]{90}{\large Keypoint}} 
 & KITE~\cite{sundaresankite}        & Two-stream architecture & \makecell{Skill dataset \\(GTX 1070, 3h)} & \makecell{PointNet++~\cite{qi2017pointnet} \\(20K points)} & \makecell{50 episodes/skill \\(GTX 1070, 1h)} & Tabletop instruction-following; semantic grasping; coffee making & \makecell{Franka Panda \\(3 RealSense D435)} \\ \cmidrule(lr){2-8}
 & \makecell{RoboPoint\\\cite{yuanrobopoint}}    & Vicuna-v1.5-13B with a ViT-L/14 336px image encoder & 660K (image, relation) pairs from 10K scenes (16$\times$A100-80G, 40h) & Motion planning & N/A  & Pick-place & Franka Panda \\ \cmidrule(lr){2-8}
 & CoPa~\cite{huang2024copa}         & GPT-4V, OWL-ViT, SAM, GraspNet & Frozen  & GPT-4V & N/A  & Hammer nail, insert flower, pour water, etc. & \makecell{Franka Panda \\ (2 RealSense D435)} \\ \cmidrule(lr){2-8}
 & RAM~\cite{kuangram}               & \makecell{CLIP, \\Stable Diffusion} & Frozen  & Motion planning (cuRobo) & N/A  & Articulated manipulation & \makecell{Franka Panda \\(RealSense D415); \\Unitree B1 with a Z1 \\(RealSense D415)} \\ \cmidrule(lr){2-8} 
 & ReKep~\cite{huangrekep}           & \makecell{DINOv2,\\ SAM} & Frozen  & Motion planning (constraint optimization) & N/A  & \makecell{Tape box;\\ dual-arm fold} & \makecell{Franka Panda \\(single \& dual arm)} \\ \cmidrule(lr){2-8}
 & OmniManip~\cite{pan2025omnimanip} & GPT-4o & N/A  & Motion planning (constraint optimization) & N/A  & \makecell{Pour tea,\\ open jar,\\ operate drawer} & \makecell{Franka Panda \\with a UMI gripper \\(2 RealSense D415)} \\ \cmidrule(lr){2-8}
 & KUDA~\cite{liu2025kuda}           & SAM & Frozen  & Motion planning (MPPI) & N/A   & Move rope/cube/granular & NR \\ \midrule
\multirow{3}{*}[-0.6cm]{\centering\rotatebox[origin=c]{90}{\large Bounding Box}}
 & GPT-4V for Robotics~\cite{wake2024gpt}         & Detic & Frozen  & Task planner & Hardware-independent executable file & \makecell{Relocate juice,\\ open drawer} & \makecell{Nextage, Fetch with \\a Shadow Dexterous \\Hand Lite} \\ \cmidrule(lr){2-8}
 & A3VLM~\cite{huanga3vlm}           & SPHINX-1K, Llama-2 & 40 images per object in PartNet-Mobility (8$\times$A100-80G, 24h)  & Action primitives & N/A  & Articulated manipulation & Kuka with a Robotiq gripper (RealSense D415)  \\ \cmidrule(lr){2-8}
 & \makecell{DexGraspVLA\\\cite{zhong2025dexgraspvla}} & \makecell{Qwen2.5-VL-\\72B-Instruct} & Frozen  &\makecell{SAM, Cutie, DINOv2, \\Diffusion Policy} & 2K+ human demos (8$\times$A800-80G, 24h) & \makecell{Dexterous grasping \\in clutter} & \makecell{RealMan RM75-6F \\with a PsiBot G0-R \\(RealSense D405C, \\RealSense D435)} \\\midrule
\multirow{3}{*}[-0.5cm]{\centering\rotatebox[origin=c]{90}{\large Segmentation Mask}}
 & MOO~\cite{stone2023open}          & OWL-ViT & Frozen  & RT-1 & Trained on RT-1 data and diverse “pick” data across a set of 90 diverse objects & \makecell{Pick, move near, \\knock, place into, \\place upright} & \makecell{Mobile manipulator \\with a gripper} \\ \cmidrule(lr){2-8}
 & SoFar~\cite{qi2025sofar}          & SAM, Florence-2 & Frozen  & PointSO & OrienText300K (8$\times$H800) & Rearrangement; navigation & \makecell{Franka, UR5e, Flexiv \\(RealSense D415);\\ Unitree GO2 \\(RealSense D455, LiDAR)} \\ \cmidrule(lr){2-8}
 & RoboDexVLM~\cite{liu2025robodexvlm} & SAM & Frozen  & AnyGrasp & N/A  & Open-vocabulary pick-place & UR5 with a Inspire hand (RealSense D435i) \\ \cmidrule(lr){2-8}
 & \makecell{ROCKET-1\\~\cite{cai2025rocket}}     & SAM 2 & Frozen  & TransformerXL & 160M video frames (8$\times$A800-80G, 72h) & Minecraft & Virtual agent \\ \midrule
\multirow{3}{*}[-0.9cm]{\centering\rotatebox[origin=c]{90}{\large Affordance Map}}
 & CLIPort~\cite{shridhar2021cliportpathwaysroboticmanipulation} & \makecell{Two-stream \\architecture\\(CLIP ResNet50, \\Transporter ResNet)} & \makecell{Trained on \\self-collected \\data} & Motion primitive & N/A  & \makecell{Ravens~\cite{zeng2021transporter} (simulation);\\Pick, place, pack, \\move, fold, sweep\\(real-world)} & \makecell{UR5e with a \\suction gripper \\(simulation); \\Franka Panda \\(real-world)} \\ \cmidrule(lr){2-8}
 & VoxPoser~\cite{huang2023voxposer} & \makecell{GPT-4, \\OWL-ViT, \\XMEM} & Frozen  & Motion planning & N/A  & \makecell{Move-avoid,\\ set table,\\ sweep trash} & \makecell{Franka Panda \\(2 Azure Kinect)} \\ \cmidrule(lr){2-8}
 & ManipLLM~\cite{10656630}          & N/A & N/A  & LLaMA with a L-Adapter &  \makecell{10K successful samples\\ (A100-40G, 10h)} & Articulated manipulation & \makecell{Franka Panda \\(RealSense 415)} \\ \cmidrule(lr){2-8}
 & ManiFoundation~\cite{xu2024manifoundation} & CVAE & 3000 objects  & Motion planning (RRT-Connect) & N/A  & \makecell{Fold cloth, \\ rearrange rope, \\ breakfast preparation} & \makecell{Kinova MOVO, Flexiv \\with a Leap Hand} \\
\bottomrule
\end{tabular}%
}
\vspace{5pt}
\begin{minipage}{\textwidth}
\footnotesize
``N/A'' indicates not applicable; ``NR'' indicates not reported.
\end{minipage}
\end{table}

%% file: table/trajectory.tex
{
\begin{table}[!t]
\centering
\caption{Overview of VLA research using trajectory as action tokens. The ``Format'' column categorizes action tokens into three types: \textit{Point Trajectory}, representing the paths of a few keypoints; \textit{Visual Trajectory}, representing a path drawn directly onto the image; and \textit{Optical Flow}, representing the motion of all pixels. $T$ is the temporal span, $K$ the number of points, and $(H, W)$ the image resolution.}
\label{tab:trajectory_action}
\renewcommand{\arraystretch}{1.3}
\resizebox{\textwidth}{!}{%
\begin{tabular}{
>{\centering\arraybackslash}m{2.3cm} 
>{\centering\arraybackslash}m{3.2cm}
>{\centering\arraybackslash}m{3.5cm} 
>{\centering\arraybackslash}m{4.0cm}
>{\centering\arraybackslash}m{2.7cm} 
>{\centering\arraybackslash}m{3.5cm} 
>{\centering\arraybackslash}m{3.5cm} 
>{\centering\arraybackslash}m{2.7cm}}
\toprule
\multirow{2}{*}{\large \textbf{Paper}} & 
\multicolumn{2}{c}{\large \textbf{Previous Module}} & 
\multirow{2}{*}{\large \textbf{Format}} & 
\multicolumn{2}{c}{\large \textbf{Next Module}} & 
\multirow{2}{*}{\large \textbf{Task}} & 
\multirow{2}{*}{\large \textbf{\makecell{Embodiment}}} \\
\cmidrule(lr){2-3} \cmidrule(lr){5-6}
& Model & Training Details & & Model & Training Details & & \\ \midrule
AVDC~\cite{ko2023avdc}             & Video diffusion model, GMFlow  & \makecell{Diffusion model trained\\ on Bridge~\cite{ebert2021bridgedataboostinggeneralization} and 20 \\human demonstrations;\\ GMFlow frozen} & \makecell{Optical Flow\vspace{0.5em}\\ $\mathbf{V}\in\mathbb{R}^{H\times W\times 2}$} & Rigid body transformation regression & N/A & \makecell{Meta-World,\\ iTHOR~\cite{iTHOR};  tabletop\\ manipulation} & Franka Panda (RealSense D435) \\ \midrule
\makecell{RT-Trajectory\\\cite{rt-trajectory}} & \makecell{Code as Policies /\\ PALM-E} & Frozen & \makecell{Visual Trajectory\vspace{0.5em} \\ $\mathbf{I} \in \mathbb{R}^{H\times W\times 3}$} & RT-1 & Trained on RT-1 dataset & \makecell{Pick-place, \\open/close drawer,\\fold towel,\\swivel chair}& Everyday Robots\\ \midrule
ATM~\cite{wen2023any}              & Track Transformer & Trained on 50 action-free video demonstrations & \makecell{Point Trajectory\vspace{0.5em} \\$\mathbf{P} \in \mathbb{R}^{T \times K \times 2}$} & \makecell{Transformer,\\ MLP} & Trained on 10 action-labeled demonstrations & \makecell{LIBERO~\cite{LIBERO} \\(simulation); \\pick-place, \\squeeze objects\\(real-world)} & UR5 with a gripper \\ \midrule
LLARVA~\cite{niu2024llarva}        & LLaVA 1.5 & Trained on 8.5M image-visual trace pairs & \makecell{Point Trajectory\vspace{0.5em} \\ $\mathbf{P}\in\mathbb{R}^{T\times K\times2}$\\($K=1$)} & N/A & N/A & \makecell{RLBench \\(simulation);\\pick cubes, \\stack/destack cubes, \\(real-world)} & Franka Panda\\ \midrule
\makecell{Im2Flow2Act\\ \cite{xu2024flow}} & \makecell{Grounding DINO,\\ TAPIR, CLIP,\\AnimateDiff} & Grounding DINO frozen; The decoder from Stable Diffusion fine-tuned; AnimateDiff fine-tuned via LoRA on human demonstration videos & \makecell{Visual Trajectory\vspace{0.5em} \\ $\mathbf{V} \in \mathbb{R}^{T \times H \times W \times 3}$} & State encoder, temporal alignment module, diffusion action head & Trained on 4800 simulated robot exploration data for 500 epochs & \makecell{Pick-place,\\ pouring,\\ open drawer,\\ fold cloth} & UR5e with a WSG-50 gripper (RealSense D415) \\ \midrule
FLIP~\cite{flip}                   & CVAE with transformer & Trained on 40 videos & \makecell{Point Trajectory\vspace{0.5em} \\ $\mathbf{P} \in \mathbb{R}^{T \times K \times 2}$} & Diffusion Policy & Trained on 10 demonstrations with action labels and 50 demonstrations without action labels & \makecell{LIBERO-LONG,\\ FMB~\cite{FMB} (simulation); \\fold cloth, \\unfold cloth \\(real-world)} & xArm6 (2 RealSense D435i) \\ \midrule
\makecell{HAMSTER \\\cite{li2025hamster}}       & VILA-1.5-13B & Fine-tuned on 770K object location tasks, 320K simulated 2D end-effector paths, 110K real robot 2D end-effector paths, 660K VQA & \makecell{Visual Trajectory\vspace{0.5em} \\ $\mathbf{I} \in \mathbb{R}^{H\times W\times 3}$} & RVT-2 / 3D-DA & Trained on 320 teleoperation episodes & \makecell{Pick-place,\\knock down objects, \\press button} & Franka Panda \\ 
\bottomrule
\end{tabular}%
}
\vspace{5pt}
\begin{minipage}{\textwidth}
\footnotesize
``N/A'' indicates not applicable.
\end{minipage}
\end{table}

%% file: table/goal_state.tex
\begin{table}[!t]
\centering
\caption{Overview of VLA research using goal state as action tokens.}
\label{tab:goal_state}
\renewcommand{\arraystretch}{1.3}
\resizebox{\textwidth}{!}{%
\begin{tabular}{
>{\centering\arraybackslash}m{0.7cm} 
>{\centering\arraybackslash}m{2.4cm} 
>{\centering\arraybackslash}m{2.5cm} 
>{\centering\arraybackslash}m{4.5cm} 
>{\centering\arraybackslash}m{2.2cm} 
>{\centering\arraybackslash}m{2.5cm} 
>{\centering\arraybackslash}m{4.5cm} 
>{\centering\arraybackslash}m{3.1cm} 
>{\centering\arraybackslash}m{2.5cm} 
>{\centering\arraybackslash}m{2.5cm}
}
\toprule
\multirow{2}{*}{\large \textbf{\makecell{Type}}} & 
\multirow{2}{*}{\large \textbf{Paper}} & 
\multicolumn{2}{c}{\large \textbf{Previous Module}} & 
\multirow{2}{*}{\large \textbf{\makecell{Format}}} & 
\multicolumn{2}{c}{\large \textbf{Next Module}} & 
\multirow{2}{*}{\large \textbf{Task}} & 
\multirow{2}{*}{\large \textbf{\makecell{Generali-\\zation}}} & 
\multirow{2}{*}{\large \textbf{\makecell{Embodi-\\ment}}} \\
\cmidrule(lr){3-4} \cmidrule(lr){6-7}
& & Model & Training Details & & Model & Training Details & & & \\  
\midrule
\multirow{4}{*}[-1.8cm]{\centering\rotatebox[origin=c]{90}{\large Single-frame}} 
 & SuSIE~\cite{black2023zeroshotroboticmanipulationpretrained} & Instruct Pix2Pix & Fine-tuned on BridgeData V2, Something-Something~\cite{Something} & Image &  Diffusion Policy  & \makecell{Trained on BridgeData V2\\ (Goal-Conditioned \\Behaviour Cloning) }& CALVIN; tabletop manipulation (real-world) & Environment, Object (zero-shot) & WidowX250 \\ \cmidrule(lr){2-10}
 & 3D-VLA~\cite{zhen20243d} & Conditional Diffusion Model & Fine-tuned via LoRA on OXE, HoNY, RH20T, EPIC-KITCHENS-100, HOI4D & RGB-D image, Point cloud & BLIP2-FlanT5XL & \makecell{Fine-tuned on 2M\\ 3D-language-action\\ data pairs} & RLBench, CALVIN; tabletop manipulation (real-world) & Environment (zero-shot) & NR \\ \cmidrule(lr){2-10}
 & \makecell{CoTDiffusion\\~\cite{Ni_2024_CVPR}} & Semantic Alignment Module, Diffusion Model & Trained on 10K trajectories with annotated ground truth keyframe for each task (coarse-grained pretraining, fine-grained train) & Image & ViT Encoder, MLP & Trained on 10K trajectories & VIMA-Bench~\cite{jiang2022vima} & \makecell{Placement, \\Object, \\Task} (zero-shot) & N/A \\ \cmidrule(lr){2-10}
 & CoT-VLA~\cite{zhao2025cotvlavisualchainofthoughtreasoning} & VILA-U & Fine-tuned on robot action data, videos without action labels (VILA-U's vision tower frozen) & Image & Full-Attention Module (based on VILA-U) & Fine-tuned on task demonstrations collected on downstream robot tasks & \makecell{LIBERO; \\Bridge-V2~\cite{walke2024bridgedatav2datasetrobot}, \\tabletop manipulation \\(real-world)} & \makecell{Environment, \\Task, \\Instruction \\(zero-shot)} & \makecell{WidowX, \\Franka Panda} \\ 
\midrule
\multirow{7}{*}[-6.0cm]{\centering\rotatebox[origin=c]{90}{\large Multi-frame}} 
 & UniPi~\cite{du2023learninguniversalpoliciestextguided} & Conditional Diffusion Model, Temporal Super Resolution & \makecell{CDM pretrained and \\fine-tuned, TSR fine-tuned \\on 14M video-text pairs,\\ 60M image-text pairs,\\ LAION-400M, and Bridge} & Video & CNN inverse dynamic model & Trained from scratch & Bridge; tabletop manipulation (real-world, simulation) & Instruction, Object, Environment (zero-shot) & NR \\ \cmidrule(lr){2-10}
 & AVDC~\cite{ko2023avdc} & Conditional Diffusion Model & Trained on Bridge & Video & \makecell{GMFlow~\cite{xu2022gmflow},\\rigid body\\transformation\\regression} & Frozen & \makecell{Meta-World,\\ iTHOR; tabletop\\ manipulation\\(real-world)} & Environment (zero-shot) & Franka Panda (Realsense D435) \\ \cmidrule(lr){2-10}
 & VLP~\cite{du2023video} & PaLM-E, UniPi & \makecell{PaLM-E fine-tuned and\\ UniPi trained from scratch \\on 10000 long-horizon\\ trajectories}& Video & LAVA with ResNet encoder & Trained on RT-1, Bridge, RT-2, Ego4D, EPIC-KITCHENS-100, LAION-400M & \makecell{Tabletop \\manipulation,\\group color,\\make a line\\(real-world,\\simulation)} & \makecell{Lighting, \\Object, \\Task \\(zero-shot)} & Mobile manipulator; Bi-manual ALOHA \\ \cmidrule(lr){2-10}
 & Gen2Act~\cite{bharadhwaj2024gen2acthumanvideogeneration} & Gemini, VideoPoet & Frozen & Human video & ViT, Bootstap, Tap-vid & ViT trained on RT-1 and 400 diverse robot trajectories; Bootstap, Tap-vid frozen & Tabletop manipulation (real-world, simulation) & \makecell{Environment,\\ Object, \\Task \\(zero-shot)} & Mobile manipulator with a gripper \\ \cmidrule(lr){2-10}
 & VPP~\cite{hu2025videopredictionpolicygeneralist} & Stable Video Diffusion & Fine-tuned on Something-Something V2, self-collected datasets, internet robot datasets & Video & Diffusion Policy & Trained on self-collected datasets & \makecell{CALVIN, \\Meta-World;\\ tabletop\\ manipulation, \\tool using \\(real-world)} & Environment, Object (zero-shot) & Franka Panda, Xarm with an Xhand \\ \cmidrule(lr){2-10}
 & FLIP~\cite{flip} & CVAE with transformer, DiT, LIV & \makecell{CVAE and DiT trained on\\ long-horizon videos \\with flow annotations; \\LIV fine-tuned on\\ long-horizon imperfect videos} & Video & Diffusion Policy & \makecell{Trained on a few demonstrations \\with action labels} & \makecell{LIBERO, FMB\\(simulation); \\fold cloth, \\unfold cloth\\(real-world)} & Environment, Instruction (zero-shot) & xArm6 (2 RealSense D435i) \\ \cmidrule(lr){2-10}
 & GEVRM~\cite{zhang2025gevrmgoalexpressivevideogeneration} & DiT & Fine-tuned on Bridge, CALVIN & Video & Diffusion Policy & \makecell{Trained on Image-action\\ label pairs} & CALVIN; Bridge & Environment (zero-shot) & UR5 with a gripper \\ 
\bottomrule
\end{tabular}%
}
\vspace{5pt}
\begin{minipage}{\textwidth}
\footnotesize
``N/A'' indicates not applicable; ``NR'' indicates not reported.
\end{minipage}
\end{table}

%% file: table/latent.tex
\begin{table}[!t]
\centering
\caption{Overview of VLA research using latent representation as action tokens.}
\label{tab:latent}
\renewcommand{\arraystretch}{1.3}
\resizebox{\textwidth}{!}{%
\begin{tabular}{
>{\centering\arraybackslash}m{1.6cm} 
>{\centering\arraybackslash}m{1.0cm}
>{\centering\arraybackslash}m{3.2cm} 
>{\centering\arraybackslash}m{2.8cm} 
>{\centering\arraybackslash}m{0.7cm} 
>{\centering\arraybackslash}m{2.4cm} 
>{\centering\arraybackslash}m{2.6cm}  
>{\centering\arraybackslash}m{2.4cm} 
>{\centering\arraybackslash}m{2.6cm} 
>{\centering\arraybackslash}m{2.8cm}
>{\centering\arraybackslash}m{2.4cm}}
\toprule
\multirow{2}{*}{\large \textbf{Paper}} & 
\multicolumn{4}{c}{\large \textbf{Latent Construction}} & 
\multicolumn{2}{c}{\large \textbf{Previous Module}} & 
\multicolumn{2}{c}{\large \textbf{Next Module}} & 
\multirow{2}{*}{\large \textbf{Task}} & 
\multirow{2}{*}{\large \textbf{\makecell{Embodiment}}} \\
\cmidrule(lr){2-5} \cmidrule(lr){6-7} \cmidrule(lr){8-9}
& Method & Encoded Information & Dataset & Aligned & Model & Training Strategy & Model & Training Strategy & & \\ \midrule
\makecell{OmniJARVIS\\~\cite{wang2024omnijarvis}} & \makecell{FSQ\\~\cite{mentzer2023finite}} &Goal behavior from interaction trajectory & Synthetic Data by IDM and LLM & \textcolor{red}{\xmark} & LLaVA-7B & Trained on data synthesized with an IDM and LLM & Transformer-based latent action decoder & Trained on data synthesized with an IDM and LLM & Minecraft; open-ended question answering, instruction following &\makecell{Mouse, \\keyboard \\(virtual agent)} \\ \midrule
\makecell{QueST\\~\cite{mete2024quest}} & FSQ & \makecell{Latent skills \\abstracted from \\raw action} & LIBERO, Meta-World Dataset & \textcolor{red}{\xmark} & \makecell{CLIP, ResNet,\\Transformer \\Decoder} & Trained on benchmark dataset & Transformer-based latent action decoder & Few-shot fine-tuned on task-specific demonstrations & LIBERO, Meta-World & Simulated robot \\ \midrule
\makecell{LAPA\\~\cite{ye2024latent}} & \makecell{NSVQ\\~\cite{vali2022nsvq}} & Visual difference between nearby frames (1 frame apart) & BridgeData V2, Something-Something V2, OXE & \textcolor{red}{\xmark} & LWM-Chat-1M(7B) & Trained on cross-embodiment dataset (8$\times$H100) & MLP head & Few-shot fine-tuned on task-specific demonstrations & Language-Table, SimplerEnv~\cite{li2024simpler}; Tabletop manipulation (real-world) & Franka Panda; 14-DOF bi-manual robot\\ \midrule
\makecell{GROOT-2\\~\cite{cai2024groot}} & \makecell{VAE\\~\cite{kingma2013auto}} & Goal behavior from interaction trajectory & Benchmark dataset & \textcolor{green}{\cmark} & \makecell{Transformer \\Encoder, BERT} & Trained on task-relevant dataset & Transformer-XL model & Trained on task-relevant dataset & ALE Atari~\cite{bellemare2013arcade}, Minecraft SkillForge~\cite{cai2023groot}, Language-Table, SimplerEnv &\makecell{Mouse, \\keyboard \\(virtual agent)} \\ \midrule
\makecell{GO-1\\~\cite{bu2025agibot}} & \makecell{VQ-VAE\\~\cite{van2017neural}} & Visual difference between nearby frames (30 frames apart) & \makecell{Web-scale \\vision-language\\ data, Ego4D,\\ Cross-embodiment \\robot data, \\AgiBot World} & \textcolor{red}{\xmark} & InternVL2.5-2B & Trained on cross-embodiment dataset & Diffusion-based action head & Pretrained on AgiBot World, Few-shot fine-tuned on task-specific data & Household ( restock, fold, wipe, pour) & AgiBot G1 \\ \midrule
\makecell{UniVLA\\~\cite{bu2025univla}} & \makecell{VQ-VAE} & Visual difference between nearby frames (1 second apart, task-centric) & Data mixture from OXE, GNM~\cite{GNM}, and Ego4D & \textcolor{green}{\cmark} & Prismatic-7B & Trained on a mixture dataset (A100 960h) & Simple action head & Few-shot fine-tuned on task-specific demonstrations & LIBERO, CALVIN, SimplerEnv, R2R~\cite{R2R}; manipulation tasks (real-world) & Piper arm (from AgileX Robotics) with a gripper \\ 
\bottomrule
\end{tabular}

}
\end{table}

%% file: table/raw_action.tex
\begin{table}[!htbp]
\centering
\caption{Overview of VLA research using raw action as action tokens.}
\label{tab:raw_action_p1}
\renewcommand{\arraystretch}{1.4}
\resizebox{\textwidth}{!}{%
\begin{tabular}{
>{\centering\arraybackslash}m{2.0cm}
>{\centering\arraybackslash}m{2.0cm}  
>{\centering\arraybackslash}m{3.5cm}  
>{\centering\arraybackslash}m{4.2cm}  
>{\centering\arraybackslash}m{5.2cm}  
>{\centering\arraybackslash}m{3.8cm}  
>{\centering\arraybackslash}m{2.6cm}  
>{\centering\arraybackslash}m{2.0cm}  
}
\toprule
\textbf{Paper} & 
\textbf{Action Head Type} & 
\textbf{Action Token Format} & 
\textbf{Model} & 
\textbf{Training Strategy} & 
\textbf{Task} & 
\textbf{Embodiment} & 
\textbf{Frequency} \\
\midrule
LangLfP~\cite{LangLfP} & CVAE & 6-DoF Cartesian position, euler angle of the end effector, 2-DoF gripper angle & \makecell{Language encoder: \\simple MLP; \\Vision encoder: \\simple CNN; \\Action head: CVAE} & Trained on 10K language-conditioned and 10M goal-image-conditioned robot data (8$\times$V100 72h) & Tabletop manipulation (3D Playroom~\cite{lynch2019learninglatentplansplay}) & 3D Playroom & 30 Hz \\
\midrule
BC-Z~\cite{BC-Z} & Multi-head MLP & 6-DoF Cartesian position, axis-angle of the end effector, delta form, 1-DoF gripper angle (3-DoF mobile base) & Language Encoder: MUSE; Video Encoder: ResNet-18; Backbone: FiLM-conditioned ResNet-18, multi-head MLP action head& Trained on 26K robot data and 19K human videos of 100 tasks & Tabletop manipulation & Everyday Robots & 10 Hz \\
\midrule
Gato~\cite{Gato} & Autoregressive transformer & \makecell{Task-specific text/\\control action} & 1.2B decoder-only transformer & \makecell{Trained jointly on 1.5T tokens\\ from 604 tasks across VQA, game,\\ robot control using 16$\times$16 \\(TPU v3 96h)} & \makecell{DM Lab, ALE Atari;\\ RGB Stacking Benchmark\\ (simulation, real-world)} & \makecell{DM Lab, ALE Atari;\\ Sawyer (real-world)} & 20 Hz\\
\midrule
VIMA~\cite{jiang2022vima} & Autoregressive transformer & Two SE(2) poses (pick/place or push start/end), discretized into bins & \makecell{Multimodal prompt encoder: \\Mask R-CNN, ViT, T5; \\Backbone: transformer} & \makecell{Trained on 650K trajectories \\of 17 tasks in VIMA-Bench\\ (8$\times$V100 24h)} &\makecell{Tabletop manipulation \\(VIMA-Bench)} & UR5 with a suction cup or a spatula & N/A \\
\midrule
RT-1~\cite{Brohan2022RT1RT} & Autoregressive transformer & 6-DoF end effector pose, 1-DoF gripper state, 3-DoF base, 1 mode token, 256 bins per dimension, delta, single-step & \makecell{Language encoder: USE; \\Vision encoder: \\FiLM-conditioned \\EfficientNet-B3; \\Backbone: TokenLearner, \\transformer} & Trained on RT-1 dataset consisting of \textasciitilde130K episodes across 700+ tasks & Mobile manipulation (office kitchen)  & Everyday Robots & 3 Hz \\
\midrule
RT-2~\cite{Brohan2023RT2VM} & Autoregressive transformer & 6-DoF end effector pose, 1-DoF gripper state, 1 termination command, delta, single-step & \makecell{PaLI-X (5B/55B) /\\ PaLM-E (12B)} & Co-trained on RT-1 dataset and web-scale vision-language data & Mobile manipulation (office kitchen) & Everyday Robots & \makecell{1-3 Hz (55B),\\ \textasciitilde5 Hz (5B)} \\
\midrule
RT-X~\cite{RT-X} & Autoregressive transformer & Same as RT-1 or RT-2 & RT-1/RT-2 & RT-1 and RT-2 trained on OXE subset including 9 embodiments to obtain RT-1-X and RT-2-X & Small-data and large-data domains within OXE & Task-specific embodiments & 3-10 Hz \\
\midrule
\makecell{RoboFlamingo\\\cite{RoboFlamingo}} & LSTM, MLP & 6-DoF end effector pose, 1-DoF binary gripper state, delta, single-step & OpenFlamingo, action head & Trained on CALVIN dataset (8$\times$A100 39h for 3B version, 104h for 9B version) & CALVIN & Franka Panda & NR \\
\midrule
LEO\ \cite{huang2023embodied} & Autoregressive transformer & Navigation: 4 discrete commands (forward, left, right, stop); Manipulation: 6-DoF pose& \makecell{Image encoder: \\OpenCLIP ConvNext; \\3D encoder: PointNet++,\\ Spatial Transformer; \\Backbone: Vicuna-7B} & Trained on LEO-align (1.03M 3D-VL pairs from Objaverse~\cite{objaverse}, ScanNet~\cite{scannet}, 3RScan~\cite{Wald2019RIO}) and LEO-instruct dataset (505K multi-task data covering QA, planning, navigation, manipulation, etc.) & 3D VQA, captioning, dialogue and planning; object navigation (MP3D ObjNav~\cite{ramrakhya2022habitat}), robotic manipulation (CLIPort) & Navigation: AI Habitat; Manipulation: CLIPort & NR \\
\midrule
GR-1~\cite{wu2023unleashinglargescalevideogenerative} & MLP & 6-DoF end effector pose, 1-DoF binary gripper state, delta, single-step & \makecell{Language encoder: CLIP;\\ Vision encoder: MAE-\\pretrained ViT, \\perceiver resampler;\\ Backbone: GPT-style \\transformer} & Video generation pretraining on 800K video clips containing 8M frames from Ego4D, robot data fine-tuning on task-specific data &\makecell{CALVIN,\\transport objects;\\articulated manipulation \\(real-world)} & \makecell{Franka Panda \\(simulation) \\Kinova Gen2 \\(real-world)} & NR \\
\midrule
Octo\ \cite{Octo} & Diffusion & 6-DoF end effector pose, 1-DoF gripper state, delta, action chunking & \makecell{Language encoder: \\T5-base (frozen);\\ Vision encoder: CNN;\\ Backbone: transformer, \\lightweight diffusion head} & \makecell{Pretrained on 800K trajectories\\ from 25 datasets in OXE\\ (TPU v4-128 pod 14h); \\Fine-tuned on  \textasciitilde100 trajectories\\ (A5000 5h for each task)} & Zero-shot: in-distribution manipulation tasks; Fine-tune: new manipulation tasks on new embodiments & Zero-shot: WidowX, UR5, Everyday Robots; Fine-tune: Franka Panda, ViperX, ALOHA & 5-15 Hz \\
\midrule
OpenVLA\ \cite{kim2024openvla} & Autoregressive transformer & 6-DoF end effector pose, 1-DoF gripper state, delta, single-step & Prismatic-7B VLM & Pretrained on 970K robot episodes from OXE (64$\times$A100 336h); Fine-tuned on 10–150 demonstrations for each task (support parameter-efficient fine-tuning) & Zero-shot: manipulation tasks in BridgeData V2 and Google Robot Evaluations; Fine-tune: Franka-Tabletop, Franka-DROID, LIBERO & WidowX, Google Robot, Franka Panda & 6 Hz on RTX 4090 \\
\midrule
TinyVLA~\cite{wen2025tinyvlafastdataefficientvisionlanguageaction} & Diffusion & 7-DoF end effector pose, 1-DoF gripper state, absolute positions, single-step & \makecell{Vision encoder: ViT;\\ Backbone: Pythia;\\ Action head: Diffusion} & VLM pretraining using LLaVA pipeline and dataset; Parameter-efficient fine-tuning on 100 robot data for each task& Tabletop manipulation & Franka Panda, Bimanual UR5 & NR \\
\midrule
HiRT\ \cite{zhang2024hirt} & MLP & 6-DoF end effector pose, 1-DoF gripper state, delta, single-step & Understanding module: InstructBLIP-7B (LoRA); Execution module: lightweight policy with a visual encoder (EfficientNet-B3 for simulation, ViT-B/16 for real-world) and a conditioned action head & Trained on Meta-World (20 tasks, 50 demonstrations each), Franka-Kitchen (5 tasks, 100 demonstrations each), and 4 real tasks with 2000 custom trajectories; VLM fine-tuned with LoRA & Meta-World, Franka-Kitchen; pick-place, press button, route cable, open drawer (real-world) & Franka Panda (real-world) & 9.8 Hz \\
\bottomrule
\end{tabular}
}
\vspace{5pt}
\begin{minipage}{\textwidth}
\footnotesize
\textit{This table continues on the next page (Part II).}
\end{minipage}
\end{table}


\begin{table}[!htbp]
\centering
\label{tab:raw_action_p2} 
\renewcommand{\arraystretch}{1.3}
\begin{minipage}{\textwidth}
\footnotesize
\textit{Table~\ref{tab:raw_action_p1} (continued): Part II.}
\end{minipage}
\vspace{10pt}
\resizebox{\textwidth}{!}{%
\begin{tabular}{
>{\centering\arraybackslash}m{2.0cm}
>{\centering\arraybackslash}m{2.0cm}  
>{\centering\arraybackslash}m{3.5cm}  
>{\centering\arraybackslash}m{4.2cm}  
>{\centering\arraybackslash}m{5.2cm}  
>{\centering\arraybackslash}m{3.8cm}  
>{\centering\arraybackslash}m{2.6cm}  
>{\centering\arraybackslash}m{2.0cm}  
}
\toprule
\textbf{Paper} &
\textbf{Action Head Type} & 
\textbf{Action Token Format} & 
\textbf{Model} & 
\textbf{Training Strategy} & 
\textbf{Task} & 
\textbf{Embodiment} & 
\textbf{Frequency} \\
\midrule
GR-2\ \cite{cheang2024gr2generativevideolanguageactionmodel} & CVAE & 7-DoF Cartesian space action trajectory (end effector pose, binary gripper state) & \makecell{Language encoder: CLIP \\text encoder (frozen); \\Video Encoder: \\VQGAN (frozen); \\Backbone: \\GPT-style Transformer; \\Action Head: CVAE} & Pretraining: Video generation on 38M internet and robot videos (Howto100M, Ego4D, RT-1, etc.); Fine-tuning: Joint video and action trajectory prediction on domain-specific robot data (e.g., CALVIN, 40K trajectories for multi-task tabletop manipulation) & CALVIN; tabletop manipulation (multi-task), pick bin (single/cluttered object, real-world) & Kinova Gen3 with a Robotiq 2F-85 gripper & 200 Hz \\
\midrule
RDT\ \cite{liu2024rdt} & Diffusion & 128-dim unified space, 64 steps action chunking & \makecell{Language encoder: \\T5-XXL (frozen);\\ Vision encoder: \\SigLIP (frozen);\\ Backbone: DiT} & Pretrained on 1M+ trajectories from 46 diverse robot datasets (RT-1, DROID, RH20T, etc.) (48$\times$H100 720h), then on 6K+ trajectories from a self-collected bimanual dataset on (48$\times$H100 72h); Fine-tuned on at least 1-5 demonstrations to learn a new skill & Zero-shot: wash cup, pour water, pour with a specific amount or with a specific hand; Few-shot: fold cloth, handover, fine-grained joystick control for a robot dog & ALOHA dual-arm robot & 381 Hz \\
\midrule
$\pi_0$\ \cite{black2410pi0} & Flow Matching & 18-DoF unified action space, 50 steps action chunking & \makecell{PaliGemma-3B,\\ action expert (300M)} & Pretrained on over 10K hours of mixed data (a subset of OXE and the $\pi$ dataset); Fine-tuned to follow language commands, learn new dexterous tasks or multi-stage tasks & Zero-shot: laundry fold, table bussing, etc.; Fine-tuned: laundry fold, box assembly, etc.& UR5e, bimanual UR5e, Franka, bimanual Trossen, bimanual ARX, AgileX, Mobile Trossen, ARX, Mobile Fibocom & 50 Hz \\
\midrule
CogACT\ \cite{li2024cogact} & Diffusion & 6-DoF end effector pose, 1-DoF binary gripper state, delta, 15 steps action chunking &\makecell{Prismatic-7B VLM,\\ DiT action module} & Pretrained on 22.5M frames from OXE (16$\times$A100 120h); Fine-tuned on 391 demonstrations for RealMan, 400 demonstrations for Franka & \makecell{SimplerEnv;\\ pick, stack, place \\(real-world)} & \makecell{Google Robot,\\ WidowX \\(SimplerEnv);\\RealMan, Franka\\ Panda} & 5-30 Hz \\
\midrule
$\pi_0$-FAST\ \cite{Pi0-FAST} & Autoregressive transformer & FAST tokens, 1 second action chunking & $\pi_0$/OpenVLA backbone & Train universal tokenizer on a cross-embodied dataset of \textasciitilde1M real robot action trajectories; Pretrained on the dataset used by $\pi_0$ requiring 80\% less computation than $\pi_0$ & Zero-shot: same tasks with $\pi_0$, DROID~\cite{khazatsky2025droidlargescaleinthewildrobot} tasks; zero-shot generalization to unseen environments (evaluate a separate policy trained on only the DROID dataset)& Embodiments in $\pi_0$; Franka Panda (DROID) & 5-50 Hz\\
\midrule
UniAct\ \cite{UniAct} & MLP & Embodiment-specific action space & \makecell{Backbone: \\LLaVA-One-Vision-0.5B; \\Universal Action Space: \\Vector-Quantized Codebook \\(256$\times$128); \\Action Decoder: \\lightweight MLPs for \\each embodiment}& Pretrained on 1M trajectories from 28 embodiments (OXE, DROID, LIBERO, etc.) (64$\times$A100 240h); Fine-tune only the MLP head or adapt to ACT decoder with minimal data & Fine-tuned: LIBERO; 19 real-world WidowX tasks; fast adaptation to unseen robots on complex tasks (AIRBOT stacking and bimanual manipulation) & Franka Panda (LIBERO); WidowX, AIRBOT (single \& bi-manual) & NR \\
\midrule
OpenVLA-OFT\ \cite{openvla-oft} & Parallel decoding transformer & 6-DoF end effector pose, 1-DoF gripper state (per arm), continuous, delta, action chunking (8 for LIBERO, 25 for ALOHA) & OpenVLA backbone, FiLM conditioning, MLP action head & Start with pretrained OpenVLA; Fine-tuning: LoRA on target tasks; LIBERO: \textasciitilde500 demonstrations/suite; ALOHA: 20-300 demonstrations/task & LIBERO; fold cloth, tool using (real-world) & Franka Panda (LIBERO); ALOHA bimanual robot (real-world) & up to 109.7 Hz \\
\midrule
JARVIS-VLA\ \cite{JARVIS-VLA} & Autoregressive transformer & 51 token bins: 22 mouse control, 29 keyboard & Qwen2-VL-7B/Llava-Next-8B & \makecell{Three-stage training on: \\277K Minecraft world knowledge \\QA pairs; 35K VQA pairs and 404K \\grounding data; 7.4M frames of \\human/agent play data, 6.4M \\synthesized GUI data; \\Cost: stage 1\&2: 128 GPU hours, \\stage 3: 512 GPU hours \\(32$\times$A800 16h)} & Minecraft MCU Benchmark~\cite{lin2023mcu} (across over 1K atomic tasks) & \makecell{Mouse, \\keyboard \\(virtual agent)} & 55 Hz \\
\midrule
HybridVLA\ \cite{HybridVLA} & Diffusion and autoregressive transformer & 6-DoF end effector pose, 1-DoF binary gripper state (per arm), delta, single state & \makecell{7B: Prismatic-7B VLM;\\ 2.7B: CLIP, Phi-2}  & Pretrained on 760K trajectories of 35 datasets; Fine-tuned on RLBench (simulation) or self-collected real-world data (both 100 trajectories per task)& 10 tabletop tasks from RLBench; 5 single-arm and 5 dual-arm real-world manipulation tasks& Franka Panda (RLBench); Franka Research 3 (single-arm); AgileX dual-arm robot & 6.1 Hz \\
\midrule
GR001 N1\ \cite{nvidia2025gr00tn1openfoundation} & Flow Matching & Embodiment-specific action space, 16 steps action chunking & \makecell{System 2: Eagle-2 VLM; \\System 1: DiT with \\cross-attention; \\Interface: embodiment-\\specific MLPs} & \makecell{Pretrained on a data pyramid of: \\3.3K hours of real robot data from \\OXE, AgiBot-Alpha, etc.; 1.7K \\hours of synthetic data; 2.5K hours \\human video data from Ego4D, \\EPIC-KITCHEN, etc. \\(\textasciitilde50K H100 hours)\\Fine-tuned on 30-300 \\demonstrations per target task} & Zero-shot: bimanual handover, novel object placement; Fine-tuned: RoboCasa Kitchen~\cite{nasiriany2024robocasa}, DexMimicGen Cross-Embodiment Suite~\cite{jiang2024dexmimicgen}, GR-1 Tabletop, Real-world GR-1 manipulation (articulated object, multi-agent coordination) & Fourier GR-1 humanoid (real-world) & \makecell{10 Hz \\(system 2); \\120 Hz \\(system 1);}\\
\bottomrule
\end{tabular}%
}
\vspace{5pt}
\begin{minipage}{\textwidth}
\footnotesize
``N/A'' indicates not applicable; ``NR'' indicates not reported.
\end{minipage}
\end{table}

%% file: table/reasoning.tex
\begin{table}[!t]
\centering
\caption{Overview of VLA research using reasoning as action tokens.}
\label{tab:reasoning_action}
\renewcommand{\arraystretch}{1.3}
\resizebox{\textwidth}{!}{%
\begin{tabular}{
>{\centering\arraybackslash}m{2.8cm} 
>{\centering\arraybackslash}m{2.7cm} 
>{\centering\arraybackslash}m{4.0cm} 
>{\centering\arraybackslash}m{3.5cm}
>{\centering\arraybackslash}m{2.5cm} 
>{\centering\arraybackslash}m{4.0cm}
>{\centering\arraybackslash}m{4.0cm}
}
\toprule
\textbf{Paper} &  
\textbf{Model} & 
\textbf{Training Details} &  
\textbf{Token Format} & 
\textbf{Augmented Token} &  
\textbf{Task} &  
\textbf{Embodiment}
\\
\midrule
\makecell{Inner Monologue\\\cite{huang2022inner}} & PALM & Frozen & \makecell{Fixed simple reasoning} & \makecell{Language \\Description} & \makecell{Tabletop \\rearrangement; \\mobile manipulation} & \makecell{UR5e with a gripper;\\Everyday Robots} \\
\midrule
\makecell{DriveVLM~\cite{tian2024drivevlm}} & Qwen-VL-7B & \makecell{Pretrained and co-tuned on\\ self-constructed dataset} & \makecell{CoT in three modules} & \makecell{Trajectory} & \makecell{Autonomous driving} & \makecell{Li Auto autonomous vehicles} \\
\midrule
\makecell{ECoT~\cite{ECoT}} & OpenVLA & \makecell{Trained on 60K CoT\\ dataset constructed\\ from BridgeData V2} & \makecell{Fixed CoT \\reasoning steps} & \makecell{Raw Action} & \makecell{Pick, place, \\move, unfold cloth } & \makecell{WidowX with a gripper} \\
\midrule
\makecell{RAD~\cite{clark2025action}} & OpenVLA & \makecell{Trained on a mixture \\of robot data and 1616\\ action-free human demos\\ on BridgeData V2 Toy Sink setup} & \makecell{Fixed CoT \\reasoning steps} & \makecell{Raw Action} & \makecell{Pick, place} & \makecell{WidowX with a gripper} \\
\midrule
\makecell{AlphaDrive\textsuperscript{†} \\\cite{bo2025alpha}} & Qwen2-VL-2B & \makecell{SFT on MetaAD} & \makecell{CoT} & \makecell{Language \\Description} & \makecell{Autonomous driving} & \makecell{N/A} \\
\midrule
\makecell{Cosmos-Reason1\textsuperscript{†} \\\cite{nvidia2025cosmos}} & \makecell{Qwen2-VL-7B \\ Nemotron 56B} & \makecell{SFT on numerous \\self-constructed\\ datasets} & \makecell{CoT} & \makecell{Language \\Description} & \makecell{Physical common \\sense reasoning; \\embodied reasoning; \\intuitive physics reasoning} & \makecell{N/A} \\
\bottomrule
\end{tabular}
}
\vspace{5pt}
\begin{minipage}{\textwidth}
\footnotesize
\textsuperscript{†} These methods do not perform action grounding or execution. Included for their relevance to VLA perception and representation learning. For the same reason, these methods are N/A (not applicable) in the embodiment column.
\end{minipage}
\end{table}

%% file: table/data_sources.tex
\begin{table}[htbp]
\centering
\small
\caption{Overview of datasets used in VLA research.}
\renewcommand{\arraystretch}{1.2}
\resizebox{\textwidth}{!}{
\begin{tabular}{
>{\centering\arraybackslash}m{1.2cm} 
>{\centering\arraybackslash}m{2.0cm} 
>{\centering\arraybackslash}m{4.0cm} 
>{\centering\arraybackslash}m{4.0cm}
>{\centering\arraybackslash}m{3.9cm} 
>{\centering\arraybackslash}m{3.9cm} 
}
\toprule
\large \textbf{Category} & \large \textbf{Name} & \large \textbf{Description} & \large \textbf{Quantity} & \large \textbf{Data Structure} & \large \textbf{Used in VLA Papers} \\
\midrule
\multirow{3}{*}[-0.7cm]{\centering\rotatebox[origin=c]{90}{Human Video}}
& Something-Something V2~\cite{goyal2017something} & a large collection of labeled video clips with humans performing everyday actions & 220K video clips & videos, labels, objects & \makecell{LAPA~\cite{ye2024latent},\\ Magma~\cite{yang2025magma},\\ CoT-VLA~\cite{zhao2025cot}} \\
\cmidrule(lr){2-6}
& Ego4D~\cite{Ego4D2022CVPR} & an egocentric dataset of daily-life activity videos & 3670 hours & audio, 3D meshes, eye gaze, stereo, synchronized videos from multiple cameras & \makecell{EmbodiedGPT~\cite{mu2023embodiedgpt},\\ Magma~\cite{yang2025magma},\\ GR00T N1~\cite{nvidia2025gr00tn1openfoundation}}\\
\cmidrule(lr){2-6}
& Ego-Exo4D~\cite{grauman2024ego} & a multi-modal and multi-view video dataset of skilled human activities & 1286 hours & videos, 7-channel audio, IMU, eyegaze, two grayscale SLAM cameras, point clouds & GR00T N1~\cite{nvidia2025gr00tn1openfoundation} \\
\cmidrule(lr){2-6}
& EPIC-KITCHENS-100~\cite{Damen2021RESCALING} & an egocentric dataset of daily activities in the kitchen & \makecell{100 hours, 20M frames,\\ 90K action segments,\\ 200 participants} & \makecell{narrations, videos,\\ action segments} & ARM4R~\cite{niu2025pre}, Magma~\cite{yang2025magma}, 3D-VLA~\cite{zhen20243d}, CoT-VLA~\cite{zhao2025cot}, GR00T N1~\cite{nvidia2025gr00tn1openfoundation} \\
\midrule
\multirow{2}{*}[-0cm]{\centering\rotatebox[origin=c]{90}{Synthetic \& Simulation Data}}
& MimicGen~\cite{mandlekar2023mimicgen} & a dataset generation systems to produce diverse demonstrations from a few human demonstrations & 50K demons generated from \textasciitilde200 human demos & robot states, actions, scene/object configuration, camera data & - \\ \cmidrule(lr){2-6}
& RoboCasa~\cite{nasiriany2024robocasa} & a large-scale cross-embodiment simulation framework in kitchen environments & \makecell{100K trajectories,\\ 120 kitchen scenes,\\ 2500 object categories} & \makecell{action, images, \\proprioception, \\object states,\\ task annotations} & GR00T N1~\cite{nvidia2025gr00tn1openfoundation} \\
\cmidrule(lr){2-6}
& AgiBot Digital World~\cite{contributors2025agibotdigitalworld} & a high-fidelity digital twin-based simulation suite & \makecell{1M demonstrations,\\ 2976 hours,\\ 100 tasks} & RGB images, tactile, proprioceptive, low-level action commands & - \\
\midrule
\multirow{11}{*}[-4.5cm]{\rotatebox[origin=c]{90}{Real-World Robotic data}}
& RT-1~\cite{Brohan2022RT1RT} & a large, multi-embodiment robotic dataset including multiple tasks, objects and environments & 130K demonstrations, 9 skills, 700 tasks & action, images, language instruction, proprioception & \makecell{Gen2Act~\cite{bharadhwaj2024gen2acthumanvideogeneration},\\ RT-1~\cite{Brohan2022RT1RT}}\\
\cmidrule(lr){2-6}
& OXE~\cite{RT-X} & a large, multi-embodiment robotic dataset assembling 60 datasets & 1M episodes, 311 scenes, 22 robots, 527 skills, 60 datasets, 5228 objects & depends on the original dataset & 
3D-VLA~\cite{zhen20243d}, CoT-VLA~\cite{zhao2025cot}, RT-X~
\cite{RT-X}, Octo~\cite{Octo}, OpenVLA~\cite{kim2024openvla}, $\pi_0$~\cite{black2410pi0}, RDT~\cite{liu2024rdt}, etc. \\
\cmidrule(lr){2-6}
& RH20T~\cite{fang2023rh20t} & a multi-embodiment robotic dataset with visual, force, audio, and action information & 110K robot episodes, 110K human demonstrations & RGB-D images, Binocular IR images, action, audio, tactile (partly) & 3D-VLA~\cite{zhen20243d} \\
\cmidrule(lr){2-6}
& RoboMIND~\cite{wu2024robomind} & a large, multi-embodiment dataset including failure cases & 107K successful trajectories, 5K failed trajectories, 479 tasks, 96 objects, 4 robots & RGB-D images, action, language instruction & HybridVLA~\cite{HybridVLA}, AgiBot World Colosseo~\cite{bu2025agibot} \\
\cmidrule(lr){2-6}
& HoNY~\cite{shafiullah2023bringing} & a dataset containing interactions at home with the \textit{Stick} & 5K trajectories, 13 hours, 216 environments & RGB-D videos, action & 3D-VLA~\cite{zhen20243d} \\
\cmidrule(lr){2-6}
& BridgeData V2~\cite{walke2024bridgedatav2datasetrobot} & a large robotic dataset with the WidowX 250 robot arm & 60K episodes, 24 environments, 13 skills & videos (an over-the-shoulder RGB-D camera, two random RGB cameras, a wrist RGB camera), language instruction & LAPA~\cite{ye2024latent}, FLIP~\cite{flip}, AVDC~\cite{ko2023avdc}, HAMSTER~\cite{li2025hamster}, SuSIE~\cite{black2023zeroshotroboticmanipulationpretrained}, RoboDual~\cite{bu2024towards} \\
\cmidrule(lr){2-6}
& DROID~\cite{khazatsky2025droidlargescaleinthewildrobot} & a diverse robotic manipulation dataset of a Franka Panda 7DoF robot arm & 76K trajectories, 350 hours, 564 scenes, 86 tasks & action, language, 3 RGB cameras, proprioception & HAMSTER~\cite{li2025hamster}, Diffusion-VLA~\cite{wen2024diffusion}, Hi Robot~\cite{shi2025hi}, RoboDual~\cite{bu2024towards}, RAM~\cite{kuangram} \\
\cmidrule(lr){2-6}
& AgiBot World~\cite{contributors2024agibotworldrepo} & a large-scale robotic manipulation dataset with Genie-1 & 1M trajectories, 3K hours, 100 scenes, 5 domains, 200 tasks, 87 skills & RGB-D videos, action, skill, proprioception & FLaRe~\cite{hu2024flare} \\
\cmidrule(lr){2-6}
& WOMD~\cite{Ettinger_2021_ICCV, Kan_2024_icra} & a diverse interactive motion dataset for autonomous driving & 103K segments, 20 seconds each, 574 hours & ego pose, images, object tracks, 3D bounding box, Lidar data & EMMA~\cite{hwang2024emma} \\
\cmidrule(lr){2-6}
& nuScenes~\cite{fong2021panoptic} & a large dataset for autonomous driving & \makecell{1K driving scenes,\\ 20 seconds each} & ego pose, image, Lidar, Radar, object 3D bounding box & EMMA~\cite{hwang2024emma}, VLM-E2E~\cite{liu2025vlm} \\
\cmidrule(lr){2-6}
& CoVLA~\cite{covla_wacv2025} & a comprehensive Vision-Language-Action dataset for autonomous driving & 80 hours, 10K video clips & videos, frame-level language captions, future trajectory actions & CoVLA~\cite{covla_wacv2025} \\

\bottomrule
\end{tabular}
}
\label{tab:vla-datasets}
\end{table}